\documentclass[]{style}

\usepackage[toc,page,header]{appendix}

\usepackage{minitoc}
\usepackage{multirow}
\usepackage{graphicx}
\usepackage{array}
\usepackage{makecell}
\usepackage{framed}
\usepackage{hyperref}
\usepackage{amsmath} 
\usepackage[colorinlistoftodos]{todonotes}
\usepackage{longtable}
\usepackage{hhline}
\usepackage{fancyvrb}
\usepackage{fvextra}
\usepackage{CJKutf8}
\usepackage{multicol}
\usepackage{cleveref}
\usepackage{tablefootnote}
\usepackage{threeparttable}
\usepackage{tabularx}
\usepackage{mdframed}
\usepackage{subcaption}
\usepackage[usestackEOL]{stackengine}
\usepackage[numbers]{natbib}
\newcommand{\commentout}[1]{}
\renewcommand{\paragraph}[1]{\noindent\textbf{#1.}\hspace*{1em}}
\usepackage{enumitem}
\setlist[itemize]{leftmargin=15pt}

\usepackage{colortbl}
\PassOptionsToPackage{table}{xcolor}

\usepackage{marvosym}

\RequirePackage{xspace}
\makeatletter
\DeclareRobustCommand\onedot{\futurelet\@let@token\@onedot}
\def\@onedot{\ifx\@let@token.\else.\null\fi\xspace}

\def\eg{\emph{e.g}\onedot}

\makeatother

\title{RoboBrain 2.0 Technical Report}

\author{BAAI RoboBrain Team}

\contribution{Please see \hyperref[sec:contributions]{Contributions and Author List} for more author details.}

\abstract{
We introduce \textbf{RoboBrain 2.0}, our latest generation of embodied vision-language foundation models, designed to unify perception, reasoning, and planning for complex embodied tasks in physical environments. It comes in two variants: a lightweight 7B model and a full-scale 32B model, featuring a heterogeneous architecture with a vision encoder and a language model. Despite its compact size, RoboBrain 2.0 achieves strong performance 
across a wide spectrum of embodied reasoning tasks. On both spatial and temporal benchmarks, the 32B variant achieves leading results, surpassing prior open-source and proprietary models.
In particular, it supports key real-world embodied AI capabilities, including spatial understanding (\eg, affordance prediction, spatial referring, trajectory forecasting) and temporal decision-making (\eg, closed-loop interaction, multi-agent long-horizon planning, and scene graph updating). This report details the model architecture, data construction, multi-stage training strategies, infrastructure and practical applications. We hope RoboBrain 2.0 advances embodied AI research and serves as a practical step toward building generalist embodied agents. The code, checkpoint and benchmark are available at \url{https://superrobobrain.github.io}. \vspace{-3em}
}

\begin{document}
\maketitle

\vspace{-2em}
\begin{figure*}[h]
    \centering
    \includegraphics[width=1.0\linewidth]{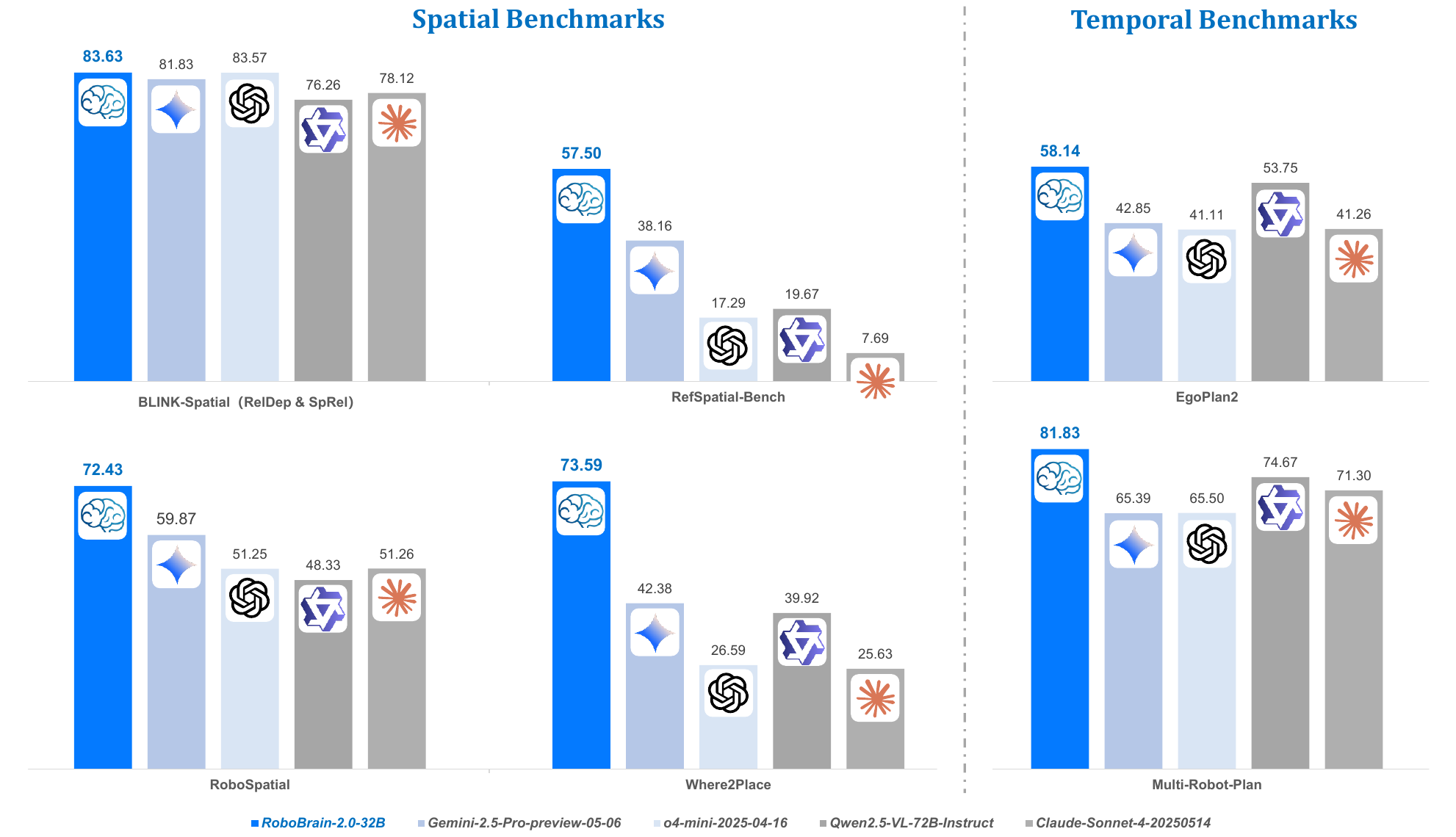}
    \vspace{-2em}
    \caption{\textbf{Benchmark comparison across spatial and temporal reasoning.} RoboBrain2.0-32B achieves best performance on both spatial and temporal reasoning benchmarks across BLINK-Spatial, RoboSpatial, RefSpatial-Bench, Where2Place, EgoPlan2 and Multi-Robot-Plan, outperforming prior open-source models and proprietary models.}
    \label{fig:teasor}
\end{figure*}

\newpage
\tableofcontents
\newpage

\section{Introduction}

In recent years, large language models (LLMs) and vision-language models (VLMs) have emerged as key driving forces in the advancement of general artificial intelligence (AGI). Within digital environments, these models have demonstrated remarkable capabilities in perception~\cite{qwen25vl,internvl3,gemini25pro}, understanding~\cite{yang2025qwen3,gpt4o}, and reasoning~\cite{seed15,deepseekr1,kimi15k15,gpto3-o4-mini,claude4}, and have been widely applied in tasks such as multimodal question answering~\cite{talmor2021multimodalqa,lu2022learn}, image generation and editing~\cite{kawar2023imagic,sheynin2024emu}, GUI control~\cite{wang2024gui,luo2025gui}, and video understanding~\cite{tang2025video,wang2024internvideo2,chen2024sharegpt4video}. They have also seen early adoption in practical domains such as education, healthcare, search, and intelligent assistants~\cite{huang2024comprehensive,zhou2024large,chu2025llm}.

\vspace{0.3em}

However, bridging the gap between ``digital intelligence'' and ``physical intelligence''—enabling models to perceive their surroundings, understand embodied tasks, and interact with the real world—remains a critical challenge on the path toward AGI. Embodied foundation models~\cite{geminirobotics,cosmos-reason1,magma} represent a promising research direction toward physical intelligence. Several recent efforts have extended the capabilities of LLMs and VLMs to embodied scenarios, advancing multimodal fusion, perception, and action execution. While these models have achieved encouraging progress, they still face three fundamental capability bottlenecks when deployed in complex and open-ended real-world environments:
\textbf{(1) Limited spatial understanding:} Current models struggle to accurately model relative and absolute spatial relationships and identify affordances in physical environments, which hinders real-world applicability;
\textbf{(2) Weak temporal modeling:} The lack of understanding of multi-stage, cross-agent temporal dependencies and feedback mechanisms limits long-horizon planning and closed-loop control;
\textbf{(3) Insufficient reasoning chains:} Existing models are often incapable of extracting causal logic from complex human instructions and aligning it with dynamic environmental states, restricting their generalization to open-ended embodied tasks.

\vspace{0.3em}

\begin{figure*}[t]
    \centering
    \includegraphics[width=1.0\linewidth]{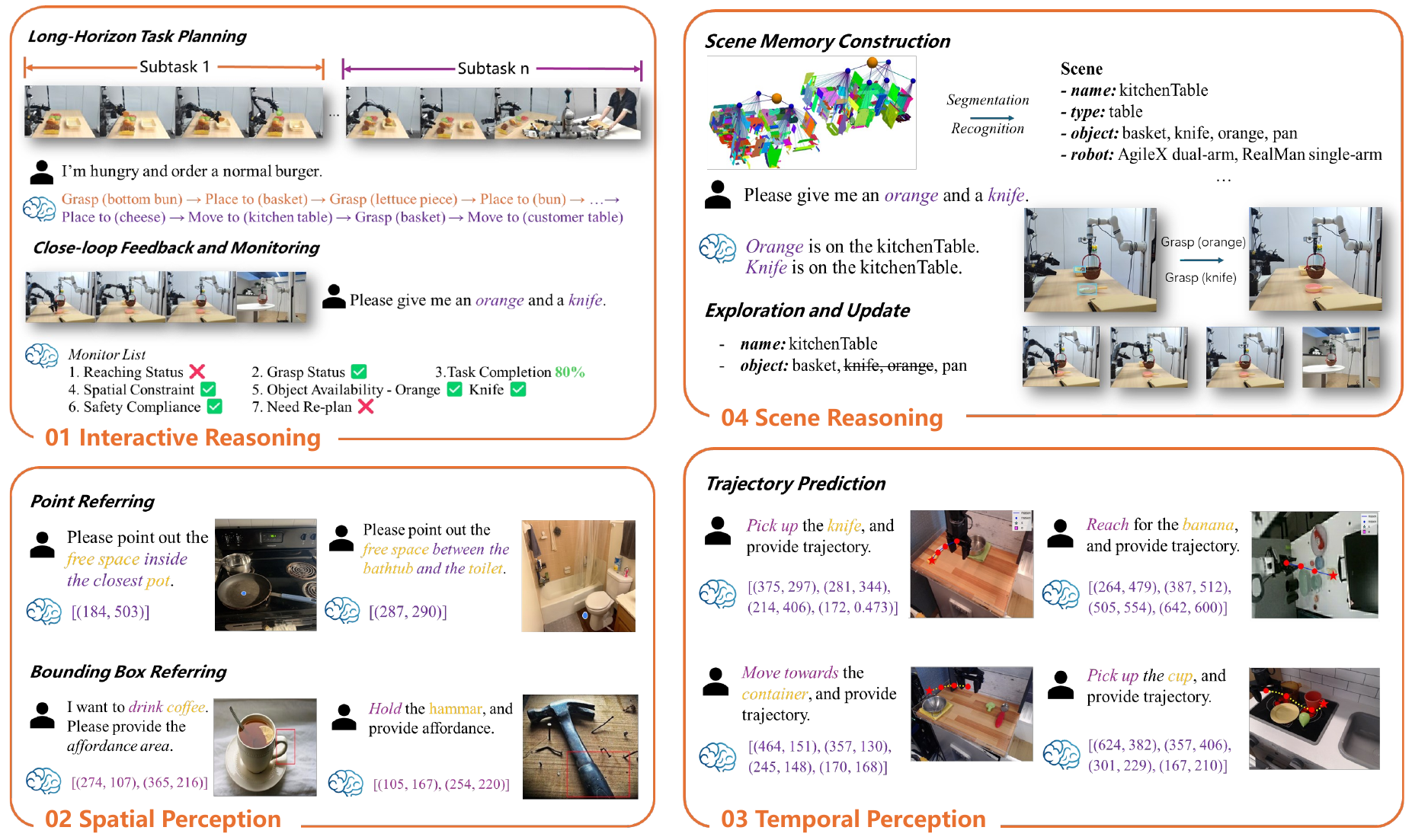}
    \caption{\textbf{The overview of RoboBrain 2.0's Capabilities.} RoboBrain 2.0 supports interactive reasoning with long-horizon planning and closed-loop feedback, spatial perception for precise point and bounding box prediction from complex instructions, temporal perception for future trajectory estimation, and scene reasoning through real-time scene graph construction and updating.}
    \vspace{-1.5em}
    \label{fig:task_cap}
\end{figure*}

To address these challenges, we present \textbf{RoboBrain 2.0}, our latest generation of embodied vision-language foundation models, tailored to bridge perception, reasoning, and planning in physically environments. RoboBrain 2.0 processes visual observations and language instructions in a unified architecture, enabling holistic understanding of the environment, goal-directed reasoning, and long-horizon planning. We release two variants of the model: the lightweight \textit{RoboBrain 2.0–7B} and the full-scale \textit{RoboBrain 2.0–32B}, designed to meet different deployment needs under varying resource constraints. On both spatial reasoning and temporal reasoning benchmarks, the 32B variant mostly achieves state-of-the-art performance, outperforming prior open-source and proprietary models, as shown in~\Cref{fig:teasor}. Model capabilities are summarized in~\Cref{fig:task_cap}.

\vspace{0.3em}

This report provides a systematic overview of the design principles, core components and key innovations. In particular, we highlight the extensive data contributions that support spatial understanding, temporal reasoning, and causal inference, which form the foundation of RoboBrain 2.0’s capabilities. To address the scarcity of spatial data, we develop a spatial data synthesis pipeline that constructs large-scale, high-quality datasets spanning tasks such as pointing, affordance prediction, and trajectory generation. To improve temporal reasoning and feedback modeling, we design multi-robot coordination templates across common scenarios via RoboOS~\cite{tan2025roboos}, generate cross-agent long-horizon planning trajectories using external models~\cite{deepseekv3}, and simulate randomized failure events to collect closed-loop feedback data that enhances model robustness. To further enrich reasoning data, we extract step-by-step thought traces from powerful reasoning VLMs~\cite{gpt4o}, conditioned on spatiotemporal task contexts. These traces serve as supervision signals for learning causal chains across vision, language, and action. 
 
\vspace{0.3em}

RoboBrain 2.0 adopts a high-efficiency heterogeneous architecture and a progressive multi-stage training strategy to support spatial understanding, temporal modeling, and long-chain causal reasoning in embodied settings. The model comprises a lightweight vision encoder with approximately \texttt{689M} parameters and a decoder-only language model with \texttt{7B/32B} parameters. It is trained using a three-stage curriculum—covering foundational spatiotemporal learning, embodied spatiotemporal enhancement, and chain-of-thought reasoning—on large-scale multimodal and embodied datasets. Training is conducted using our open-source framework \textit{FlagScale}, which integrates hybrid parallelism, pre-allocated memory optimization, high-throughput I/O pipelines, and robust fault tolerance. These infrastructure innovations significantly reduce training and deployment costs while ensuring scalability for large-scale multimodal models. We evaluate RoboBrain 2.0 on over \texttt{12} public benchmarks covering spatial understanding, temporal modeling and multimodal reasoning, achieving state-of-the-art results on \texttt{6} of them despite its compact size. We release code, checkpoints, and benchmarks as open-source resources to benefit the research community. These materials facilitate reproducible research, accelerate embodied AI development, and enable practical deployment in robotic systems.

\vspace{0.3em}

To provide a comprehensive view of RoboBrain 2.0’s architecture, training methodology, and capabilities, this report is organized as follows: \Cref{sec:arch} introduces the overall model design, including the coordination between the vision encoder and language model, as well as image and video input strategies. \Cref{sec:train_data} describes the data curation and construction process, covering three major categories: general multimodal understanding, spatial reasoning, and temporal modeling. \Cref{sec:trainstrategy} presents our multi-stage training strategies, including foundational spatiotemporal learning, embodied enhancement, and chain-of-thought reasoning. \Cref{sec:infrastructure} outlines the infrastructure stack supporting scalable training and inference, including hybrid parallelization, memory optimization, data loading, and failure recovery. \Cref{sec:evaluation} reports extensive evaluation results on public benchmarks, highlights RoboBrain 2.0’s capabilities in spatial reasoning, temporal feedback, and embodied planning. Finally, \Cref{sec:conclusion} discusses current limitations, and outlines future research directions.

\section{Architecture}
\label{sec:arch}

\begin{figure*}[t]
    \centering
    \includegraphics[width=1.0\linewidth]{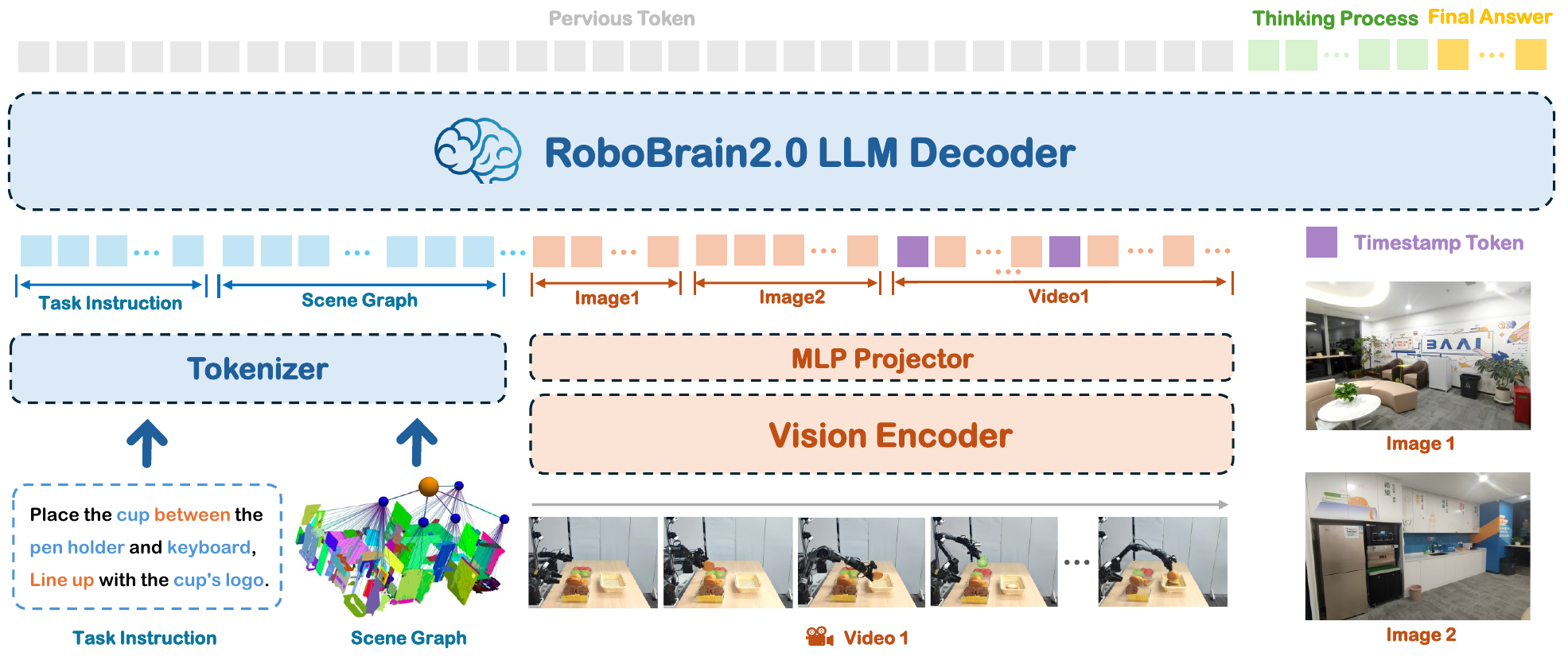}
    \caption{\textbf{The Architecture of RoboBrain 2.0}. The model supports multi-image, long video, and high-resolution visual inputs, along with complex task instructions and structured scene graphs on the language side. Visual inputs are processed via a vision encoder and an MLP projector, while textual inputs are tokenized into a unified token stream. All inputs are fed into an LLM decoder that performs long-chain-of-thought reasoning and generates a variety of outputs depending on the task, including structured plans, spatial relations, or relative and absolute coordinates.}
    \label{fig:model}
    \vspace{-1.5em}
\end{figure*}

RoboBrain 2.0 employs a modular encoder-decoder architecture that unifies perception, reasoning, and planning for complex embodied tasks. As shown in~\Cref{fig:model}, it processes multi-view visual observations and natural language instructions through four core components: (1) a tokenizer for textual/structured inputs, (2) a vision encoder, (3) an MLP projector mapping visual features to the language model's token space, and (4) a language model backbone initialized from Qwen2.5-VL~\cite{qwen25vl}.
Unlike conventional VLMs~\cite{gpt4o,claude4} focused on general static VQA, RoboBrain 2.0 maintains strong general VQA capabilities while specializing in embodied reasoning tasks like spatial perception, temporal modeling, and long-chain causal reasoning. The architecture encodes high-resolution images, multi-view inputs, video frames, language instructions, and scene graphs into a unified multimodal token sequence for comprehensive processing.

\subsection{Input Modalities and Tokenization}

RoboBrain 2.0 supports a diverse set of input modalities tailored for embodied AI tasks:

\vspace{0.3em}

\begin{itemize}
    \item \textbf{Language instructions:} Natural language commands describing high-level goals or low-level actions. RoboBrain 2.0 processes natural language commands spanning different abstraction levels: from high-level, spatially grounded instructions (\eg, \textit{``Carry the apple to the nearest table, aligned with the leftmost cup''}) to low-level motor commands (\eg, \textit{``Navigate to the nearest table'', ``Grasp the apple'', ``Detect position aligned with the leftmost cup'', ``Place the apple into the box''}).
    
    \item \textbf{Scene graph:} A structured JSON representation of the explored environment, containing information about discovered objects, their categories, spatial locations, and embodiment configuration (\eg, \texttt{name: KitchenTable1, type: table, object: [basket, knife], robot: RealMan-single-arm}).
    
    \item \textbf{Multi-view static images:} Images captured from multiple viewpoints, such as head-mounted cameras, wrist-mounted cameras, or multi-view projections from a 3D environment. These are processed independently by the vision encoder and concatenated into a unified token sequence.
    
    \item \textbf{Video frames:} Video sequences (\eg, egocentric views from the agent), optionally annotated with timestamp tokens~\cite{qwen25vl} to facilitate temporal grounding and reasoning.
\end{itemize}

\vspace{0.3em}

Language instructions and scene graphs are tokenized using the language tokenizer. Visual inputs—including multi-view images and video frames—are processed by the vision encoder into dense visual embeddings, which are then projected into the LLM's token space through an MLP projector, enabling unified multi-modal reasoning within the decoder.

\subsection{Vision Encoder and Projection}

RoboBrain 2.0 vision encoder supports dynamic-resolution image and video inputs through adaptive positional encoding and windowed attention mechanisms~\cite{qwen25vl}. This design choice enables efficient processing of high-resolution and multi-view visual observations common in embodied tasks. 

\vspace{0.3em}

To accommodate the long-horizon and temporally grounded nature of such tasks, we adopt frame-wise visual tokenization with multi-dimensional RoPE~\cite{qwen25vl} for spatiotemporal encoding. Each visual embedding is projected via a lightweight MLP into the token space of the language model. For multi-view scenarios, visual tokens from different camera perspectives are serialized and augmented with view-specific positional identifiers before being fused with other input modalities.

\subsection{LLM Decoder and Output Representations}

RoboBrain 2.0 employs a decoder-only language model designed to unify high-level reasoning and spatially grounded output generation. Unlike conventional VLMs that primarily return short-form answers to static prompts, RoboBrain 2.0 flexibly supports both concise responses and multi-step chain-of-thought reasoning. This capability enables deeper understanding of complex instructions and physical scenes.

\vspace{0.3em}

To enable the decoder to handle embodied tasks, the decoder is trained to produce a diverse range of outputs, including semantically grounded expressions (\eg, referring to objects or actions), spatial coordinates (\eg, absolute positions or bounding boxes), and intermediate reasoning traces. Rotary positional encodings and temporally conditioned tokens allow the model to maintain coherence across multi-round perception-action loops, which are essential for long-horizon planning in dynamic environments. Output formats supported by RoboBrain 2.0 include:
\textbf{(1) Free-form text:} Used for task decomposition, scene graph updates, agent invocation, and human-agent dialogue.
\textbf{(2) Spatial coordinates:} Used to represent point locations, bounding boxes, or trajectories in the image space for downstream controllers.
\textbf{(3) Reasoning traces (Optional):} Long-chain-of-thought explanations to support deep problem solving and decision transparency.

\vspace{0.3em}

This unified decoding formulation allows RoboBrain 2.0 to effectively handle a wide range of embodied tasks, from spatial grounding and visual understanding to long-horizon multi-agent planning and causal reasoning.

\section{Training Data}
\begin{figure*}[h!]
  \centering
  \setlength{\abovecaptionskip}{0.5em}
  \setlength{\belowcaptionskip}{0em}
  \includegraphics[width=1.0\linewidth]{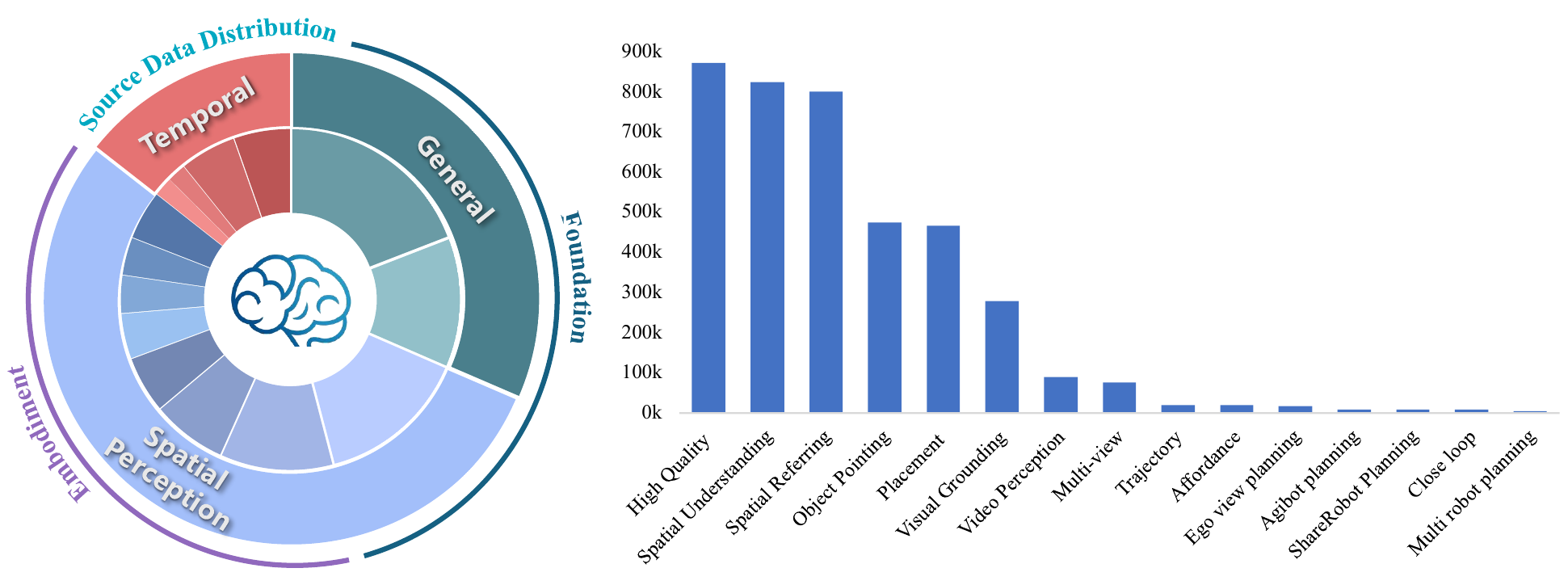}
  \caption{\textbf{Training Data Distribution for RoboBrain 2.0.} This figure illustrates the distribution of training data supporting RoboBrain 2.0’s capabilities, including interactive reasoning with long-horizon planning and closed-loop feedback, spatial perception for precise point and bounding box prediction from complex instructions, and multi-agent collaboration tasks, which is meticulously categorized into three primary types: general multimodal understanding, spatial perception, and temporal modeling.}
  \label{fig:train_data}
\end{figure*}

As shown in \Cref{fig:train_data}, RoboBrain 2.0 is trained on a diverse and extensive dataset designed to enhance its capabilities in spatial understanding, temporal modeling and long-chain causal reasoning in embodied settings. The training data encompasses a wide range of modalities, including high-resolution images, multi-view inputs, video sequences, scene graph and natural language instructions. This comprehensive dataset is meticulously categorized into three primary types: general multimodal understanding, spatial perception, and temporal modeling, ensuring the model can effectively perceive, reason, and plan in complex physical environments.

\label{sec:train_data}

\subsection{General MLLM VQA}
\textbf{High Quality Data.} The general training dataset for RoboBrain 2.0 includes 873K high-quality samples, primarily derived from LLaVA-665K~\cite{liu2024improved} and LRV-400K~\cite{liu2023mitigating}, spanning standard Visual Question Answering(VQA), region-level queries, OCR-based VQA, and visual dialogues. \textbf{(1) LLaVA-665K} serves as the primary source and contains diverse VQA-style data, including standard VQA datasets, OCR-based questions, region-level queries, visual conversations, and language-only dialogues. To improve training efficiency, multiple question-answer(QA) pairs from the same image merge into single conversations; invalid ShareGPT~\cite{vicuna_sharegpt} entries are filtered out, and overly long conversations (>2048 tokens) are truncated (resulting in 40K valid samples). Specifically, A-OKVQA~\cite{schwenk2022okvqa} samples are augmented by duplicating choices to balance multiple-choice formats, OCR-VQA~\cite{mishra2019ocr} contributes 80K sampled conversations focused on scene text understanding, Visual Genome(VG)~\cite{krishna2017visual} provides dense object-level annotations limited to 10 entries per image with additional captions, and RefCOCO~\cite{refcoco} dialogues are split into short multi-turn segments (<10 exchanges). Language-only conversations, which are generally longer than visual ones, are sampled in single-modality batches to improve throughput by 25\% without performance degradation. After removing bounding-box-dependent QA pairs, 531K high-quality samples are retained from this source.
\textbf{(2) LRV-400K} is synthetically generated using GPT-4~\cite{gpt4} under a few-shot instruction-following setting. It produces 400K image-conditioned instructions across 16 vision-language tasks with textual answers. Unlike prior works that rely on sparse image captions, this dataset leverages the dense annotations in VG (\eg, bounding boxes, dimensions, and $\sim$21 object regions per image). GPT-4 generates both declarative and interrogative prompts for each image, with 10 tasks randomly sampled per instance. After filtering out bounding-box-related QA pairs, 342K samples are selected for training.

\subsection{Spatial Data}

\textbf{Visual Grounding.}
The visual grounding dataset is constructed to enhance multimodal understanding through precise object-level localization, leveraging the extensive annotations from LVIS~\cite{gupta2019lvis}. We carefully curate 152K high-resolution images from LVIS, ensuring broad coverage of diverse object categories and complex visual scenes. Each object annotation is converted into standardized bounding box coordinates \((x_1,y_1,x_2,y_2)\) representing the top-left and bottom-right corners, enabling consistent spatial referencing. To facilitate rich visual dialogue, we generated 86K conversational sequences, each containing multiple rounds of QA pairs that progressively explore visual relationships, attribute reasoning, and contextual understanding. The dataset maintains a balanced distribution across object categories while preserving challenging cases of occlusion, viewpoint variation, and rare instances to support robust visual grounding. 

\vspace{+1mm}
\textbf{Object Pointing.}
The object pointing dataset is constructed to enable RoboBrain 2.0 to identify the locations of specified objects through pointing within an image. We leverage the Pixmo-Points~\cite{deitke2024molmo} dataset, which includes 2.3M point annotations across 223K images as our data source. However, direct utilization of Pixmo-Points data for RoboBrain 2.0 training presents challenges due to densely repeated object instances (\eg, books on a shelf). To address this, we implement a two-step filtering process: (1) we discard annotations with more than ten labeled points to simplify training, and (2) we use GPT-4o~\cite{gpt4o} as a scene analyzer to select only indoor-relevant objects, such as kitchenware, furniture, and decorations, excluding irrelevant or outdoor scenes. This process yields 190K QA pairs for 64K images with reduced clutter, making the data more suitable for embodied contexts. 
To construct QA pairs for pointing tasks, we construct 28 human-designed templates, such as ``\textit{Point out all instances of \{label\} in the image.}'' or ``\textit{Help me find \{label\} in the image by pointing to them.}'' Here, \textit{\{label\}} refers to object categories from the annotations. Templates are randomly selected to ensure linguistic diversity and improve the model's generalization ability in referencing tasks.
For object reference pointing, we incorporate object reference data sourced from RoboPoint~\cite{yuan2024robopointvisionlanguagemodelspatial}, which includes 347K QA annotations across 288K images. To address the potential issue of excessive points hindering training convergence, we randomly sample up to ten points per question. Additionally, the normalized coordinates are converted into absolute values to better support RoboBrain 2.0 training.

\vspace{+1mm}
\textbf{Affordance.}
The affordance dataset focuses on understanding object functionality and spatial vacant areas for placement. For object affordance recognition, we utilize part-level annotations from PACO-LVIS~\cite{ramanathan2023paco}, covering 75 object categories and 200 part categories across 46K images. Bounding boxes and segmentation masks are extracted for both whole objects and their functional parts. These annotations are transformed into bounding box coordinates \((x_1,y_1,x_2,y_2)\), serving as ground truth labels for affordance prediction tasks. Questions are constructed using GPT-4o~\cite{gpt4o} to query object functionality and part usage, \eg, ``\textit{Which part of a handbag can be grasped to carry it?}'' for the handle of a handbag. For whole-object affordances, questions avoid naming the object directly, such as ``\textit{What device can be moved to control the cursor on a screen?}'' for a mouse (computer equipment). This automatic process results in 561K QA pairs.
For spatial affordance learning, we include region reference data from RoboPoint~\cite{yuan2024robopointvisionlanguagemodelspatial}. This dataset consists of 270K images with 320K QA pairs and 14 spatial relationship labels. Each annotation is converted into a set of absolute coordinates \([(x_1,y_1),(x_2,y_2),...]\), and ground truth points are resampled to a maximum of ten points per answer for optimization. This dataset enables RoboBrain 2.0 to reason about spatial affordances for object placement in real-world settings.

\vspace{+1mm}
\textbf{Spatial Understanding.}
To enhance RoboBrain 2.0’s 3D spatial reasoning, we present the Spatial Understanding Dataset, comprising 826K samples. This dataset emphasizes object-centric spatial attributes (\eg, position, orientation) and inter-object relations (\eg, distance, direction), covering both qualitative and quantitative aspects. 
It covers 31 distinct spatial concepts, substantially surpassing the $\sim$15 typically found in previous datasets.
We partially adopt the RefSpatial~\cite{zhou2025roborefer} pipeline to construct 2D web image and 3D video datasets via automated template- and LLM-based generation: 
\textbf{(1) 2D web images} aim to provide core spatial concepts and depth perception across diverse indoor and outdoor scenes. To bridge scale and category gaps between these domains, we utilize the large-scale OpenImage~\cite{kuznetsova2020open} dataset. Since direct 3D reasoning from 2D images is challenging, we convert them into pseudo-3D scene graphs. Specifically, after filtering 1.7M images to 466K, we first use RAM~\cite{zhang2024recognize} for object category prediction and GroundingDINO~\cite{liu2024grounding} for 2D boxes Detection. Then we enhance using Qwen2.5-VL~\cite{qwen2.5-vl} and a heuristic method to generate hierarchical captions given the 2D bounding box, ranging from coarse (\eg, ``cup'') to fine-grained (\eg, ``the third cup from the left''). This enables unambiguous spatial referring in cluttered environments and captures both coarse and fine-grained spatial references. Next, we use UniDepth V2~\cite{piccinelli2025unidepthv2} and WildeCamera~\cite{zhu2023tame} for depth and camera intrinsics to enable 3D point cloud reconstruction. Finally, combining this with object boxes from GroundingDINO~\cite{liu2024grounding} and masks from SAM 2.1~\cite{ravi2024sam}, each scene graph includes object labels, 2D boxes, instance masks, and object-level point clouds, yielding axis-aligned 3D boxes. Object captions serve as nodes, and spatial relations form the edges. QA pairs are generated via templates and LLMs (\eg, QwQ~\cite{qwq32b}), including object-location questions derived from the hierarchical captions.
\textbf{(2) 3D scene-based videos} integrates multimodal 3D scene understanding data from five original datasets: MMScan~\cite{mmscan}, 3RScan~\cite{3rscan}, ScanQA~\cite{scanqa}, SQA3D~\cite{sqa3d}, and SpaceR~\cite{ouyang2025spatial}. We conduct template-based question filtering through rigorous data processing to ensure task relevance, perform multi-stage quality screening (\eg, consistency checks, outlier removal), and standardize all formats into a unified representation. This curation enables fine-grained environmental perception with enhanced reliability, supporting tasks ranging from object localization to complex spatial reasoning in 3D scenes.
\textbf{(3) 3D embodied videos} focus on fine-grained spatial understanding in indoor environments. We leverage the CA-1M~\cite{lazarow2024cubify} dataset, filtering 2M frames to 100K high-quality ones. Compared to 2D, the availability of accurate 3D bounding boxes allows us to construct richer scene graphs with more diverse spatial relations, thereby generating more quantitative QA pairs (\eg, size, distances). 

\vspace{+1mm}
\textbf{Spatial Referring.}
After enhancing foundational 3D spatial understanding, we extend these capabilities to physical-world interactions by introducing the Spatial Referring Dataset~\cite{zhou2025roborefer}, consisting of 802K samples. Unlike prior datasets in visual grounding or object pointing, which often deal with ambiguous or multiple referents, this dataset targets a single unambiguous target, aligning with robotic applications such as precise pick-and-place that demand accurate object identification and localization.
Following the RefSpatial~\cite{zhou2025roborefer} construction pipeline, for location data, we sample caption-point pairs from scene graphs built on 2D web images (OpenImage~\cite{kuznetsova2020open}) and 3D embodied videos (CA-1M~\cite{lazarow2024cubify}), using hierarchical captions. For placement data, we leverage fully annotated 3D datasets to generate top-down occupancy maps encoding object positions, orientations, and metric spatial relations (\eg, ``10cm right of the chair''), facilitating accurate spatial referring.

\subsection{Temporal Data}

\vspace{+1mm}
\textbf{Ego-View Planning.}
We construct Ego-View Planning dataset by partially processing the EgoPlan-IT~\cite{chen2024egoplanbenchbenchmarkingmultimodallarge} dataset, which contains 50K automatically generated samples. For each selected task instance, we extract multiple frames from prior actions to represent task progress, and one frame to capture the current viewpoint. To enhance linguistic variety, we use multiple prompt templates that describe the task goal, video context, and current observation. Each question includes the correct next action along with up to three distractor actions randomly sampled from negative examples. This setup supports multimodal instruction tuning with diverse visual and textual input, aimed at improving egocentric task planning performance.

\vspace{+1mm}
\textbf{ShareRobot Planning.}
The ShareRobot dataset~\cite{ji2025robobrain} is a large-scale, fine-grained resource for robotic manipulation, offering multi-dimensional annotations tailored for task planning. Its planning component provides detailed low-level instructions aligned with individual video frames, effectively transforming high-level task descriptions into structured and executable sub-tasks. Each data instance includes precise planning annotations to support accurate and consistent task execution. The dataset comprises 1M QA pairs from 51K instances, spanning 102 diverse scenes across 12 robot embodiments and 107 atomic tasks filtered according to the Open-X-Embodiment taxonomy~\cite{o2024open}. All planning data were meticulously annotated by human experts following the RoboVQA~\cite{sermanet2024robovqa} format, enabling models to learn robust multi-step planning strategies grounded in diverse real-world scenarios. The scale, quality, and diversity of ShareRobot help improve the model's ability to perform fine-grained reasoning and task decomposition in complex embodied environments.

\vspace{+1mm}
\textbf{Agitbot Planning.}
The AgiBot Planning dataset is a large-scale robotics task planning dataset built upon the AgiBot-World~\cite{bu2025agibot} dataset, comprising 9,148 QA pairs across 19 manipulation tasks with 109,378 first-person perspective images. Each sample contains 4-17 consecutive frames documenting task progression with multimodal conversational format. AgiBot-Planning provides step-by-step planning instructions that transform high-level goals into executable sub-tasks. Each data point includes current objectives, historical steps, and required subsequent actions. The dataset covers diverse scenarios from household refrigerator operations to supermarket shopping tasks across different environments. The meticulously crafted annotations use standardized conversational formats, enabling models to learn from varied real-world contexts. Through continuous visual sequences and fine-grained action plans, AgiBot-Planning enhances RoboBrain 2.0’s ability to perform long-horizon task planning and spatial reasoning in complex embodied scenarios.

\vspace{+1mm}
\textbf{Multi-Robot Planning.}
The Multi-Robot Planning dataset is constructed by simulating collaborative task scenarios across three environments—household, supermarket, and restaurant—based on RoboOS~\cite{tan2025roboos}. Each sample is generated using structured templates that specify a detailed scene graph, robot specifications, and associated tool lists. For every scenario, we design high-level, long-horizon collaborative task goals that require coordination among multiple robots present in the scene, and generate corresponding workflow graphs that decompose the tasks into subtasks with detailed reasoning explanations. Based on these decompositions, we further generate agent-specific robotic tool plans that translate high-level task goals into precise low-level Observation-Action pairs for each subtask. Specifically, we define 1,659 types of multi-robot collaboration tasks across the three environments and produce 44,142 samples using DeepSeek-V3~\cite{deepseekv3}.

\vspace{+1mm}
\textbf{Close-Loop Interaction.}
The Close-Loop Interaction dataset is designed to facilitate advanced embodied reasoning~\cite{zhou2024code}, featuring a large-scale collection of synthesized Observation-Thought-Action (OTA) trajectories that combine first-person visual observations with structured thought tokens. It spans 120 diverse indoor environments—including kitchens, bathrooms, bedrooms, and living rooms—containing over 4,000 interactive objects and receptacles. The dataset is constructed within the AI2Thor~\cite{kolve2017ai2} simulator through a rigorous multi-stage pipeline based on Embodied-Reasoner~\cite{zhang2025embodied}, which includes: (1) crafting task instructions from constrained templates to ensure scene-appropriate validity; (2) deriving key action sequences from an object-affiliation graph encoding functional relationships; and (3) strategically incorporating search actions to emulate realistic exploration. To enrich the depth of reasoning, GPT-4o generates detailed thought processes—covering situational analysis, spatial reasoning, self-reflection, task planning, and verification—which are seamlessly integrated between observations and actions, forming coherent reasoning chains that guide models through complex, long-horizon interactive tasks.

\begin{table*}[!ht]
    \centering
    \caption{Detailed configuration for each training stage of the RoboBrain 2.0.}
    \vspace{0.2cm}
    \label{tab:training_setting}
    \setlength{\tabcolsep}{12pt}
    \renewcommand{\arraystretch}{1.2}
    \resizebox{0.93\textwidth}{!}{%
    \begin{tabular}{@{}ll|c|c|c|c}
        \toprule
        & & \textbf{Stage-1} & \textbf{Stage-2} & \multicolumn{2}{c}{\textbf{Stage-3}} \\ \cmidrule(l){3-3} \cmidrule(l){4-4} \cmidrule(l){5-6}
        & & \textbf{SFT} & \textbf{SFT} & \textbf{COT-SFT} & \textbf{RFT (RLVR)} \\
        \midrule 
        \multirow{2}{*}{\rotatebox[origin=c]{90}{\small \textit{Data}}}
        & \textbf{Dataset}  & Foundation & Embodied & Embodied (Phase 1) & Embodied (Phase 2) \\
        & \#Samples & 4.8M & 224K & 195K & 45K \\
        \midrule 
        \multirow{2}{*}{\rotatebox[origin=c]{90}{\small \textit{Model}}}
        & \textbf{Trainable Part} & Full Model & Full Model & Full Model & Full Model \\
        & \#Tunable Parameters & 8.29B or 33.45B & 8.29B or 33.45B & 8.29B or 33.45B & 8.29B or 33.45B \\
        \midrule 
        \multirow{14}{*}{\rotatebox[origin=c]{90}{\small \textit{Training}}}
        & \textbf{Per-device Batch Size} & 2 & 2 & 4 & 1 \\
        & \textbf{Gradient Accumulation} & 2 & 2 & 2 & 2 \\   
        & \textbf{LR: $\{\psi_v^{\text{ViT}}, \phi_v^{\text{LLM}}\}$} & 1$\times 10^{-4}$ & 1 $\times 10^{-5}$ & 1 $\times 10^{-5}$ & 1 $\times 10^{-6}$ \\
        & \textbf{Epoch} & 1 & 1 & 1 & 3 \\
        & \textbf{Optimizer} & AdamW & AdamW & AdamW & AdamW \\
        & \textbf{Deepspeed} & -- & -- & Zero3 & Zero3 \\
        & \textbf{Weight Decay} & 0.1 & 0.1 & 0.1 & 0.0 \\
        & \textbf{Warmup Ratio} & 0.01 & 0.01 & 0.03 & 0.00 \\
        & \textbf{LR Schedule} & Cosine & Cosine & Cosine & Cosine \\
        & \textbf{Max Seq. Length} & 16384 & 16384 & 32768 & 32768 \\
        & \textbf{Max Compl. Length} & -- & -- & -- & 1024 \\
        & \textbf{Num. of Compl.} & -- & -- & -- & 8 \\
        & \textbf{GPU Nums} & 16/64 $\times$ 8 & 16/64 $\times$ 8 & 4 $\times$ 8 & 4 $\times$ 8 \\
        \bottomrule
    \end{tabular}
    }
    \vspace{1mm}
    \vspace{-0.5em}
\end{table*}

\section{Training Strategy}
\label{sec:trainstrategy}

RoboBrain 2.0 achieves embodied capabilities (spatial understanding, temporal modeling, and chain-of-thought reasoning) through a progressive three-phase training strategy, as shown in~\Cref{tab:training_setting}. Starting from a robust vision-language foundation, we introduce escalating complexity in embodied supervision, enabling the model to evolve from static perception to dynamic reasoning and actionable planning in real-world environments.

\subsection{Stage 1: Foundational Spatiotemporal Learning}

The first stage focuses on building general capabilities in spatial perception and temporal understanding. We fine-tune the model on large-scale multimodal datasets covering dense captioning, object localization, interleaved image-text documents, and basic video QA, along with referring expression comprehension. These datasets span common physical scenes and interaction patterns, helping the model develop fundamental grounding for objects, spatial relations, and motion events. This stage lays the groundwork for understanding egocentric video streams and spatially anchored instructions.

\subsection{Stage 2: Embodied Spatiotemporal Enhancement}

To better align the model with embodied tasks, we introduce a carefully curated collection of high-resolution, multi-view, and egocentric video datasets, along with  instruction-augmented navigation and interaction data. Tasks include viewpoint-aware referring expressions, 3D affordance estimation, and object-centric scene graph construction.  
This stage of training emphasizes the modeling of long-horizon temporal dependencies, enabling the model to reason over extended sequences of actions and observations. Additionally, it incorporates multi-agent coordination scenarios, where the model learns to interpret and predict the behaviors of other agents in shared environments. To support these capabilities, we employ extended sequence lengths and multi-camera input encoding, allowing the model to process and fuse visual information from multiple viewpoints simultaneously. Through this training stage, the model can integrate historical visual cues with current instructions, fostering more coherent long-horizon planning, robust scene understanding, and adaptive decision-making in dynamic, interactive settings.

\subsection{Stage 3: Chain-of-Thought Reasoning in Embodied Contexts}

In the third stage, we augment the model’s high-level reasoning capabilities using Chain-of-Thought (CoT) methodology, following the two-phase framework of Reason-RFT~\cite{tan2025reason}: CoT-based Supervised Fine-Tuning (CoT-SFT) and Reinforcement Fine-Tuning (RFT). We leverage multi-turn reasoning examples from both synthetic and real-world embodied scenarios, encompassing long-horizon task planning, manipulation prediction, closed-loop interaction, spatiotemporal understanding, and multi-robot collaboration, sourced from~\Cref{sec:train_data}.
Specifically, (1) CoT-SFT Phase: We annotate 10\% of the constructed training data with CoT rationales annotated by GPT-4o~\cite{gpt4o} with custom prompts, then perform supervised fine-tuning for initial model from Stage 2. (2) RFT Phase: An additional 10\% of the constructed training data is sampled to collect model's responses, with incorrect answers curated into a reformatted training set (\eg, multiple-choice questions or LaTeX/numerical answers). Optimization employs Group Relative Policy Optimization (GRPO)~\cite{deepseekr1}, guided by a composite reward function evaluating both answer accuracy and format correctness.
\section{Infrastructures}
\label{sec:infrastructure}

\subsection{Large-Scale Training Infrastructure}
To improve the efficiency and stability of multimodal model training, we have developed and integrated a series of key optimization techniques, including hybrid parallelism strategies, memory pre-allocation, distributed data loading, kernel fusion, and fine-grained compute-communication overlapping. These optimizations significantly enhance both resource utilization and training throughput.
For data preprocessing, we build upon the Megatron–Energon framework~\cite{megatron-energon} and incorporate custom optimization strategies. Our system supports dynamic mixing of multiple datasets containing diverse modalities, including plain text, single image, multiple images, and video, while also allowing for strict sample order preservation within each dataset. A custom WebDataset-based format~\cite{webdataset} enables compatibility with various data modalities and greatly reduces preprocessing time while improving flexibility and scalability in data handling.

\subsubsection{Multi-Dimensional Hybrid Parallelism}
Multimodal models differ significantly from conventional LLMs in both architecture and data characteristics~\cite{liu2024improved}. On the architectural side, multimodal models are inherently heterogeneous: the vision module (\eg, ViT with Adaptor) is typically a small-scale encoder-only component, while the language module is a much larger decoder-only transformer. On the data side, training samples include plain text, single images, multi-image sequences, and videos. The number of image tokens, text tokens, and the length of the fused token sequence can vary dramatically between samples.

These heterogeneities pose substantial challenges to distributed training frameworks. To address this, we implemented several targeted strategies in our custom framework, FlagScale~\cite{flagscale}:
\begin{itemize}
    \item \textbf{Non-uniform Pipeline Parallelism}~\cite{megatron}: Since the ViT module appears early in the model and has relatively low computational cost, we reduce the number of LLM layers in the first pipeline stage, thereby improving training throughput without increasing memory overhead.
    \item \textbf{Separate Recompute Strategy}: During the annealing stage, the vision input may contain up to 20,000–30,000 tokens, frequently causing an Out-of-Memory (OOM) error in the ViT module. To mitigate this, we enable recompute~\cite{recompute,megatron-recompute} only in the ViT module to reduce memory usage of intermediate activations, while disabling recompute in the LLM module to preserve computational efficiency.
\end{itemize}

\subsubsection{Pre-Allocate Memory}
In the supervised fine-tuning training process of RoboBrain 2.0, input lengths vary significantly across samples. PyTorch’s default caching memory allocator~\cite{pytorch-memory} can lead to memory fragmentation under such dynamic input conditions, frequently resulting in OOM errors.
A common but inefficient workaround is to call \texttt{torch.cuda.empty\_cache()} before every forward pass, which severely degrades performance.Instead, we take a more efficient approach by analyzing PyTorch’s memory allocation mechanism. Fragmentation often results from the lack of a sufficiently large and contiguous cached memory block for new tensors, prompting new allocations and worsening fragmentation.
To address this, we introduce a memory pre-allocation strategy: we compute the maximum sequence length across the entire dataset before training, and pad all samples to this maximum length in the first step. This ensures that tensors can reuse pre-allocated memory blocks, reducing fragmentation and maintaining throughput.

\subsubsection{Data Pre-Processing}
We adopt native Megatron-Energon~\cite{megatron-energon} for unified data loading, eliminating the need for external training frameworks. Additionally, we optimized the preprocessing pipeline to reduce time consumption by up to 90\%.

We evaluated and compared two preprocessing strategies:

\begin{itemize}
    \item \textbf{Preprocessing Both JSON and Images.} Using the default Megatron-Energon data pipeline, both JSON metadata and images are compressed into binary files for WebDataset. However, this approach suffers from two major issues: (1) Low efficiency: Preprocessing 320,000 samples can take over 2 hours. (2) Inconsistent image readers: Megatron-Energon uses \texttt{cv2}, while models such as RoboBrain 2.0 use \texttt{PIL}, introducing subtle differences that may affect training performance.
    
    \item \textbf{Preprocessing JSON Only (Recommended).} In our optimized pipeline, only JSON files are preprocessed, and images are kept in their original form. Image preprocessing is deferred to the TaskEncoder module using the same preprocessor as Qwen2.5-VL. (1) High efficiency: Preprocessing 320,000 samples takes less than 10 minutes. (2) Alignment with model input: Ensures image handling is fully aligned between preprocessing and training, eliminating inconsistency and improving model performance.
\end{itemize}

\subsubsection{Distributed Data Loading}
To minimize the I/O burden on compute nodes, we reduce redundant data loading in large-scale distributed training. Unlike single-node setups, GPUs in distributed training systems play different roles depending on the chosen parallel strategy.
Data loading typically occurs along the data parallel (DP) dimension, where each DP rank handles a unique data shard. However, in multi-dimensional hybrid parallelism (\eg, DP-PP-TP), only a subset of GPU processes actually need to load data:
(1) In each Pipeline Parallel (PP)~\cite{megatron-pp} group, only the first and last stages need to perform data loading.
(2) Within Tensor Parallel (TP)~\cite{megatron-tp} groups, only one GPU per group is required to load data, with others receiving data via broadcast.
This design significantly reduces redundant I/O operations and improves overall data throughput.

\subsubsection{Fault Tolerance}
To handle both hardware and software failures during training, we co-designed fault-tolerant mechanisms between our FlagScale~\cite{flagscale} training framework and the system platform. Common errors, such as \texttt{LostCard}, \texttt{KubeNodeNotReady}, are automatically detected and trigger automatic job recovery and restart, ensuring minimal disruption.
Furthermore, our custom DataLoader module based on Megatron-Energon supports full data state recovery, allowing seamless resumption from the most recent checkpoint with complete consistency in data loading and sample shuffling states.

\subsection{Reinforcement Fine-Tuning Infrastructure}

We employ Reinforcement Learning with Verifiable Rewards (RLVR) to enhance RoboBrain 2.0 using VeRL~\cite{verl}, an open-source RL framework specifically designed for post-training LLMs and VLMs. Based on the HybridFlow architecture~\cite{sheng2025hybridflow}, VeRL features a hybrid-controller model that integrates both a global controller for inter-RL-role dataflow coordination and distributed controllers for intra-RL-role parallel processing. This architecture enables efficient execution of complex post-training workflows while ensuring scalability.
VeRL's support for multiple RL algorithms (\eg, GRPO) and seamless LLM integration makes it particularly suitable for RoboBrain 2.0's reinforcement fine-tuning (RFT) requirements. The framework enables high-performance model tuning with minimal overhead through its optimized dataflow management and parallel processing capabilities. Its efficient handling of large-scale training tasks and rigorous reward verification establishes VeRL as an ideal platform for advancing RoboBrain 2.0's capabilities via RLVR.

\subsection{Inference Infrastructure}
To improve the efficiency of model inference, we adopt FlagScale~\cite{flagscale}, also a multi-backend inference framework, which can automatically search for the optimal inference engine and configuration parameters based on the performance characteristics of different models on heterogeneous hardware accelerators, thereby effectively reducing inference latency.
Given the high sensitivity of embodied AI models to accuracy, we further introduce a mixed-bit quantization strategy~\cite{mixed-precesion,haq}. This strategy enhances inference efficiency and resource utilization while maintaining model performance. Specifically, the vision encoder retains full-precision floating-point computation to ensure the accuracy of key feature extraction. In contrast, during the language module, weights are quantized to 8-bit integers, while activations are preserved in 16-bit floating-point format. This mixed-precision approach significantly reduces computational overhead and memory usage with negligible impact on model accuracy.
Moreover, the quantization process is minimally invasive to existing inference pipelines and can be flexibly integrated into current systems. In end-to-end embodied tasks, weight-only quantization alone achieves approximately a 30\% reduction in inference latency, demonstrating the effectiveness and practicality of the proposed method in real-world deployment scenarios.

\newcommand{\gemini}{Gemini 2.5 Pro}
\newcommand{\openaio}{OpenAI o1}
\newcommand{\claude}{Claude 3.7 Sonnet}
\newcommand{\openaigpt}{OpenAI GPT-4o}
\newcommand{\qwen}{Qwen 2.5-VL 72B}
\newcommand{\openaicua}{OpenAI CUA}

\section{Evaluation Results}
\label{sec:evaluation}

We conducted a comprehensive evaluation of RoboBrain-2.0, focusing on its performance across spatial and temporal reasoning capabilities on embodiment. To ensure consistency and rigor in evaluation, we adopted FlagEvalMM~\cite{FlagEvalMM}, our flexible framework for systematic multimodal model assessment. Evaluations on spatial reasoning benchmarks (\eg, CV-bench~\cite{tong2024cambrian}, Blink~\cite{fu2024blink}, Where2Place~\cite{yuan2024robopointvisionlanguagemodelspatial}, ShareRobot-Bench~\cite{ji2025robobrain}), presented in \Cref{subsec:spatial_capability}, underscore the model's strengths in embodied spatial reasoning. An in-depth analysis of multi-robot collaboration~\cite{tan2025roboos} and long-horizon planning (\eg, EgoPlan2~\cite{chen2024egoplanbenchbenchmarkingmultimodallarge}, RoboBench) capabilities is provided in \Cref{subsec:temporal_capability}, highlighting the model's advancements in temporal reasoning tasks. Qualitative examples and prompt details are provided in \Cref{sec:app:qualitative} and \Cref{sec:app:prompt_detail}, respectively.

\subsection{Spatial Reasoning Capability}
\label{subsec:spatial_capability}

RoboBrain-32B-2.0 and RoboBrain-7B-2.0 demonstrate exceptional performance across nine spatial reasoning benchmarks: \textbf{BLINK}, \textbf{CV-Bench}, \textbf{EmbSpatial}, \textbf{RoboSpatial}, and \textbf{RefSpatial-Bench} (Table~\ref{tab:Synthetic}), as well as \textbf{SAT}, \textbf{VSI-Bench}, \textbf{Where2Place}, and \textbf{ShareRobot-Bench} (Table~\ref{tab:Spatial}).  Below is a detailed analysis highlighting their state-of-the-art (SOTA) achievements and near-SOTA competitive results.

\begin{table*}[!ht]
    \centering
    \caption{\textbf{Performance across five spatial reasoning benchmarks.} The best results among different models are highlighted in \textbf{bold}, while the second-best results are \underline{underlined}.}
    \vspace{-0.5em}
    \resizebox{0.95\textwidth}{!}{
    \begin{tabular}{l|ccc|c|c|c|ccc}
        \toprule
        \multicolumn{1}{l|}{\multirow{2}{*}{\textbf{Models / Metrics}}}  & \multicolumn{3}{c|}{\textbf{BLINK}}  &  \textbf{CV-Bench} &  \textbf{EmbSpatial} &  \textbf{RoboSpatial} &  \multicolumn{3}{c}{\textbf{RefSpatial-Bench}}\\ 
         \cmidrule(lr){2-4} \cmidrule(lr){5-5} \cmidrule(lr){6-6} \cmidrule(lr){7-7} \cmidrule(lr){8-10}
        & \textbf{Dep.} & \textbf{Spa.} & \textbf{All $\uparrow$} & \textbf{All $\uparrow$} & \textbf{All $\uparrow$} & \textbf{All $\uparrow$} & \textbf{Loc.} & \textbf{Pla.} & \textbf{All $\uparrow$} \\ 
        \midrule
        \rowcolor[HTML]{F2F2F2} \multicolumn{10}{l}{\textbf{General Baselines}} \\ \midrule
        Gemini-2.5-Pro-preview-05-06~\cite{gemini25pro} & 79.03 & 84.62 & 81.83 & 84.59 & \textbf{78.74} & \underline{59.87} & \underline{44.58} & \underline{31.73} & \underline{38.16}\\ 
        Gemini-2.5-Flash-preview-04-17~\cite{gemini25pro} & 77.42  & 79.02 & 78.22 & 84.03  & 74.75  & 54.10  & 37.50  & 23.00 & 30.25 \\
        GPT-o4-mini-2025-05-16~\cite{gpto3-o4-mini} & 79.03 & \textbf{88.11} & 83.57 & \underline{85.21} & 78.29 & 51.25 & 15.00 & 19.58 & 17.29\\ 
        GPT-4o-2024-11-20~\cite{gpt4o} & 72.58 & 83.22 & 77.90  & 78.63 & 71.92 & 44.42 & 8.00 & 9.55 & 8.78\\ 
        Claude-Sonnet-4-2025-05-14~\cite{claude4} & 75.81 & 80.42 & 78.12  & 78.43 & 64.26 & 51.26 & 5.00 & 10.37 & 7.69\\
        Qwen2.5-VL-32B-Instruct~\cite{qwen2.5-vl} & 77.42 & 85.31 & 81.37 & 81.59 & 74.45 & 52.16 & 16.83 & 10.60 & 13.72\\ 
        Qwen2.5-VL-72B-Instruct~\cite{qwen2.5-vl} & 74.19 & 78.32 & 76.26 & 82.68 & 73.30 & 48.33 & 23.50 & 15.83 & 19.67 \\ 
        \midrule
        \rowcolor[HTML]{F2F2F2} \multicolumn{10}{l}{\textbf{Embodied Baselines}} \\ \midrule
        Cosmos-Reason1-7B~\cite{cosmos-reason1}  & 63.71 & 73.43 & 68.57 & 74.71 & 65.22 & 38.81 & 9.84 & 1.04 & 5.44 \\ 
        VeBrain-8B~\cite{luo2025vebrain}  & 78.23 & 81.12 & 79.68 & 78.57 & 70.52 & 42.48 & 0.03 & 0.57 & 0.30 \\ 
        Magma-8B~\cite{magma} & 65.32  & 66.43 & 65.88 & 60.98  & 64.59  & 33.71  & 1.00  & 8.00 & 4.50 \\
        RoboBrain-7B-1.0~\cite{ji2025robobrain}  & 75.81 & 78.32 & 77.07 & 76.22 & 68.13 & 51.53 & 14.43 & 5.41 & 9.92 \\ 
        \rowcolor[HTML]{DAEFF9} RoboBrain-7B-2.0  & \textbf{84.68} & 83.22 & \textbf{83.95} & \textbf{85.75}& 76.32 & 54.23 & 36.00 & 29.00 & 32.50 \\ 
        \rowcolor[HTML]{DAEFF9} RoboBrain-32B-2.0 & \underline{79.84} & \underline{87.41} & \underline{83.63} & 83.92 & \underline{78.57} & \textbf{72.43} & \textbf{54.00} & \textbf{54.00} & \textbf{54.00} \\ 
        \bottomrule
    \end{tabular}
    }
    \label{tab:Synthetic}
\end{table*}

\begin{table*}[ht]
    \centering
    \caption{\textbf{Performance across four spatial reasoning benchmarks.} The best results among different models are highlighted in \textbf{bold}, while the second-best results are \underline{underlined}.}
    \vspace{-0.5em}
    \resizebox{0.93\textwidth}{!}{
    \begin{tabular}{l|c|c|ccc|cc}
        \toprule
        \multicolumn{1}{l|}{\multirow{2}{*}{\textbf{Models / Metrics}}}  & \multicolumn{1}{c|}{\textbf{~~~~~~SAT~~~~~~}} & \multicolumn{1}{c|}{\textbf{VSI-Bench}} & \multicolumn{3}{c|}{\textbf{Where2Place$^{*}$}} &  \multicolumn{2}{c}{\textbf{ShareRobot-Bench}} \\ 
         \cmidrule(lr){2-2} \cmidrule(lr){3-3} \cmidrule(lr){4-6} \cmidrule(lr){7-8}
        & \textbf{All $\uparrow$} & \textbf{All $\uparrow$} & \textbf{Seen} & \textbf{Unseen} & \textbf{All $\uparrow$} & \textbf{Afford. $\uparrow$} & \textbf{Traj. (DFD $\downarrow$ )} \\ 
        \midrule
        \rowcolor[HTML]{F2F2F2} \multicolumn{8}{l}{\textbf{General Baselines}} \\ \midrule
        Gemini-2.5-Pro-preview-05-06~\cite{gemini25pro} & 79.33 & \underline{47.81} & 42.92 & 41.13 & 42.38 & 10.26 & 0.7666 \\ 
        Gemini-2.5-Flash-preview-04-17~\cite{gemini25pro} & 74.00  & \textbf{48.83} & 31.54 & 21.73 & 28.60  & 2.50  & 0.9087   \\
        GPT-o4-mini-2025-05-16~\cite{gpto3-o4-mini} & \underline{82.00} & 41.96 & 26.63 & 26.49 & 26.59 & 8.27 & 0.5726 \\ 
        GPT-4o-2024-11-20~\cite{gpt4o} & 66.67 & 43.60 & 20.28 & 20.71 & 20.41 & 6.00 & 0.6850 \\ 
        Claude-Sonnet-4-2025-05-14~\cite{claude4} & 75.33 & 47.02 & 21.56 & 35.11 & 25.63 & 8.00 & 0.7591 \\ 
        Qwen2.5-VL-32B-Instruct~\cite{qwen2.5-vl} & 80.00 & 36.07 & 18.22 & 32.55 & 22.52 & 11.97 & 0.9222 \\ 
        Qwen2.5-VL-72B-Instruct~\cite{qwen2.5-vl} & 58.67 & 35.51 & 35.74 & 49.65 & 39.92 & 23.80 & \underline{0.5034} \\ 
        \midrule
        \rowcolor[HTML]{F2F2F2} \multicolumn{8}{l}{\textbf{Embodied Baselines}} \\ \midrule
        Cosmos-Reason1-7B~\cite{cosmos-reason1}  & 60.67 & 25.64 & 5.07 & 6.53 & 5.51 & 9.98 & 0.8524  \\ 
        VeBrain-8B~\cite{luo2025vebrain}  & 58.00 & 26.30 & 12.27 & 9.17 & 11.34 & 3.66 & 1.1659 \\ 
        Magma-8B~\cite{magma} & 71.33  & 12.65 & 9.93 & 13.14 & 10.89  & --  & 0.7478 \\
        RoboBrain-7B-1.0~\cite{ji2025robobrain}  & 59.33 & 31.12 & 54.58 & 49.45 & 53.04 & 10.20 & 0.6248 \\ 
        \rowcolor[HTML]{DAEFF9} RoboBrain-7B-2.0  & 75.33 & 36.10 & \underline{64.33} & \underline{61.88}  & \underline{63.59} & \underline{28.05} & 0.5512 \\ 
        \rowcolor[HTML]{DAEFF9} RoboBrain-32B-2.0 & \textbf{86.67} &  42.69 & \textbf{73.95} & \textbf{72.74} & \textbf{73.59} & \textbf{35.28} & \textbf{0.2368} \\ 
        \bottomrule
    \end{tabular}
    }
    \label{tab:Spatial}
\vspace{-0.5em}
\end{table*}

\begin{itemize}

\item{\textbf{BLINK.}}
In the BLINK~\cite{fu2024blink} benchmark, models are evaluated on depth perception (Dep.) and spatial relation understanding (Spa.). RoboBrain-7B-2.0 achieves a SOTA average score of \textbf{83.95} (Dep.: \textbf{84.68}, Spa.: 83.22), outperforming all general baselines, including GPT-o4-mini-2025-05-16 (83.57), Gemini-2.5-Pro-preview-05-06 (81.83), Qwen2.5-VL-32B-Instruct (81.37), Claude-Sonnet-4-2025-05-14 (78.12), GPT-4o-2024-11-20 (77.90), and Qwen2.5-VL-72B-Instruct (76.26), as well as embodied baselines like VeBrain-8B (79.68) and Cosmos-Reason1-7B (68.57). RoboBrain-32B-2.0 follows closely with an average of 83.63 (Dep.: 79.84, Spa.: \textbf{87.41}), surpassing all general and embodied baselines except RoboBrain-7B-2.0, demonstrating strong spatial reasoning capabilities.

\vspace{+1mm}

\item{\textbf{CV-Bench.}}
The CV-Bench~\cite{tong2024cambrian} benchmark assesses a model’s accuracy in 2D/3D spatial understanding and visual processing. RoboBrain-7B-2.0 secures a SOTA accuracy of \textbf{85.75}, slightly ahead of RoboBrain-32B-2.0 (83.92), both outperforming all general baselines, including GPT-o4-mini-2025-05-16 (85.21), Gemini-2.5-Pro-preview-05-06 (84.59), Qwen2.5-VL-72B-Instruct (82.68), Qwen2.5-VL-32B-Instruct (81.59), GPT-4o-2024-11-20 (78.63), and Claude-Sonnet-4-2025-05-14 (78.43), as well as embodied baselines like VeBrain-8B (78.57) and Cosmos-Reason1-7B (74.71).

\vspace{+1mm}

\item{\textbf{EmbSpatial.}}
The EmbSpatial~\cite{du2024embspatial} benchmark evaluates models on embodied spatial tasks. RoboBrain-32B-2.0 achieves a near SOTA accuracy of \textbf{78.57}, slightly less than Gemini-2.5-Pro-preview-05-06 (78.74) and surpassing all other general baselines, including GPT-o4-mini-2025-05-16 (78.29), Qwen2.5-VL-32B-Instruct (74.45), Qwen2.5-VL-72B-Instruct (73.30), GPT-4o-2024-11-20 (71.92), and Claude-Sonnet-4-2025-05-14 (64.26). RoboBrain-7B-2.0 follows with a competitive score of 76.32, outperforming most general baselines and all embodied baselines, indicating strong embodied spatial reasoning.

\vspace{+1mm}

\item{\textbf{RoboSpatial.}}
The RoboSpatial~\cite{song2025robospatial} benchmark measures spatial reasoning in robot environments, such as object localization and manipulation. RoboBrain-32B-2.0 achieves a clear SOTA score of \textbf{72.43}, substantially ahead of general baselines like Gemini-2.5-Pro-preview-05-06 (59.87), Qwen2.5-VL-72B-Instruct (48.33), GPT-o4-mini-2025-05-16 (51.25), and Claude-Sonnet-4-2025-05-14 (51.26). RoboBrain-7B-2.0 scores 54.23, outperforming all general baselines except RoboBrain-32B-2.0, demonstrating significant improvements in spatial reasoning for robotic tasks.

\vspace{+1mm}

\item{\textbf{RefSpatial-Bench.}}
The RefSpatial-Bench~\cite{zhou2025roborefer} benchmark evaluates models on spatial referring expressions, requiring precise point predictions under spatial constraints, with metrics for Location (Loc.) and Placement (Pla.) accuracy. RoboBrain-32B-2.0 achieves SOTA scores of \textbf{54.00} (Loc.) and \textbf{54.00} (Pla.), significantly outperforming all general baselines, including Gemini-2.5-Pro-preview-05-06 (44.58, 31.73), Qwen2.5-VL-72B-Instruct (23.50, 15.83), Qwen2.5-VL-32B-Instruct (16.83, 10.60), GPT-o4-mini-2025-05-16 (15.00, 19.58), GPT-4o-2024-11-20 (8.00, 9.55), and Claude-Sonnet-4-2025-05-14 (5.00, 10.37). RoboBrain-7B-2.0 scores 36.00 (Loc.) and 29.00 (Pla.), outperforming all general baselines except RoboBrain-32B-2.0, showing competitive precision in complex spatial referring tasks.

\vspace{+1mm}

\item{\textbf{SAT.}}
The SAT~\cite{ray2024sat} benchmark measures general spatial reasoning abilities across various scenes and tasks. RoboBrain-32B-2.0 achieves a clear SOTA score of \textbf{86.67}, significantly outperforming all general baselines, including GPT-o4-mini-2025-05-16 (82.00), Gemini-2.5-Pro-preview-05-06 (79.33), Qwen2.5-VL-72B-Instruct (58.67), and Claude-Sonnet-4-2025-05-14 (75.33). RoboBrain-7B-2.0 achieves 75.33, surpassing most general and embodied baselines, showcasing its strong spatial reasoning capability.

\vspace{+1mm}

\let\thefootnote\relax\footnotetext{$^{*}$Upon inspection, we found that the test set included several incorrect cases, which were manually screened and excluded.}

\item{\textbf{VSI-Bench.}}
The VSI-Bench~\cite{yang2025vsibench} evaluates visual-spatial integration capabilities. Gemini-2.5-Flash-preview-04-17 achieves the best performance with \textbf{48.83}. RoboBrain-32B-2.0 achieves 42.69, outperforming most general and embodied baselines, including GPT-o4-mini-2025-05-16 (41.96) and Qwen2.5-VL-72B-Instruct (35.51). RoboBrain-7B-2.0 reaches 36.10, indicating solid visual-spatial integration skills.

\vspace{+1mm}

\item{\textbf{Where2Place.}}
The Where2Place~\cite{yuan2024robopointvisionlanguagemodelspatial} benchmark measures a model’s ability to predict object placements in both seen and unseen scenarios under spatial constraints. RoboBrain-32B-2.0 achieves a SOTA average of \textbf{73.59} (Seen: \textbf{73.95}, Unseen: \textbf{72.74}), substantially surpassing all general and embodied baselines, including Qwen2.5-VL-72B-Instruct (39.92), Gemini-2.5-Pro-preview-05-06 (42.38), Claude-Sonnet-4-2025-05-14 (25.63), and VeBrain-8B (11.34). RoboBrain-7B-2.0 also performs strongly with an average of 63.59 (Seen: 64.33, Unseen: 61.88), outperforming all baselines except RoboBrain-32B-2.0.

\vspace{+1mm}

\item{\textbf{ShareRobot-Bench-Affordance.}}
The ShareRobot Affordance task~\cite{ji2025robobrain} evaluates models on object functionality and interaction understanding. RoboBrain-32B-2.0 secures a SOTA performance with an accuracy of \textbf{35.28}, ahead of all general baselines, including Qwen2.5-VL-72B-Instruct (23.80), Qwen2.5-VL-32B-Instruct (11.97), GPT-4o-2024-11-20 (6.00), and Claude-Sonnet-4-2025-05-14 (8.00). RoboBrain-7B-2.0 achieves 28.05, outperforming all general and embodied baselines except RoboBrain-32B-2.0.

\vspace{+1mm}

\item{\textbf{ShareRobot-Bench-Trajectory.}}
The ShareRobot Trajectory task~\cite{ji2025robobrain} assesses navigation and motion prediction, using Dynamic Fréchet Distance (DFD), where lower values denote better performance. RoboBrain-32B-2.0 achieves a SOTA DFD of \textbf{0.2368}, outperforming all general and embodied baselines, including Qwen2.5-VL-72B-Instruct (0.5034), GPT-o4-mini-2025-05-16 (0.5726), and Gemini-2.5-Pro-preview-05-06 (0.7666). RoboBrain-7B-2.0 follows with a competitive DFD of 0.5512, demonstrating strong path-planning capabilities.

\end{itemize}

\subsection{Temporal Reasoning Capability}
\label{subsec:temporal_capability}

RoboBrain-32B-2.0 and RoboBrain-7B-2.0 exhibit outstanding performance across three critical measures of temporal reasoning benchmarks: \textbf{Multi-Robot Planning}, \textbf{Ego-Plan2}, and \textbf{RoboBench}, as shown in Table~\ref{tab:Temporal}. Below is a detailed analysis highlighting their state-of-the-art (SOTA) achievements and near-SOTA results.

\begin{table*}[!ht]
    \centering
    \caption{\textbf{Performance across three temporal reasoning benchmarks.} The best results among different models are highlighted in \textbf{bold}, while the second-best results are \underline{underlined}.}
    \vspace{-0.5em}
    \resizebox{0.95\textwidth}{!}{
    \begin{tabular}{l|cccc|ccccc|c}
        \toprule
        \multicolumn{1}{l|}{\multirow{2}{*}{\textbf{Models / Metrics}}} & \multicolumn{4}{c|}{\textbf{Multi-Robot Planning}} & \multicolumn{5}{c|}{\textbf{Ego-Plan2}} & \multicolumn{1}{c}{\textbf{RoboBench}}\\ 
        \cmidrule(lr){2-5} \cmidrule(lr){6-10} \cmidrule(lr){11-11}
        & \textbf{Super.} & \textbf{Rest.} & \textbf{House.} & \textbf{All $\uparrow$} & \textbf{Daily.} & \textbf{Hobbies.} & \textbf{Rec.} & \textbf{Work.} & \textbf{All $\uparrow$} & \textbf{Plan. $\uparrow$}\\ 
        \midrule
        \rowcolor[HTML]{F2F2F2} \multicolumn{11}{l}{\textbf{General Baselines}} \\ \midrule
        Gemini-2.5-Pro-preview-05-06~\cite{gemini25pro} & 63.51 & 54.77 & 78.39 & 65.39 & 44.19 & 43.05 & 46.45 & 39.60 & 42.85 & 63.49 \\ 
        Gemini-2.5-Flash-preview-04-17~\cite{gemini25pro} & 59.44 & 55.78 & 76.88 & 63.86 & 38.72 & 35.59  & 43.72 & 33.42 & 37.09 & 69.33 \\
        GPT-o4-mini-2025-05-16~\cite{gpto3-o4-mini} & 63.32 & 55.28 & 78.89 & 65.50 & 47.61 & 35.93 & 42.62 & 37.13 & 41.11 & 70.01 \\ 
        GPT-4o-2024-11-20~\cite{gpt4o} & 77.89 & 67.34 & 79.40 & 74.50 & 47.38 & 40.00 & 44.81 & 35.64 & 41.79 & 68.60\\ 
        Claude-Sonnet-4-2025-05-14~\cite{claude4} & 73.08 & 61.81 & 80.40 & 71.30 & 43.51 & 41.02 & 42.62 & 38.87 & 41.26 & \underline{70.21} \\ 
        Qwen2.5-VL-32B-Instruct~\cite{qwen2.5-vl} & 67.84 & 61.81 & 75.38 & 68.00 & \textbf{64.46} & 51.53 & \underline{57.92} & \underline{50.00} & \underline{56.25} & 45.92 \\ 
        Qwen2.5-VL-72B-Instruct~\cite{qwen2.5-vl} & 77.39 & 68.34 & 79.40 & 74.67 & 60.36 & \underline{48.14} & \textbf{63.39} & 46.29 & 53.75 & 66.94 \\ 
        \midrule
        \rowcolor[HTML]{F2F2F2} \multicolumn{11}{l}{\textbf{Embodied Baselines}} \\ \midrule
        Cosmos-Reason1-7B~\cite{cosmos-reason1} & 35.17 & 25.62 & 40.70 & 33.66 & 30.75 & 27.12 & 31.69 & 20.30 & 26.87 & 53.17\\ 
        VeBrain-8B~\cite{luo2025vebrain} & 41.70 & 35.67 & 39.69 & 38.83 & 31.79 & 35.31 & 31.19 & 34.43 & 27.30 & 46.77 \\ 
        Magma-8B~\cite{magma} & -- & -- & -- & -- & 4.56 & 3.39 & 6.56 & 2.97 & 4.09 & -- \\ 
        RoboBrain-7B-1.0~\cite{ji2025robobrain} & 4.52 & 7.04 & 5.03 & 5.50 & -- & -- & -- & -- & -- & 38.93 \\ 
        \rowcolor[HTML]{DAEFF9} RoboBrain-7B-2.0 & \underline{83.92} & \textbf{77.39} & \underline{84.42} & \textbf{81.50} & 39.41 & 32.20 & 33.88 & 26.98 & 33.23 & \textbf{72.16} \\ 
        \rowcolor[HTML]{DAEFF9} RoboBrain-32B-2.0 & \textbf{84.42} &  \underline{72.36} &  \textbf{85.43} & \underline{80.33} & \underline{64.01} & \textbf{53.22} & \underline{57.92} & \textbf{52.48} & \textbf{57.23} & 68.33 \\ 
        \bottomrule
    \end{tabular}
    }
    \label{tab:Temporal}
\end{table*}

\vspace{+1mm}

\begin{itemize}

\item{\textbf{Multi-Robot Planning.}}\
In the Multi-Robot Planning task~\cite{tan2025roboos}, models are evaluated on their ability to coordinate multiple robots across different scenarios: Super (Supermarket), Rest (Restaurant), and House (Household). RoboBrain-32B-2.0 achieves a SOTA average score of \textbf{80.33} (Super: \textbf{84.42}, Rest: 72.36, House: \textbf{85.43}), significantly outperforming all general baselines, including GPT-4o-2024-11-20 (74.50), Qwen2.5-VL-72B-Instruct (74.67), Claude-Sonnet-4-2025-05-14 (71.30), Gemini-2.5-Pro-preview-05-06 (65.39), and Qwen2.5-VL-32B-Instruct (68.00). It also surpasses the embodied baseline RoboBrain-7B-2.0 (81.50). RoboBrain-7B-2.0 follows closely with an average of 81.50 (Super: 83.92, Rest: 77.39, House: 84.42), outperforming all general baselines and matching the performance of RoboBrain-7B-1.5-OS in Rest and House scenarios.
\vspace{+1mm}

\item{\textbf{Ego-Plan2.}}\
The Ego-Plan2~\cite{chen2024egoplanbenchbenchmarkingmultimodallarge} benchmark assesses a model’s capability to plan daily activities across four categories: Daily (Daily Routines), Hobbies, Rec (Recreation), and Work. RoboBrain-32B-2.0 secures a SOTA average score of \textbf{57.23} (Daily: \textbf{64.01}, Hobbies: \textbf{53.22}, Rec: \textbf{57.92}, Work: \textbf{52.48}), significantly outperforming all general and embodied baselines, including Qwen2.5-VL-32B-Instruct (56.25), Qwen2.5-VL-72B-Instruct (53.75), Gemini-2.5-Pro-preview-05-06 (42.85), GPT-4o-2024-11-20 (41.79), Claude-Sonnet-4-2025-05-14 (41.26), GPT-o4-mini-2025-05-16 (41.11), VeBrain-8B (27.30), and Cosmos-Reason1-7B (26.87). In contrast, RoboBrain-7B-2.0 achieves an average of 33.23 (Daily: 39.41, Hobbies: 32.20, Rec: 33.88, Work: 26.98), which is lower than general baselines like Qwen2.5-VL-32B-Instruct and Qwen2.5-VL-72B-Instruct but surpasses embodied baselines such as VeBrain-8B and Cosmos-Reason1-7B.
\vspace{+1mm}

\item{\textbf{RoboBench.}}\
The RoboBench Benchmark (Planning part) evaluates a model's ability to plan robotic mobile manipulation tasks according to their pre-defined skills across three categories: cross-embodiment, cross-object, and cross-view. On this benchmark, RoboBrain-7B-2.0 achieves a state-of-the-art (SOTA) score of \textbf{72.16}, surpassing all general and embodied baselines, including Claude-Sonnet-4-2025-05-14 (70.21), GPT-o4-mini-2025-05-16 (70.01). The performance of RoboBrain-32B-2.0, with a score of 68.33, outperforming several general baselines like GPT-4o-2024-11-20 (68.60) and Qwen2.5-VL-72B-Instruct (66.94), as well as other embodied baselines such as Cosmos-Reason1-7B (53.17) and VeBrain-8B (46.77).

\end{itemize}
\section{Conclusion and Future Works}
\label{sec:conclusion}

In this report, we introduced RoboBrain 2.0, our latest generation of embodied vision-language foundation models, developed to support unified perception, reasoning, and planning in complex physical environments. Built on a modular architecture with a dedicated vision encoder and a decoder-only language model, RoboBrain 2.0 enables high-resolution image and video comprehension, as well as spatial and temporal reasoning. Through a progressive three-stage training strategy—encompassing foundational spatiotemporal learning, embodied enhancement, and chain-of-thought reasoning—the model demonstrates strong generalization across a wide variety of challenging embodied tasks. Despite its compact size, RoboBrain 2.0 achieves state-of-the-art results on most of public embodied spatial and temporal reasoning benchmarks, outperforming both open-source and proprietary models in spatial understanding, closed-loop interaction, and long-horizon planning. Its capabilities span a broad spectrum of embodied scenarios, including affordance prediction, spatial referring, trajectory forecasting, multi-agent coordination, and scene graph construction and updating.

\vspace{0.3em}

We regard RoboBrain 2.0 as a solid foundation toward developing more general embodied AI, emphasizing the importance of tightly integrated perception, reasoning, and planning. Moving forward, we plan to expand RoboBrain 2.0 along two key directions:

\begin{itemize}

\item \textbf{Embodied VLM-powered VLA:} We aim to integrate cutting-edge embodied VLMs into the Vision-Language-Action (VLA) framework. By harnessing the powerful spatiotemporal perception and high-level reasoning capabilities of VLMs, this direction seeks to substantially enhance the generality and robustness of action generation. The resulting system will support more nuanced understanding and precise execution of complex, open-ended instructions in real-world scenarios.

\item \textbf{System-Level Integration:} To improve RoboBrain 2.0's practical utility, we will pursue tight integration with advanced robotics platforms and operating systems. This will enable serverless deployment, adaptation-free skill registration, and low-latency real-time control. In parallel, we envision building a collaborative embodied AI ecosystem—an ``intelligence app store''—that supports plug-and-play components for perception, reasoning, and control in real-world robotic systems.

\end{itemize}

\vspace{0.3em}

We release RoboBrain 2.0 at \url{https://superrobobrain.github.io}, including model checkpoints, training recipes, and evaluation tools, to support broader research and downstream applications in embodied AI. We hope this work bridges the gap between vision-language intelligence and real-world physical interaction.

\clearpage

\bibliographystyle{plainnat}
\bibliography{main}

\clearpage
\section{Contributions and Author List}
\label{sec:contributions}
\setlength{\parskip}{0pt}
\setlength{\itemsep}{0pt}
\setlength{\parsep}{0pt}
\begin{multicols}{2}

\subsubsection*{Core Contributors}
\textbf{Model Training}
\begin{itemize}
    \vspace{0.5em}\item Mingyu Cao$^*$
    \vspace{0.5em}\item Huajie Tan$^*$
    \vspace{0.5em}\item Yuheng Ji$^*$
    \vspace{0.5em}\item Xiansheng Chen$^*$$^\dagger$
    \vspace{0.5em}\item Minglan Lin$^*$$^\dagger$
    \vspace{0.5em}\item Zhiyu Li
    \vspace{0.5em}\item Zhou Cao
    \vspace{0.5em}\item Pengwei Wang$^\dagger$
\end{itemize}

\vspace{1em}

\textbf{Data \& Evaluation}
\begin{itemize}

    \vspace{0.5em}\item Enshen Zhou
    \vspace{0.5em}\item Yi Han
    \vspace{0.5em}\item Yingbo Tang
    \vspace{0.5em}\item Xiangqi Xu
    \vspace{0.5em}\item Wei Guo
    \vspace{0.5em}\item Yaoxu Lyu
    \vspace{0.5em}\item Yijie Xu
    \vspace{0.5em}\item Jiayu Shi
    \vspace{0.5em}\item Mengfei Du
    \vspace{0.5em}\item Cheng Chi$^\dagger$
    \vspace{0.5em}\item Mengdi Zhao
    \vspace{0.5em}\item Xiaoshuai Hao
\end{itemize}

\vspace{1em}
\textbf{Research Leads}
\begin{itemize} \setlength{\itemsep}{0pt} \setlength{\parsep}{0pt}
    \vspace{0.5em}\item Yonghua Lin
    \vspace{0.5em}\item Zhongyuan Wang$^{\text{\Letter}}$
    \vspace{0.5em}\item Tiejun Huang
    \vspace{0.5em}\item Shanghang Zhang$^{\text{\Letter}}$
\end{itemize}

\vspace{8em}
\subsubsection*{Contributors}
\vspace{-0.3em}

\textbf{Real-Robot Experiments}
\begin{itemize}
    \vspace{0.5em}\item Junkai Zhao
    \vspace{0.5em}\item Xiaojie Zhang
    \vspace{0.5em}\item Shanyu Rong
    \vspace{0.5em}\item Huaihai Lyu
    \vspace{0.5em}\item Zhengliang Cai
    \vspace{0.5em}\item Yankai Fu
    \vspace{0.5em}\item Ning Chen
    \vspace{0.5em}\item Bolun Zhang
    \vspace{0.5em}\item Lingfeng Zhang
    \vspace{0.5em}\item Shuyi Zhang
    \vspace{0.5em}\item Dong Liu    
\end{itemize}


\textbf{Product \& Operations}
\begin{itemize}
    \vspace{0.5em}\item Xi Feng
    \vspace{0.5em}\item Songjing Wang
    \vspace{0.5em}\item Xiaodan Liu
    \vspace{0.5em}\item Yance Jiao
\end{itemize}


\textbf{Infrastructure}
\begin{itemize}
    \vspace{0.5em}\item Mengsi Lyu
    \vspace{0.5em}\item Zhuo Chen
    \vspace{0.5em}\item Chenrui He
    \vspace{0.5em}\item Yupu Feng    
    \vspace{0.5em}\item Yulong Ao
\end{itemize}


\textbf{Evaluation}
\begin{itemize}
    \vspace{0.5em}\item Xue Sun
    \vspace{0.5em}\item Zheqi He
    \vspace{0.5em}\item Jingshu Zheng
    \vspace{0.5em}\item Xi Yang
\end{itemize}


\textbf{System Management}
\begin{itemize}
    \vspace{0.5em}\item Donghai Shi
    \vspace{0.5em}\item Kunchang Xie
    \vspace{0.5em}\item Bochao Zhang
    \vspace{0.5em}\item Shaokai Nie
    \vspace{0.5em}\item Chunlei Men
\end{itemize}

\end{multicols}

\let\thefootnote\relax\footnotetext{$^{*}$ Equal Contribution (Co-first Authors).}
\let\thefootnote\relax\footnotetext{$^{\dagger}$ Project Leaders.}
\let\thefootnote\relax\footnotetext{$^{\text{\Letter}}$ Corresponding Author. Team Email: \url{robobrain@baai.ac.cn}}
\clearpage

\beginappendix

\section{Qualitative examples}
This section provides a comprehensive set of qualitative examples that illustrate the capabilities of RoboBrain 2.0 in various embodied AI tasks. These examples demonstrate the model's proficiency in spatial reasoning, temporal planning, and interactive reasoning, showcasing its potential for real-world applications.

\label{sec:app:qualitative}

\subsection{Examples for Pointing}
\begin{figure*}[!ht]
    \centering
    \includegraphics[width=0.85\linewidth]{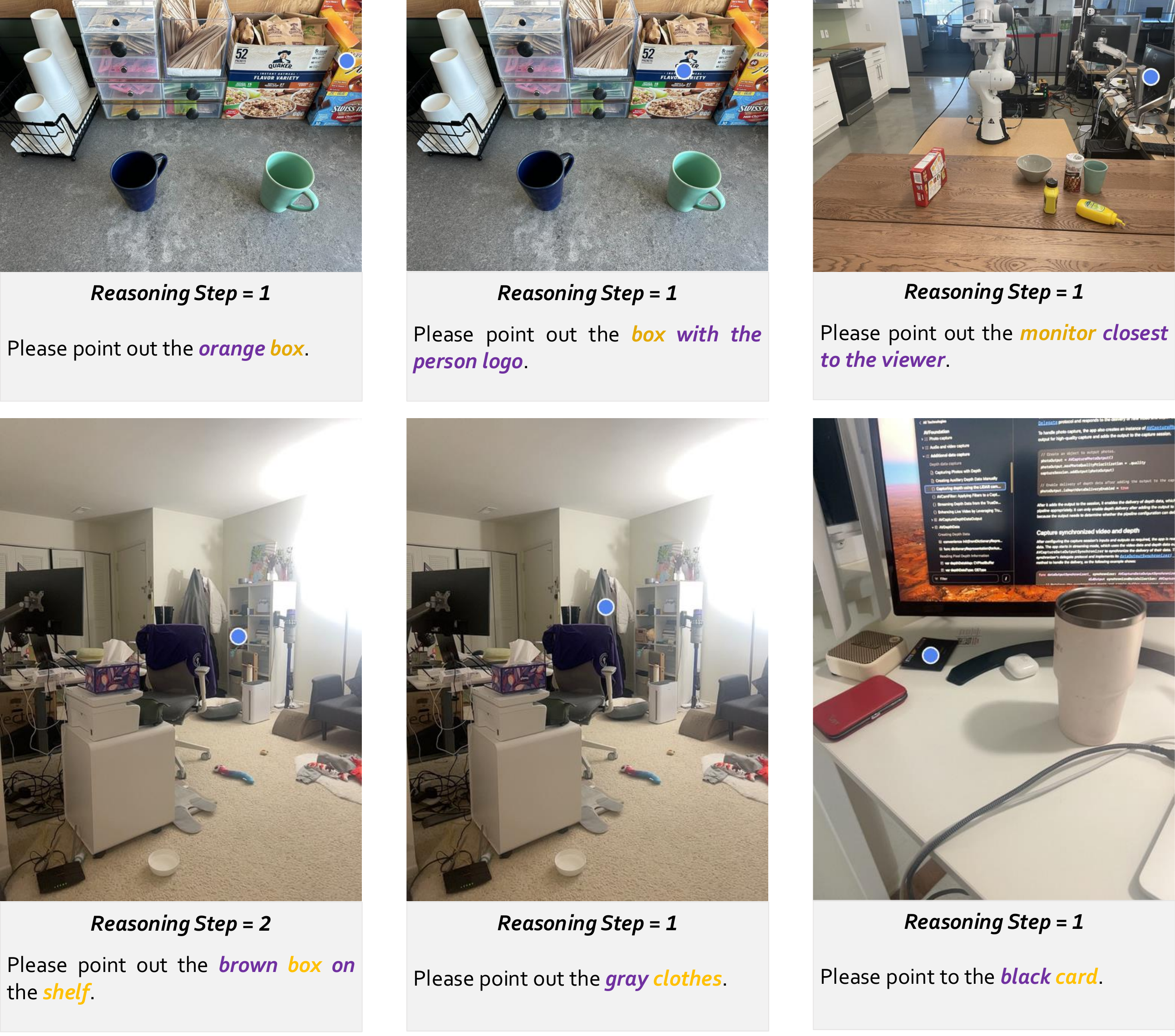}
    \caption{\textbf{Pointing Examples of RoboBrain 2.0}. The blue point represents the model's spatial referring prediction.}
    \label{fig:point_1}
    \vspace{-1.5em}
\end{figure*}

\begin{figure*}[!ht]
    \centering
    \includegraphics[width=0.9\linewidth]{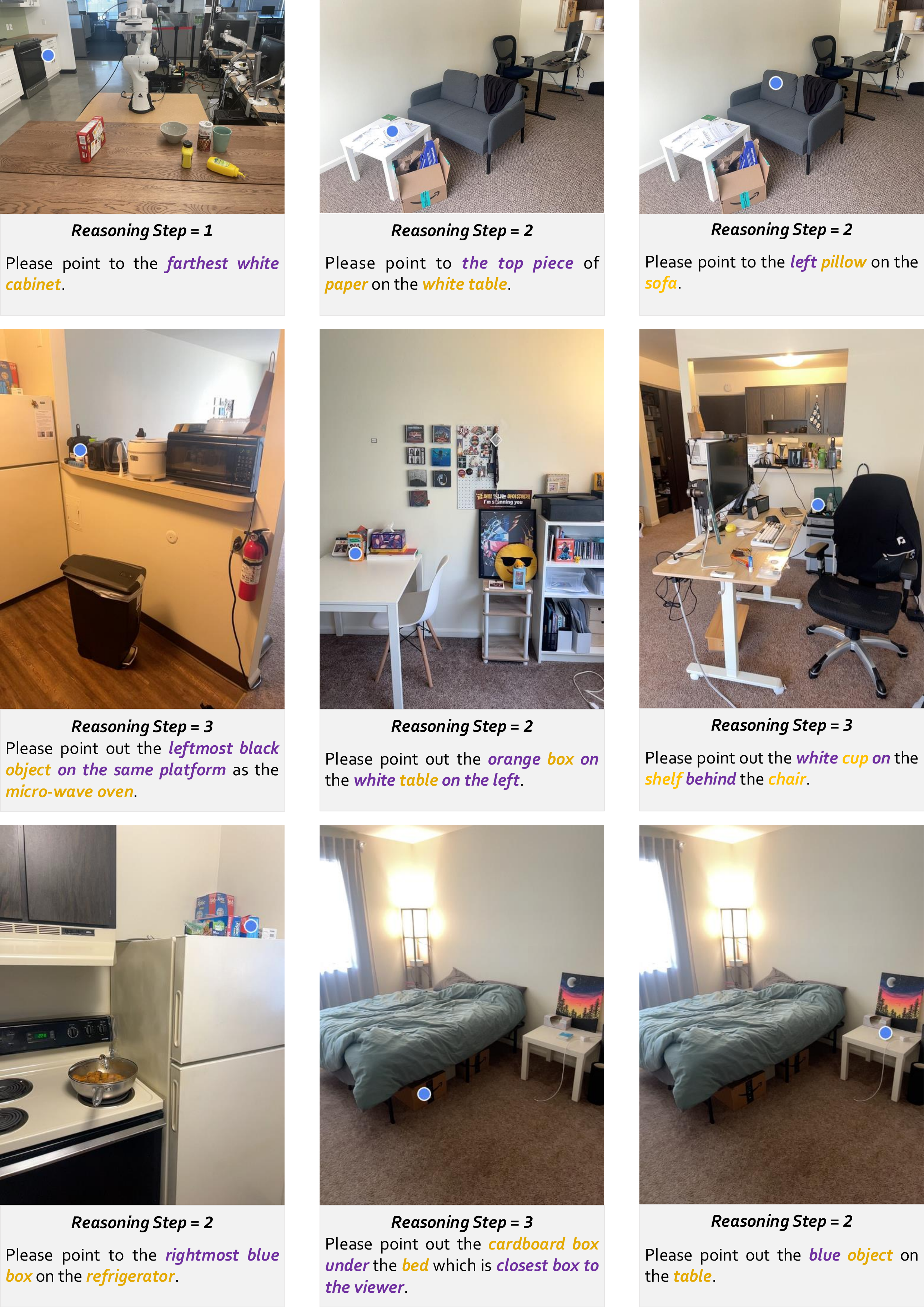}
    \caption{\textbf{Pointing Examples of RoboBrain 2.0}. The blue point represents the model's spatial referring prediction.}
    \label{fig:point_2}
    \vspace{-1.5em}
\end{figure*}

\begin{figure*}[!ht]
    \centering
    \includegraphics[width=0.9\linewidth]{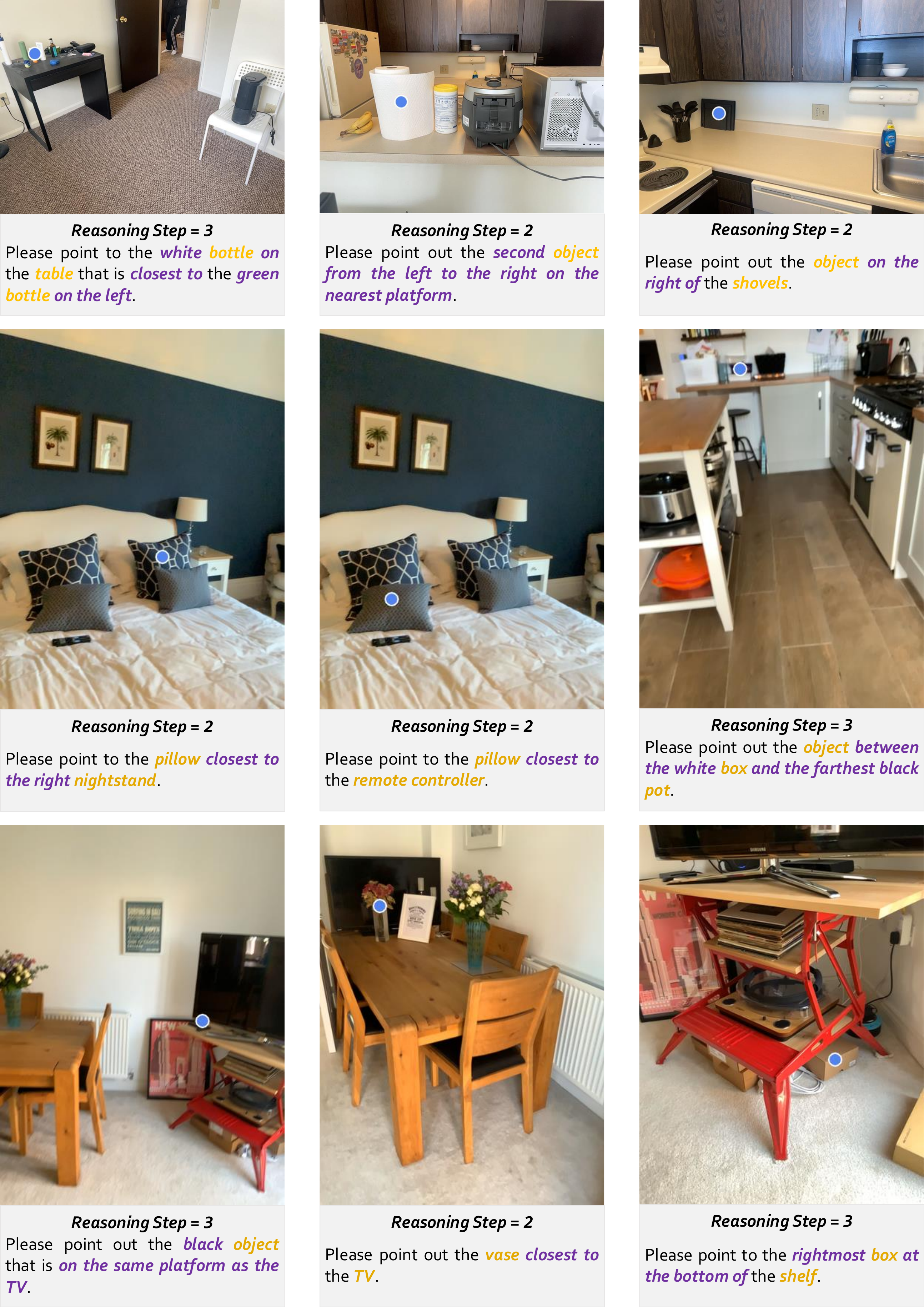}
    \caption{\textbf{Pointing Examples of RoboBrain 2.0}. The blue point represents the model's spatial referring prediction.}
    \label{fig:point_3}
    \vspace{-1.5em}
\end{figure*}

\begin{figure*}[!ht]
    \centering
    \includegraphics[width=0.9\linewidth]{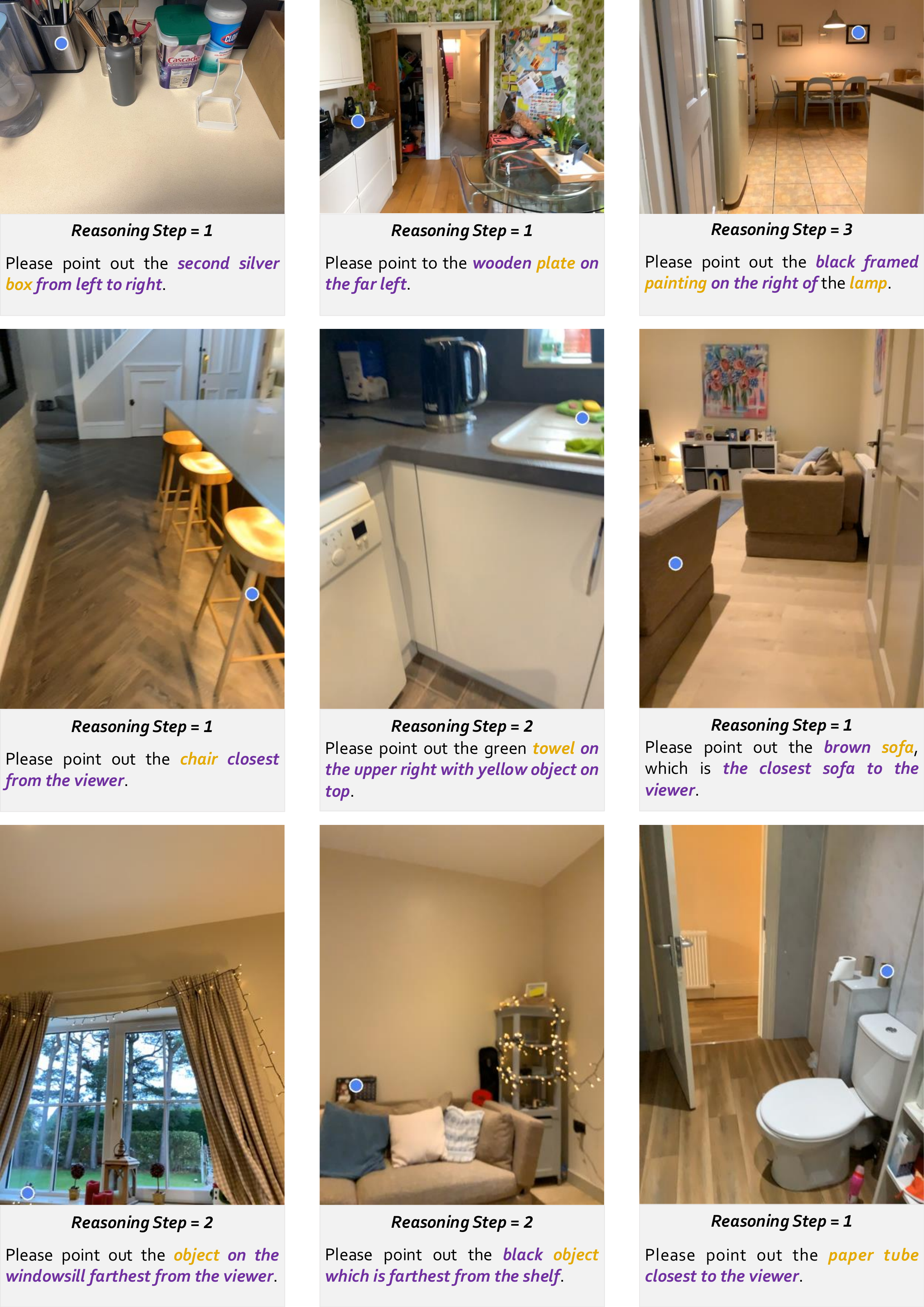}
    \caption{\textbf{Pointing Examples of RoboBrain 2.0}. The blue point represents the model's spatial referring prediction.}
    \label{fig:point_4}
    \vspace{-1.5em}
\end{figure*}

\begin{figure*}[!ht]
    \centering
    \includegraphics[width=0.9\linewidth]{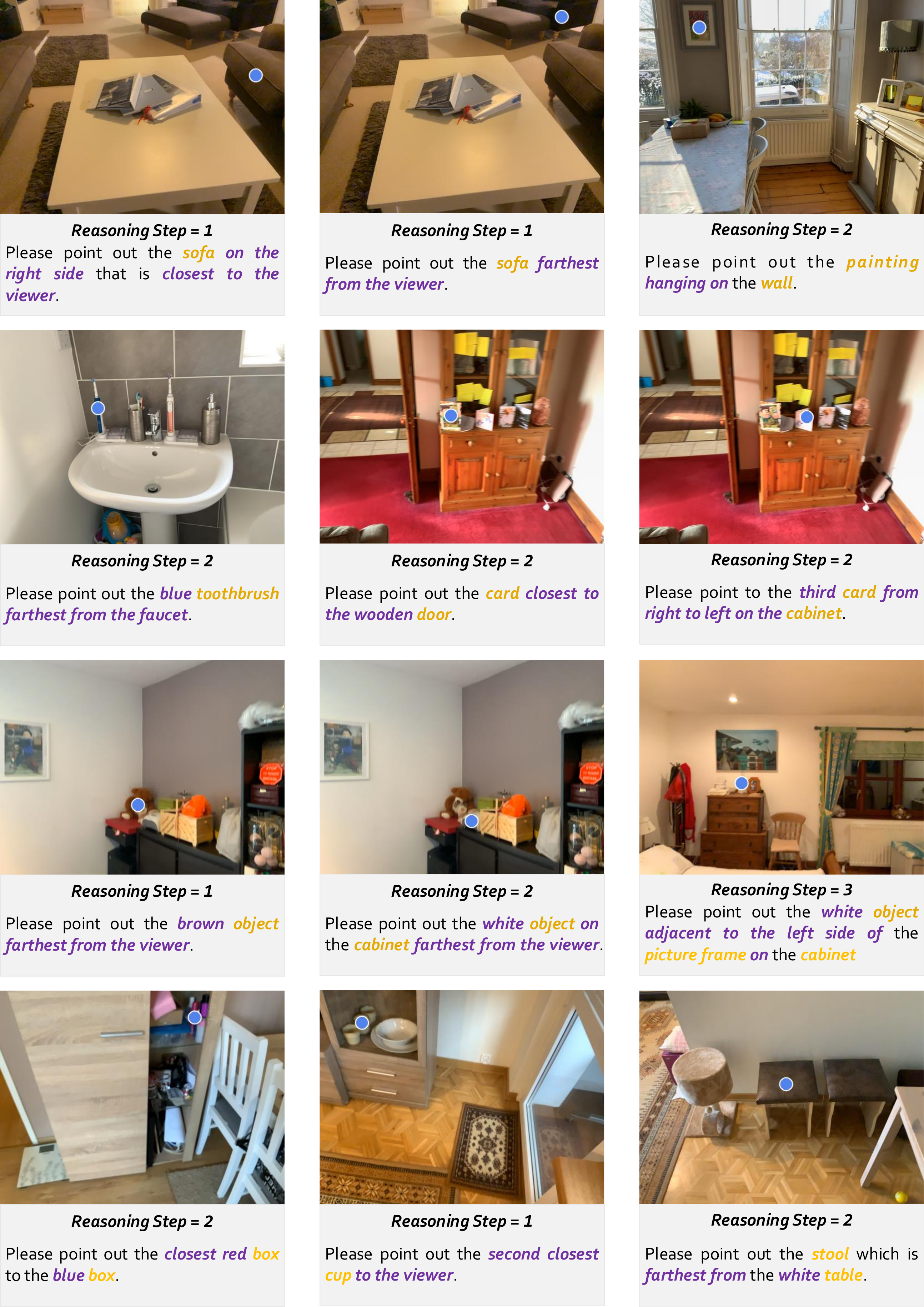}
    \caption{\textbf{Pointing Examples of RoboBrain 2.0}. The blue point represents the model's spatial referring prediction.}
    \label{fig:point_5}
    \vspace{-1.5em}
\end{figure*}

\begin{figure*}[!ht]
    \centering
    \includegraphics[width=0.9\linewidth]{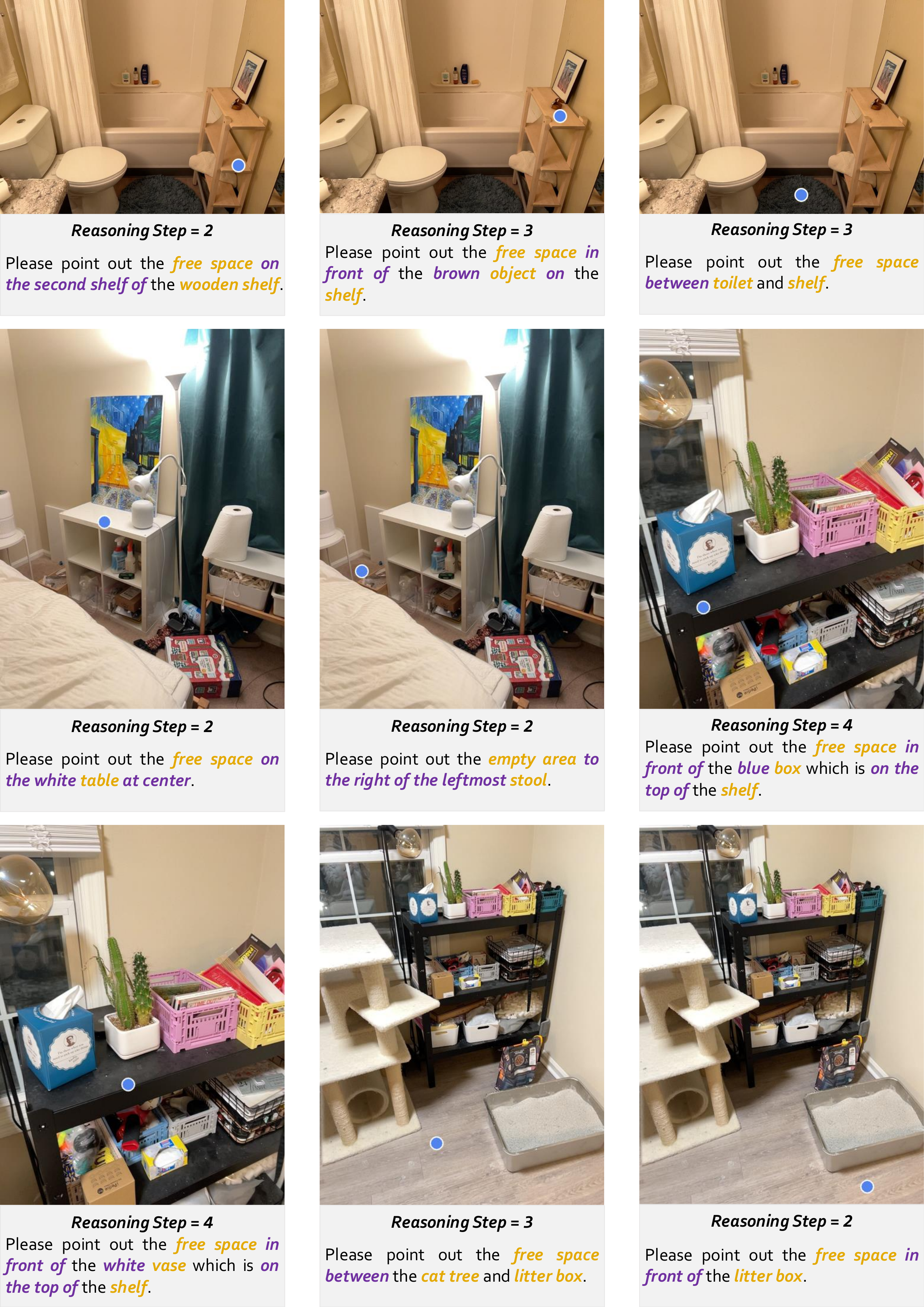}
    \caption{\textbf{Pointing Examples of RoboBrain 2.0}. The blue point represents the model's spatial referring prediction.}
    \label{fig:point_6}
    \vspace{-1.5em}
\end{figure*}

\begin{figure*}[!ht]
    \centering
    \includegraphics[width=0.9\linewidth]{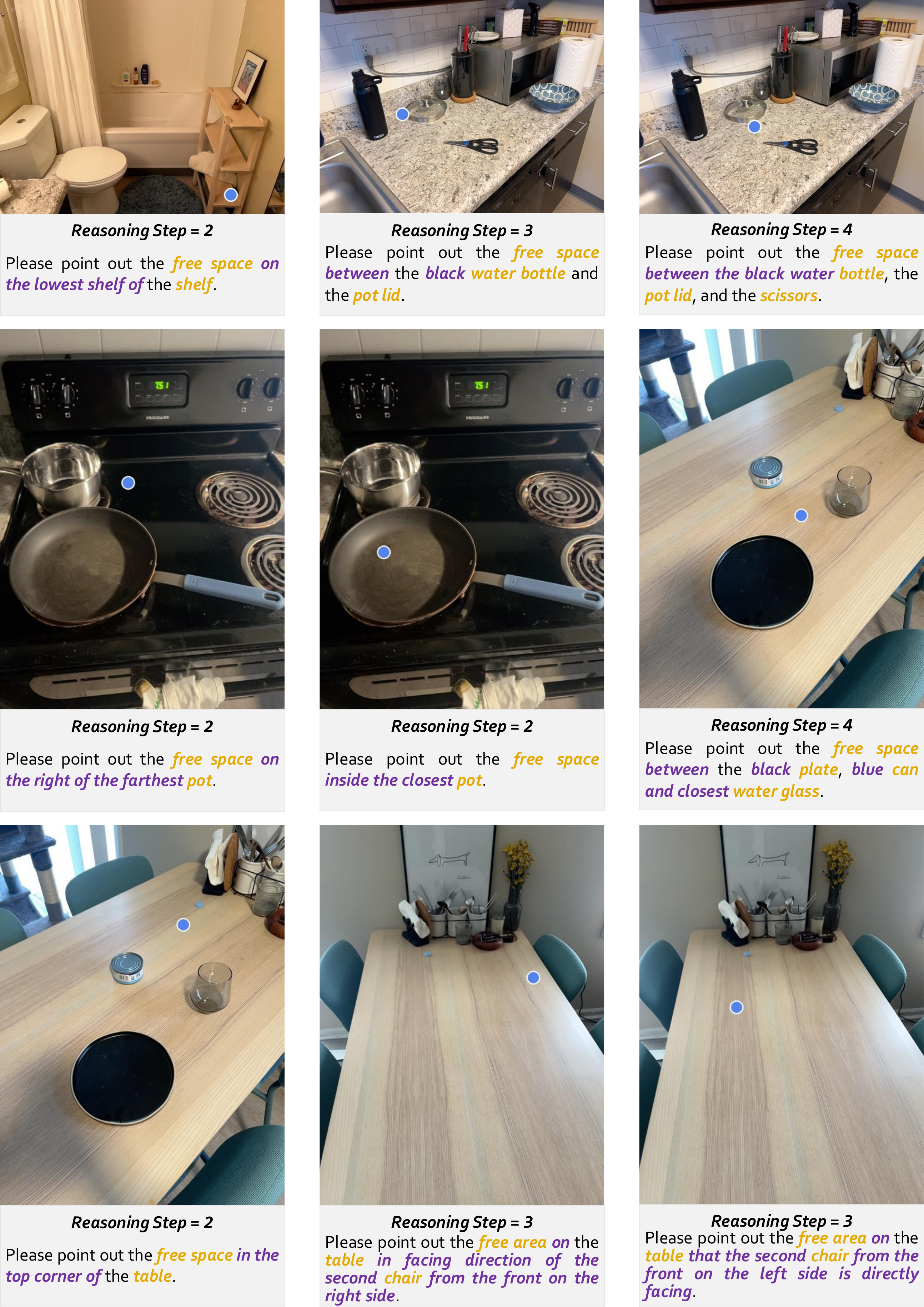}
    \caption{\textbf{Pointing Examples of RoboBrain 2.0}. The blue point represents the model's spatial referring prediction.}
    \label{fig:point_7}
    \vspace{-1.5em}
\end{figure*}

\begin{figure*}[!ht]
    \centering
    \includegraphics[width=0.9\linewidth]{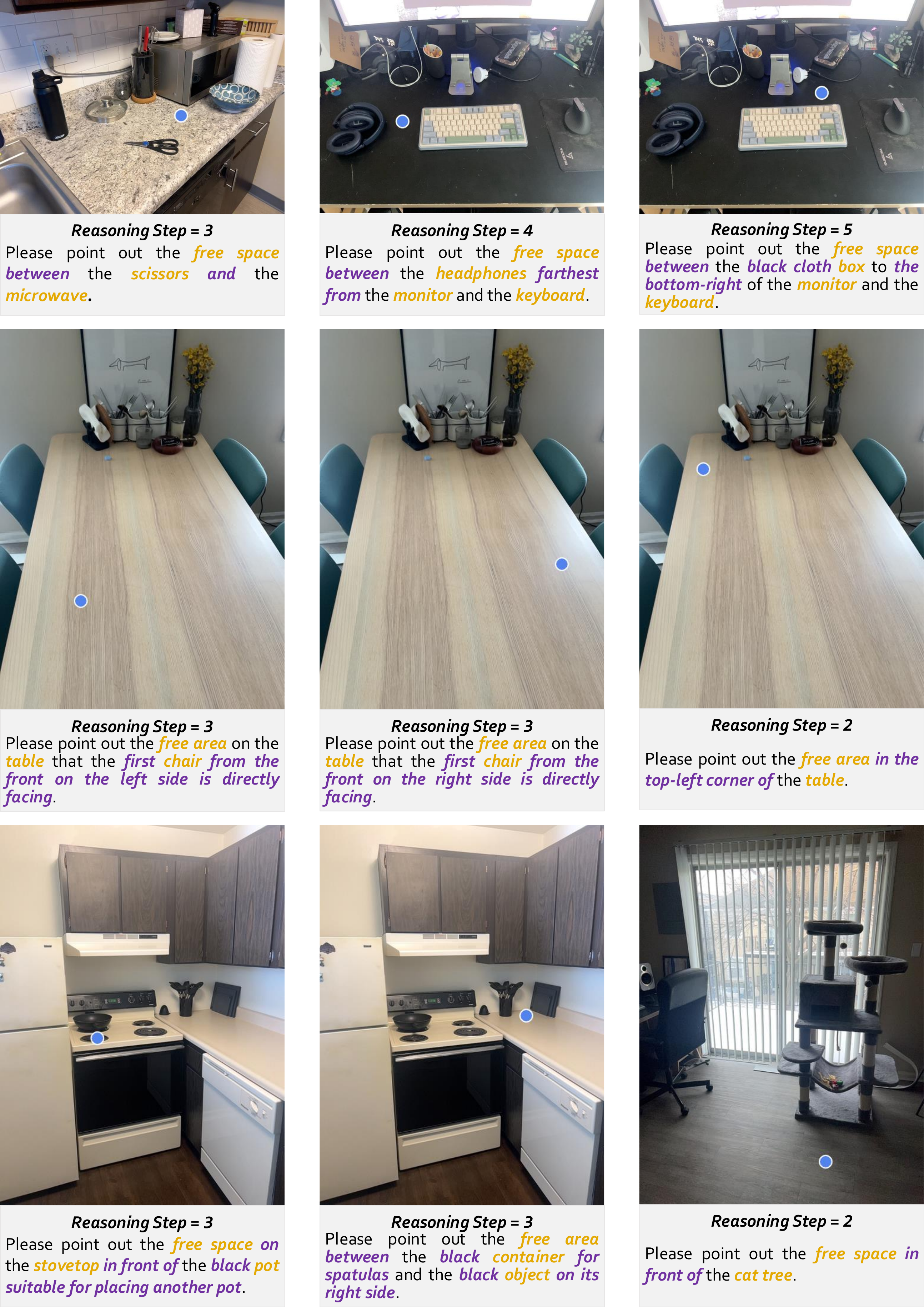}
    \caption{\textbf{Pointing Examples of RoboBrain 2.0}. The blue point represents the model's spatial referring prediction.}
    \label{fig:point_8}
    \vspace{-1.5em}
\end{figure*}

\begin{figure*}[!ht]
    \centering
    \includegraphics[width=0.9\linewidth]{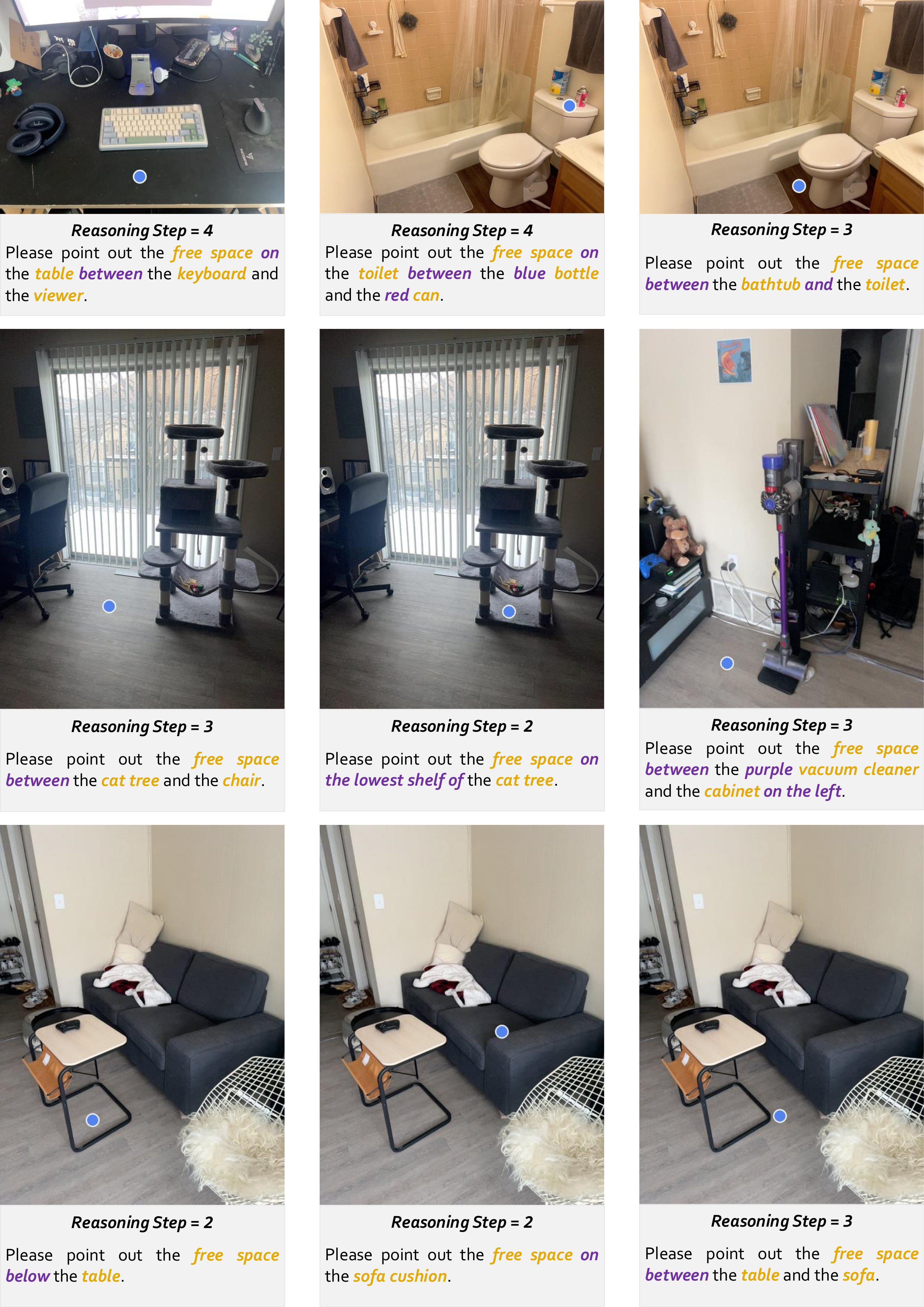}
    \caption{\textbf{Pointing Examples of RoboBrain 2.0}. The blue point represents the model's spatial referring prediction.}
    \label{fig:point_9}
    \vspace{-1.5em}
\end{figure*}

\begin{figure*}[!ht]
    \centering
    \includegraphics[width=0.9\linewidth]{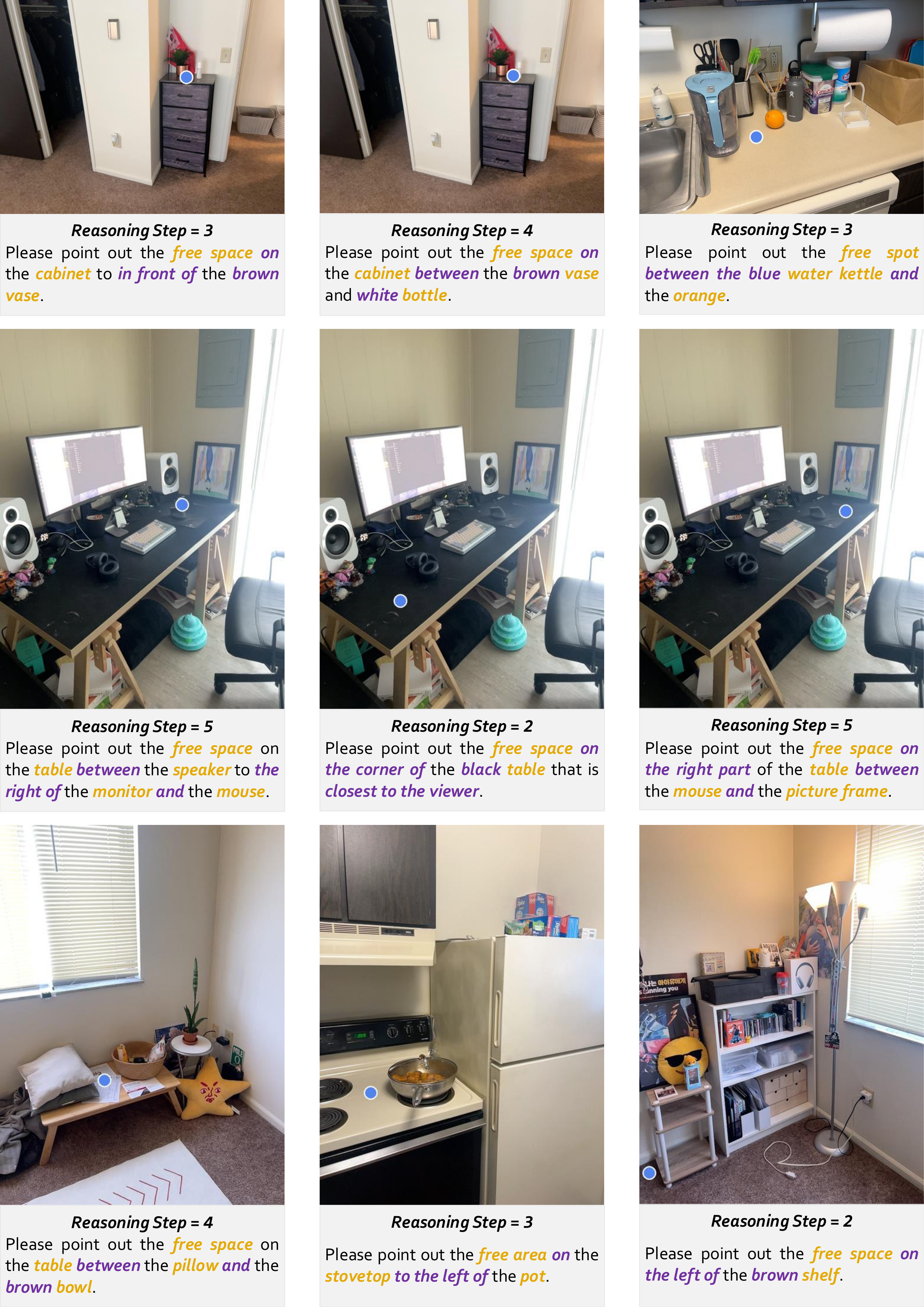}
    \caption{\textbf{Pointing Examples of RoboBrain 2.0}. The blue point represents the model's spatial referring prediction.}
    \label{fig:point_10}
    \vspace{-1.5em}
\end{figure*}

\begin{figure*}[!ht]
    \centering
    \includegraphics[width=0.9\linewidth]{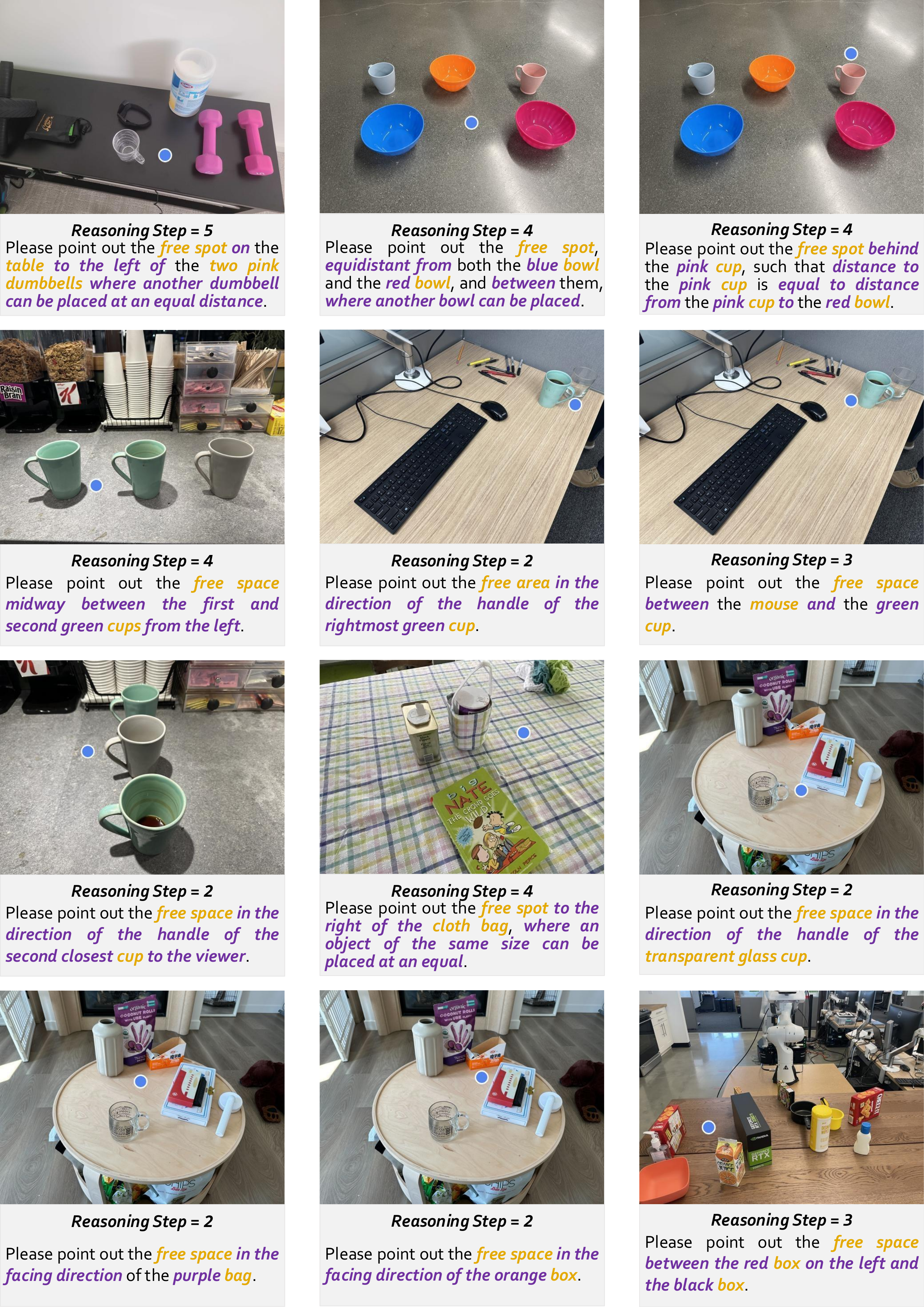}
    \caption{\textbf{Pointing Examples of RoboBrain 2.0}. The blue point represents the model's spatial referring prediction.}
    \label{fig:point_11}
    \vspace{-1.5em}
\end{figure*}

\begin{figure*}[!ht]
    \centering
    \includegraphics[width=0.9\linewidth]{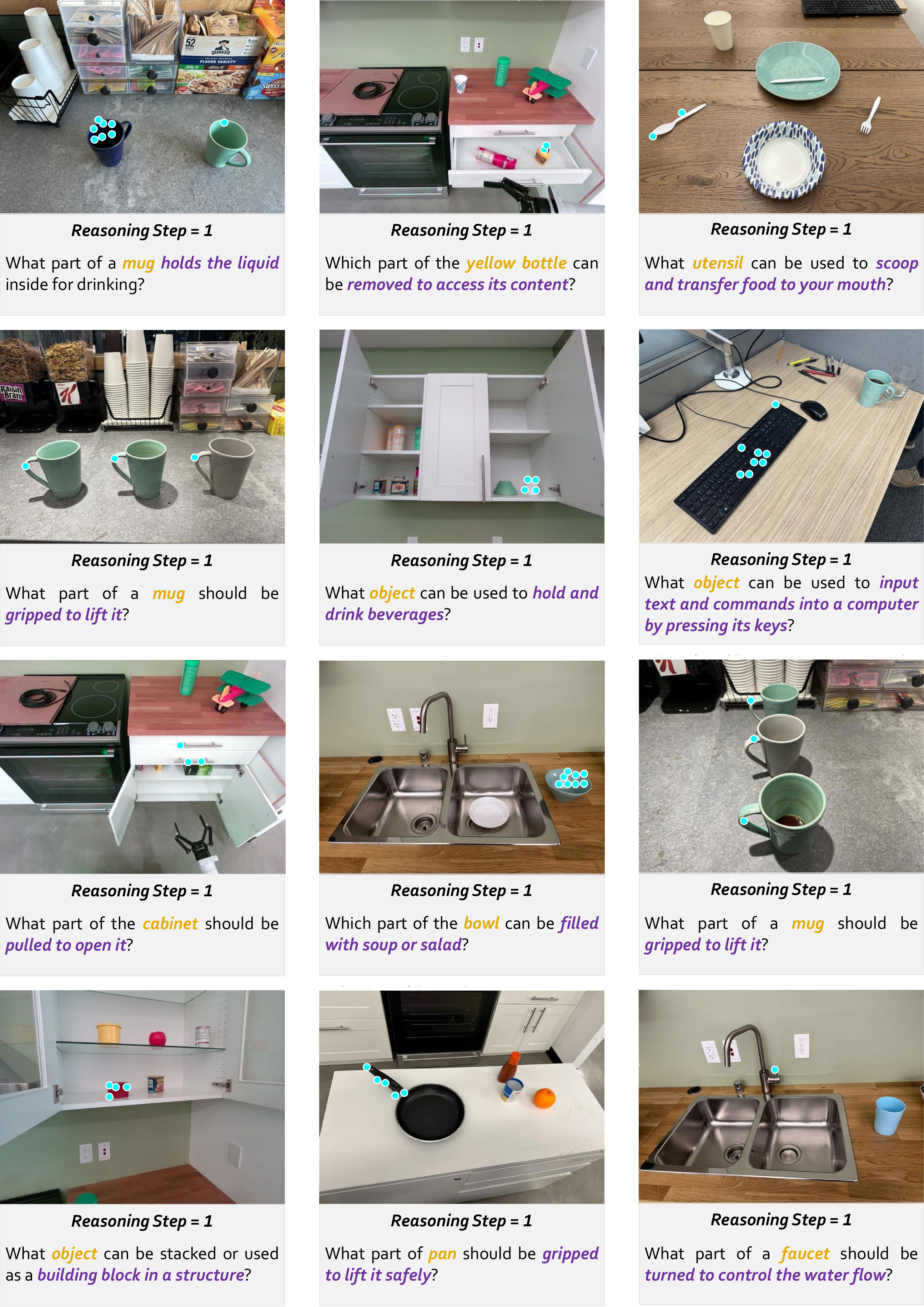}
    \caption{\textbf{Pointing Examples of RoboBrain 2.0}. The objects or their parts are pointed according to their affordances queried in the instruction.}
    \label{fig:point_12}
    \vspace{-1.5em}
\end{figure*}

\begin{figure*}[!ht]
    \centering
    \includegraphics[width=0.9\linewidth]{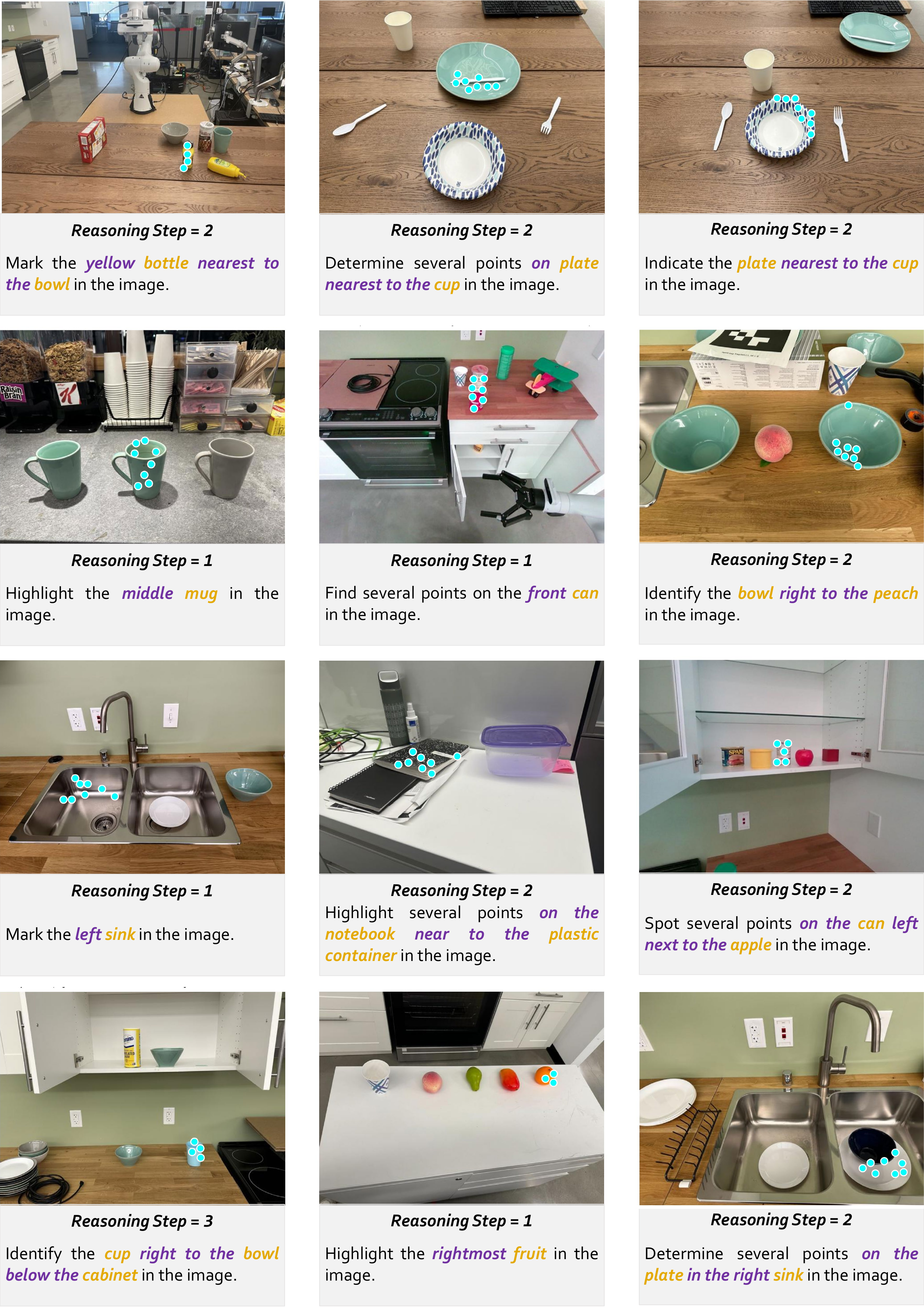}
    \caption{\textbf{Pointing Examples of RoboBrain 2.0}. The objects referred by spatial relations or object attributes are pointed out.}
    \label{fig:point_13}
    \vspace{-1.5em}
\end{figure*}

\begin{figure*}[!ht]
    \centering
    \includegraphics[width=0.9\linewidth]{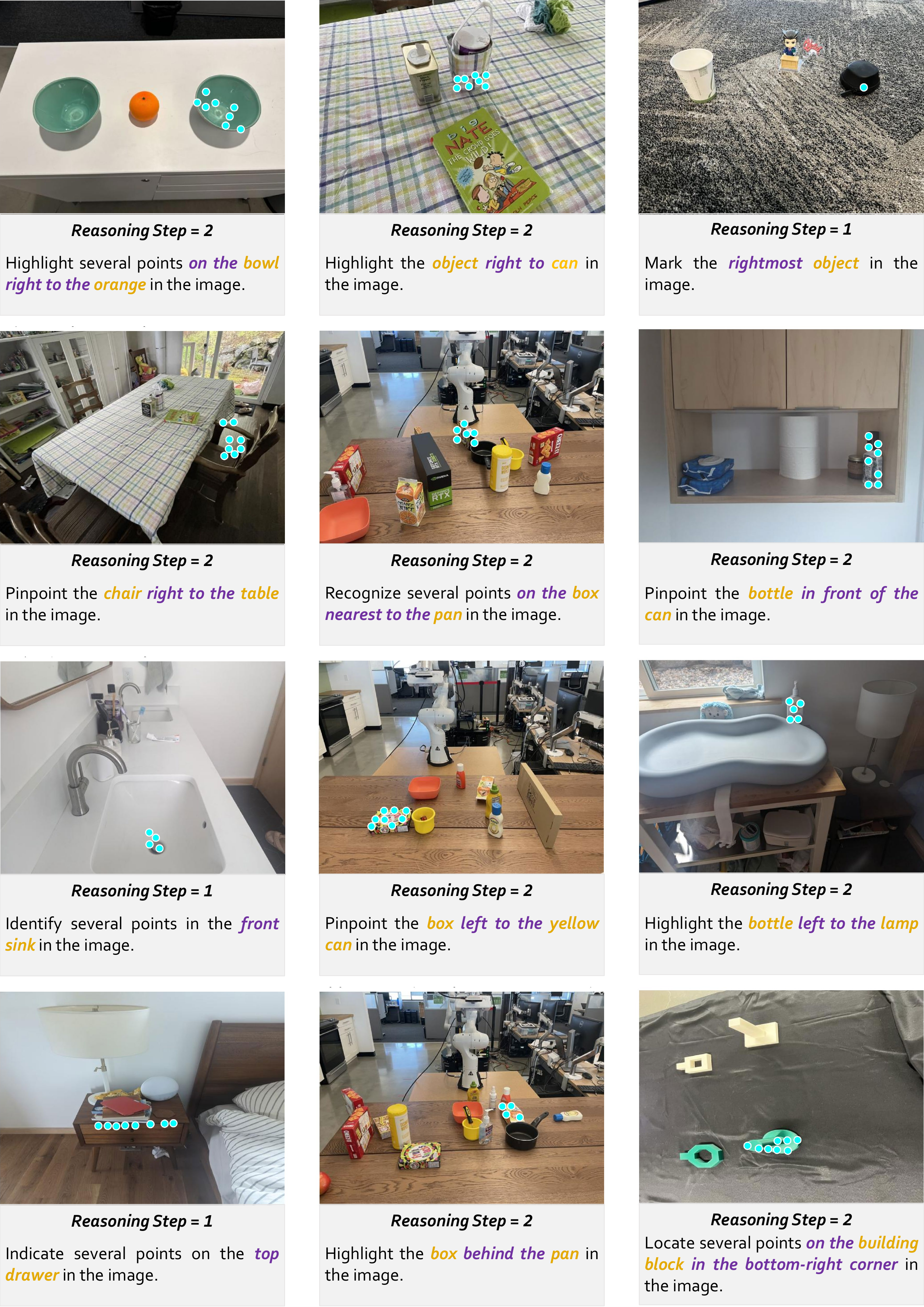}
    \caption{\textbf{Pointing Examples of RoboBrain 2.0}. The objects referred by spatial relations or object attributes are pointed out.}
    \label{fig:point_14}
    \vspace{-1.5em}
\end{figure*}

\begin{figure*}[!ht]
    \centering
    \includegraphics[width=0.9\linewidth]{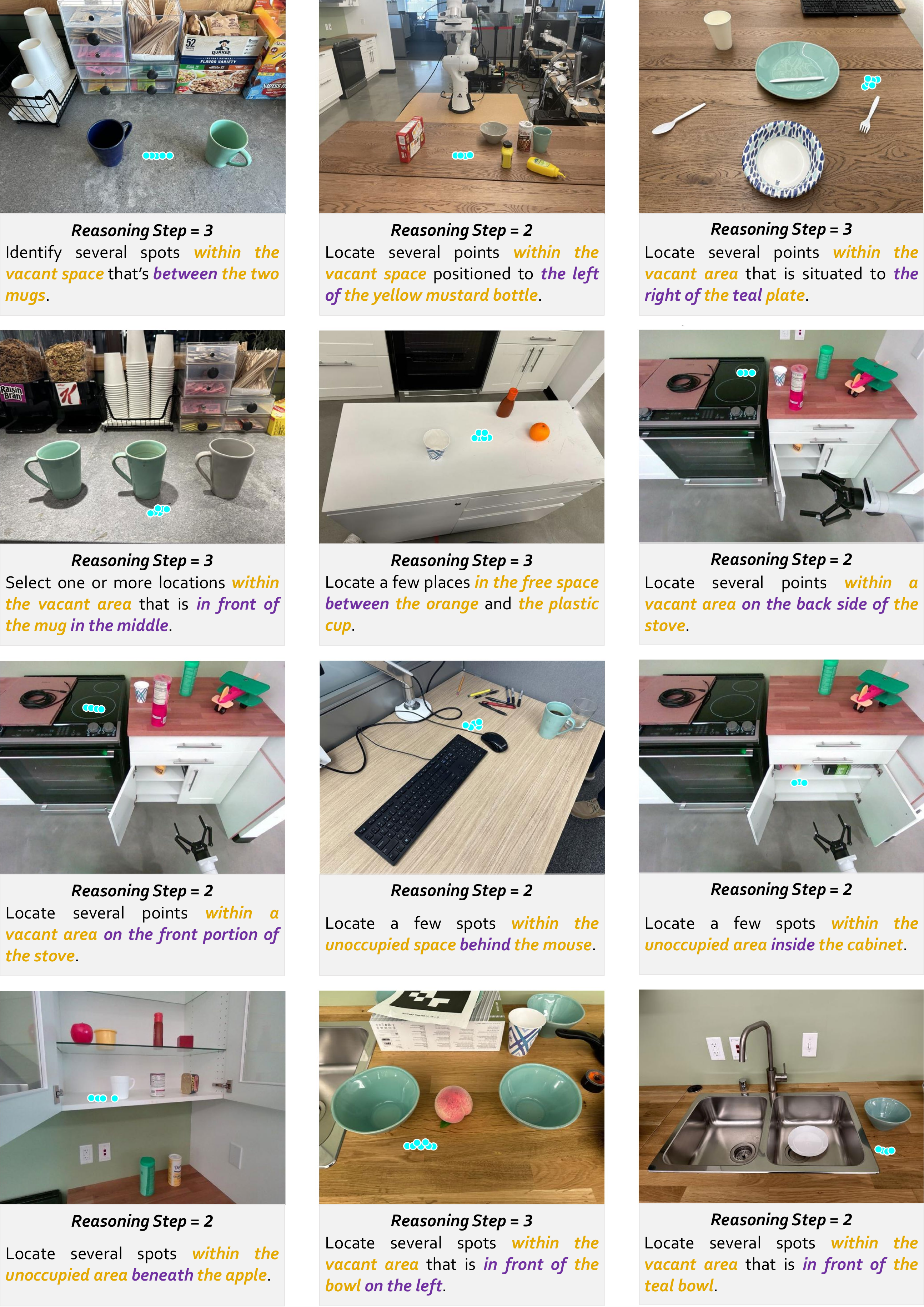}
    \caption{\textbf{Pointing Examples of RoboBrain 2.0}. The free space indicated by spatial relations and the referenced objects are pointed out.}
    \label{fig:point_15l}
    \vspace{-1.5em}
\end{figure*}

\begin{figure*}[!ht]
    \centering
    \includegraphics[width=0.9\linewidth]{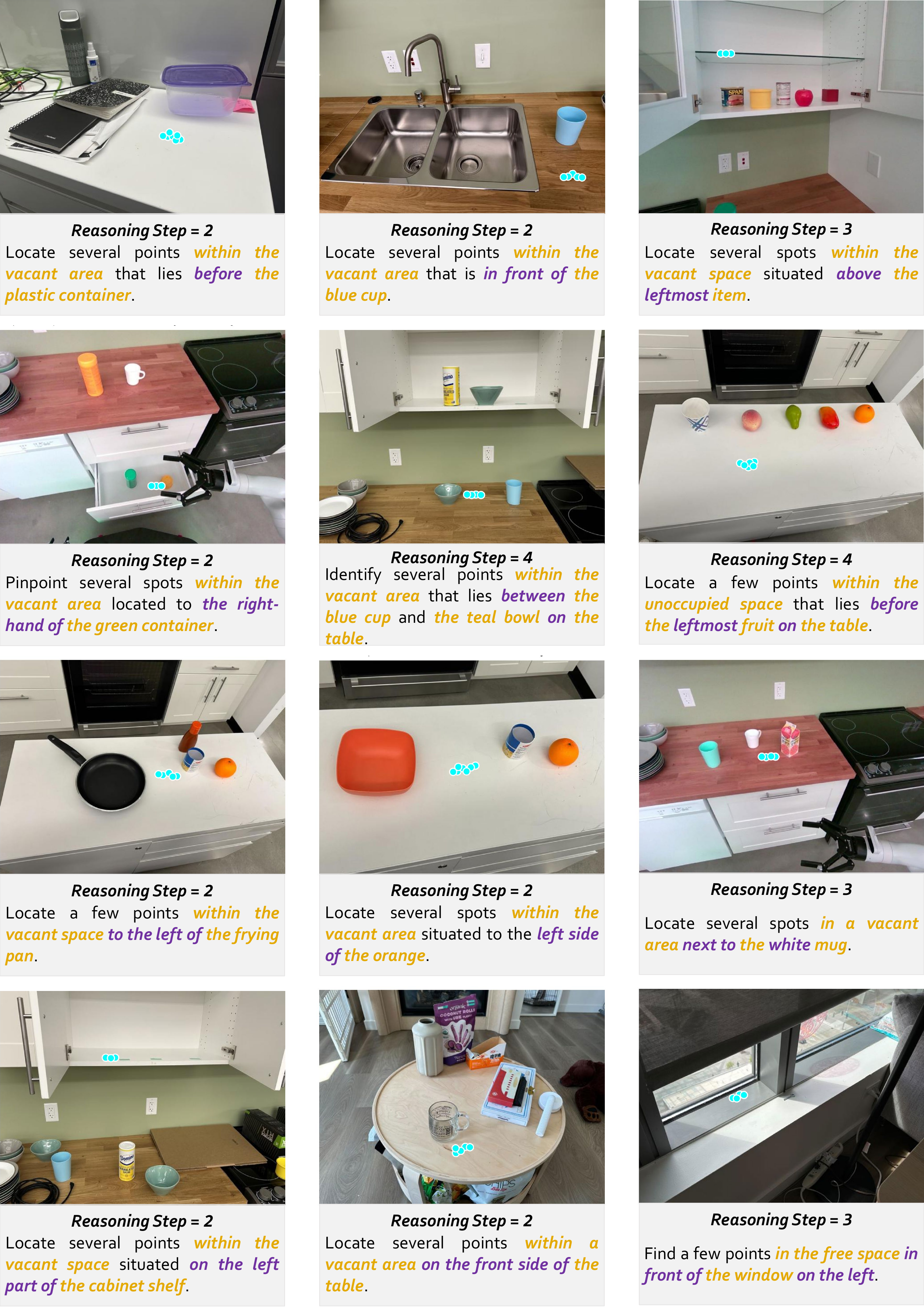}
    \caption{\textbf{Pointing Examples of RoboBrain 2.0}. The free space indicated by spatial relations and the referenced objects are pointed out.}
    \label{fig:point_16}
    \vspace{-1.5em}
\end{figure*}

In the pointing task, RoboBrain 2.0 is required to identify and point to specific objects within an image based on complex spatial instructions. For instance, given the instruction ``Please point out the orange box,'' the model accurately identifies the orange box in the image. Similarly, for more complex instructions such as ``Please point out the brown box on the shelf,'' RoboBrain 2.0 demonstrates its ability to understand spatial relationships and accurately points to the correct object. The model's proficiency in this task is further exemplified by its performance on a variety of pointing examples, as shown in ~\Cref{fig:point_1}-\Cref{fig:point_16}. These examples highlight the model's robust spatial reasoning capabilities, enabling it to handle a wide range of pointing tasks with high precision. Whether the instructions involve simple object identification or more intricate spatial relationships, RoboBrain 2.0 consistently demonstrates its ability to accurately locate and point to the specified objects. This capability is crucial for applications in robotics and automation, where precise object localization is essential for effective interaction with the physical environment.

\begin{figure*}[!ht]
    \centering
    \includegraphics[width=0.9\linewidth]{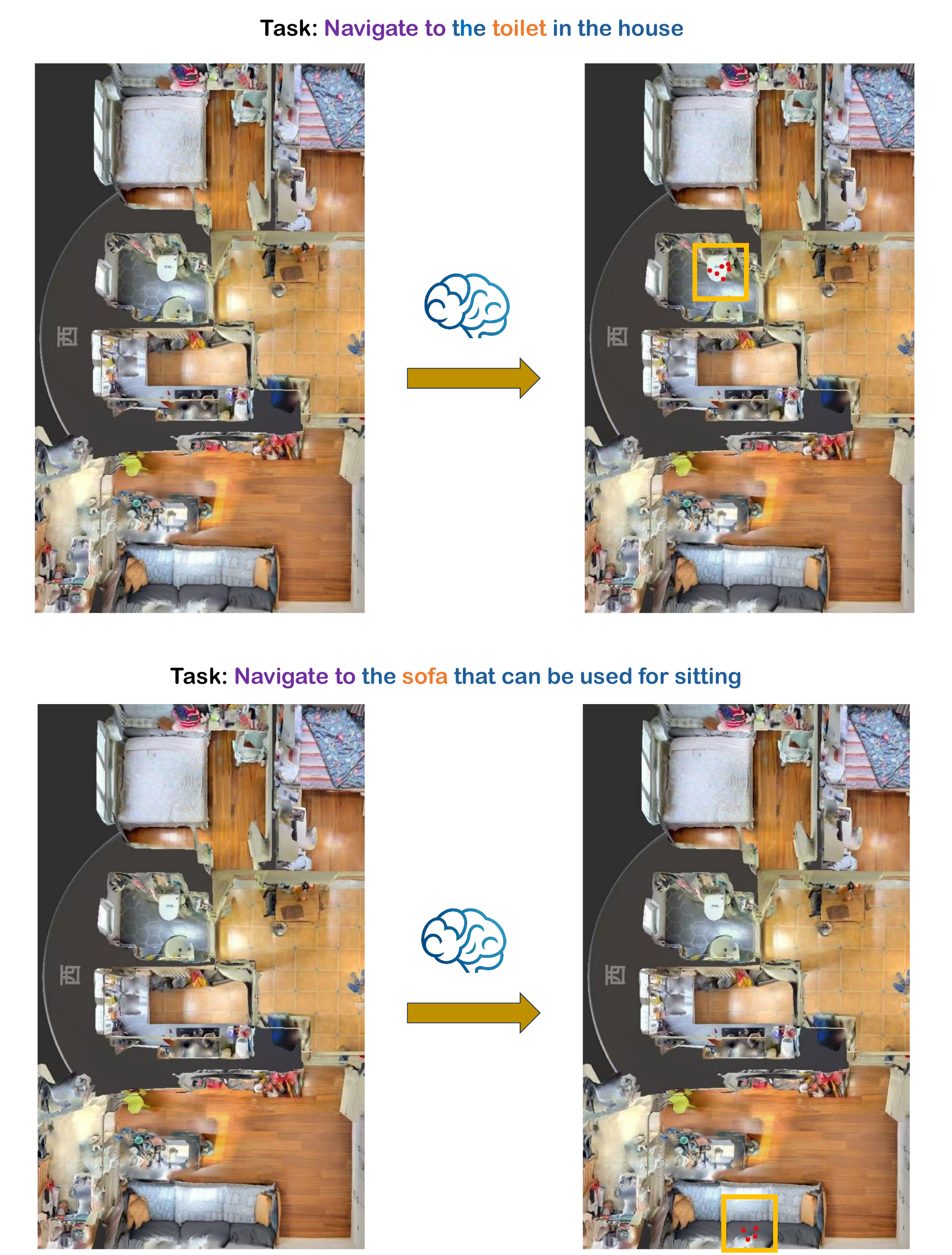}
    \caption{\textbf{Pointing Examples of RoboBrain 2.0}. The free space indicated by spatial relations and the referenced objects are pointed out.}
    \label{fig:point_17}
    \vspace{-1.5em}
\end{figure*}

\clearpage
\subsection{Examples for Affordance}
\begin{figure*}[!ht]
    \centering
    \includegraphics[width=0.95\linewidth]{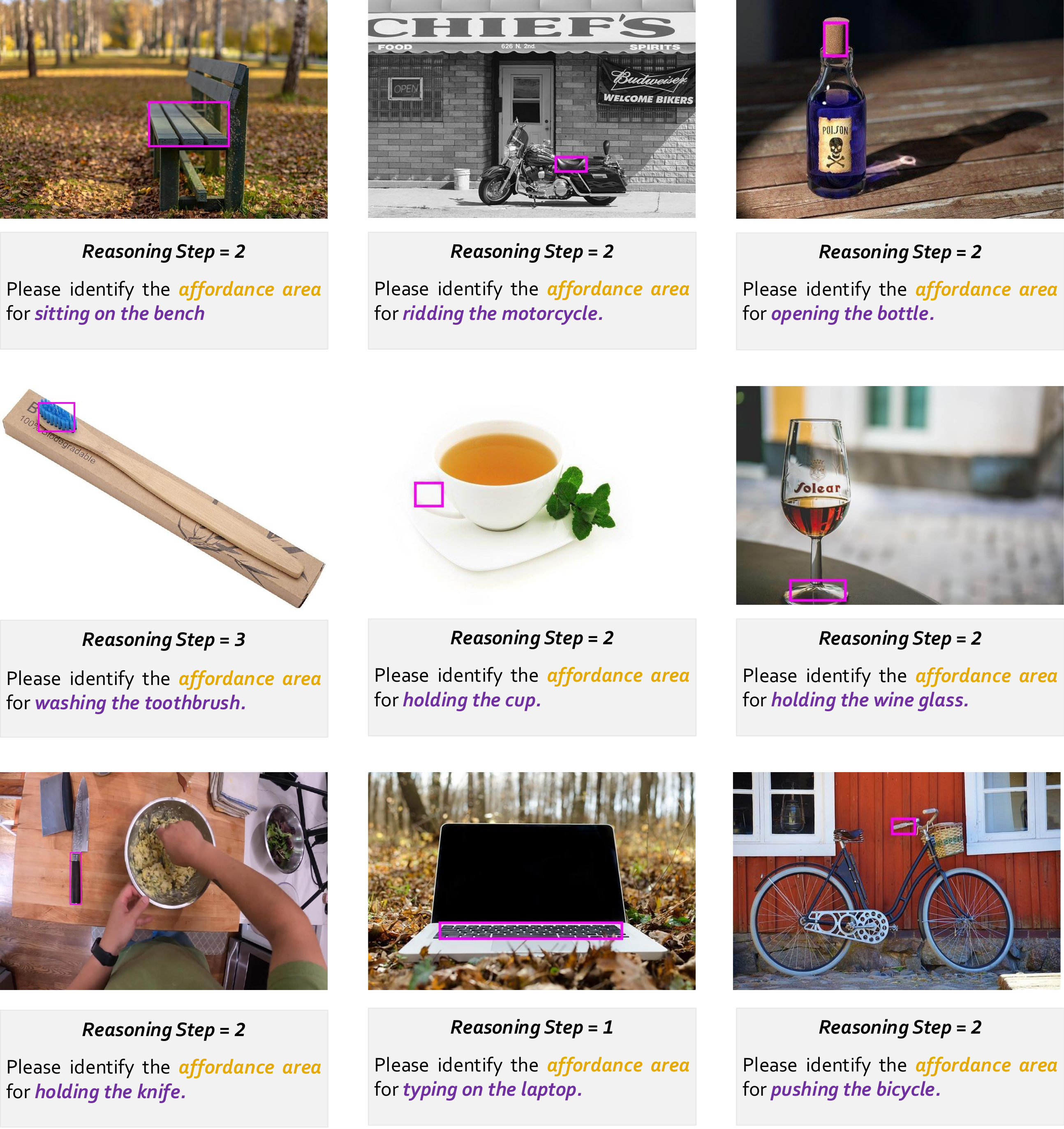}
    \caption{\textbf{Affordance Examples of RoboBrain 2.0}. The purple bounding boxes denote the actionable affordance areas for specific tasks.}
    \label{fig:Affordance_1}
    \vspace{-1.5em}
\end{figure*}

\begin{figure*}[!ht]
    \centering
    \includegraphics[width=0.93\linewidth]{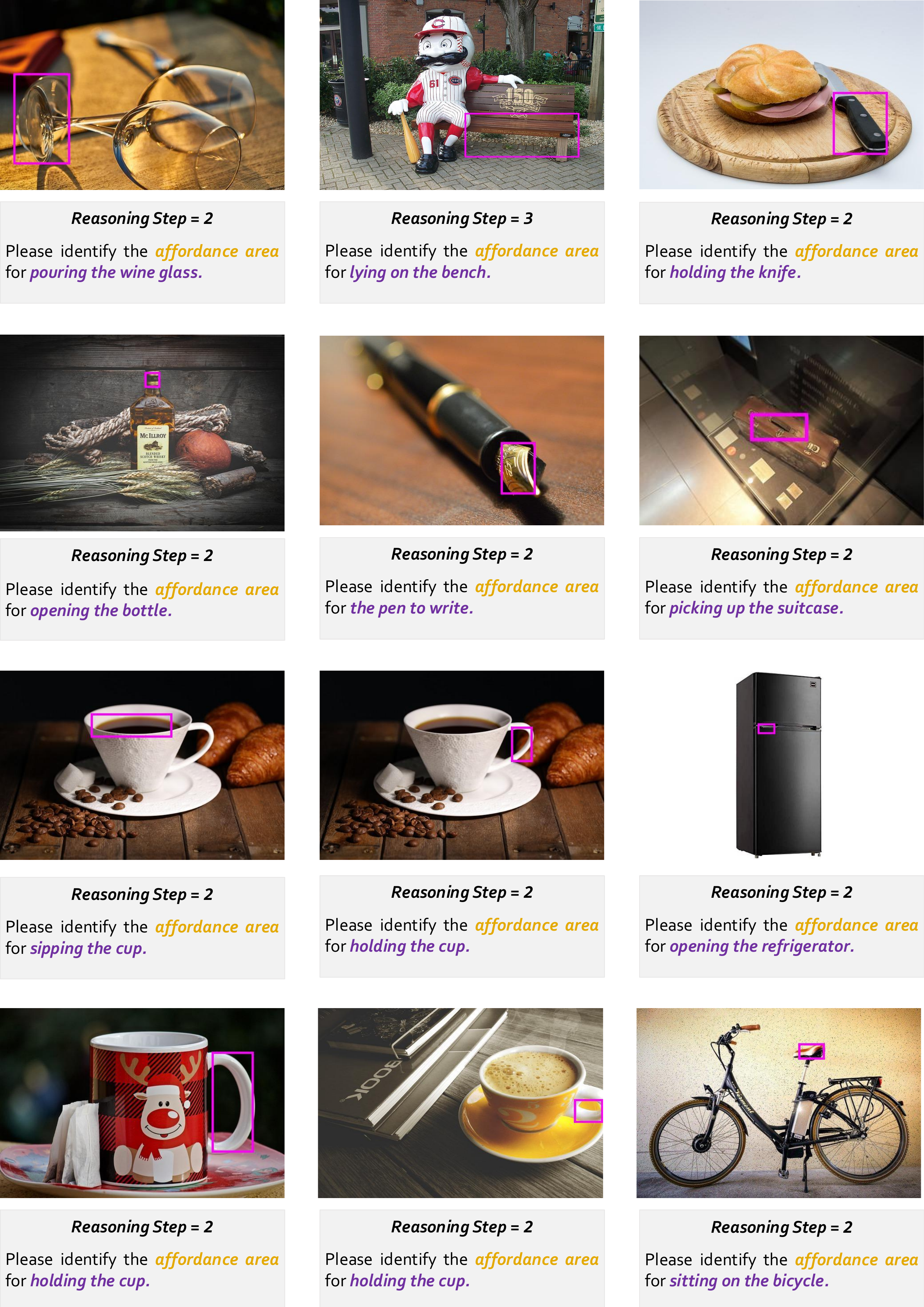}
    \caption{\textbf{Affordance Examples of RoboBrain 2.0}. The purple bounding boxes denote the actionable affordance areas for specific tasks.}
    \label{fig:Affordance_2}
    \vspace{-1.5em}
\end{figure*}

The affordance task assesses RoboBrain 2.0's understanding of object functionalities and interaction possibilities. For example, when asked ``What part of a mug holds the liquid for drinking?'' the model correctly identifies the interior of the mug as the part that holds the liquid. In another example, the instruction ``Which part of a handbag can be grasped to carry it?'' is accurately answered by identifying the handle of the handbag. These examples showcase the model's ability to reason about object affordances, making it capable of understanding how objects can be interacted with in the real world. As shown in ~\Cref{fig:Affordance_1}-\Cref{fig:Affordance_2}, the model demonstrates its proficiency in identifying functional parts of objects and their potential uses.


\clearpage
\subsection{Examples for Trajectory}
\begin{figure*}[!ht]
    \centering
    \vspace{1em}\includegraphics[width=0.91\linewidth]{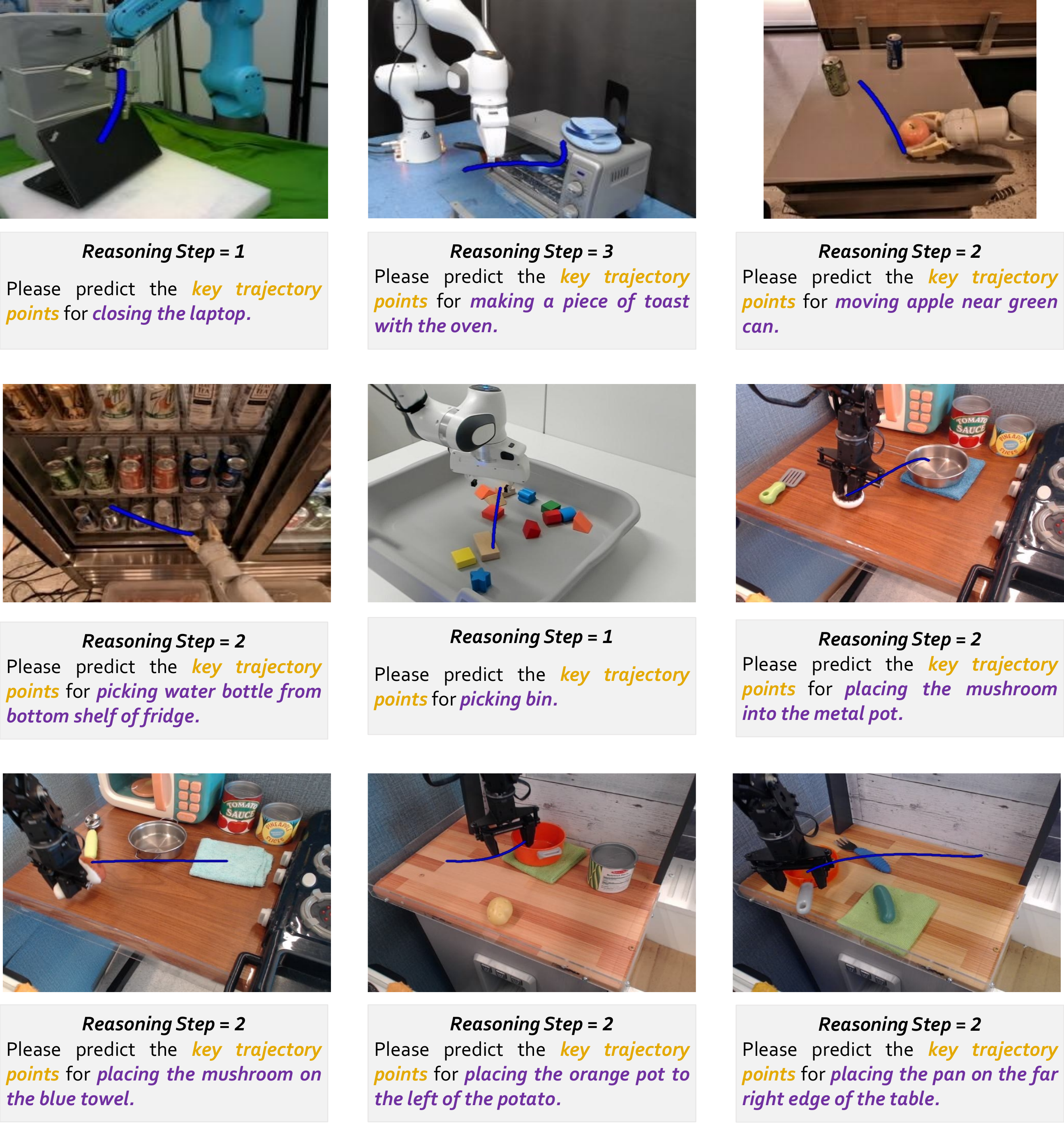}
    \caption{\textbf{Trajectory Examples of RoboBrain 2.0}. The blue trajectories, composed of key trajectory points, represent the model-predicted paths for task completion.}
    \label{fig:traj_1}
    \vspace{-1.5em}
\end{figure*}

\begin{figure*}[!ht]
    \centering
    \includegraphics[width=0.9\linewidth]{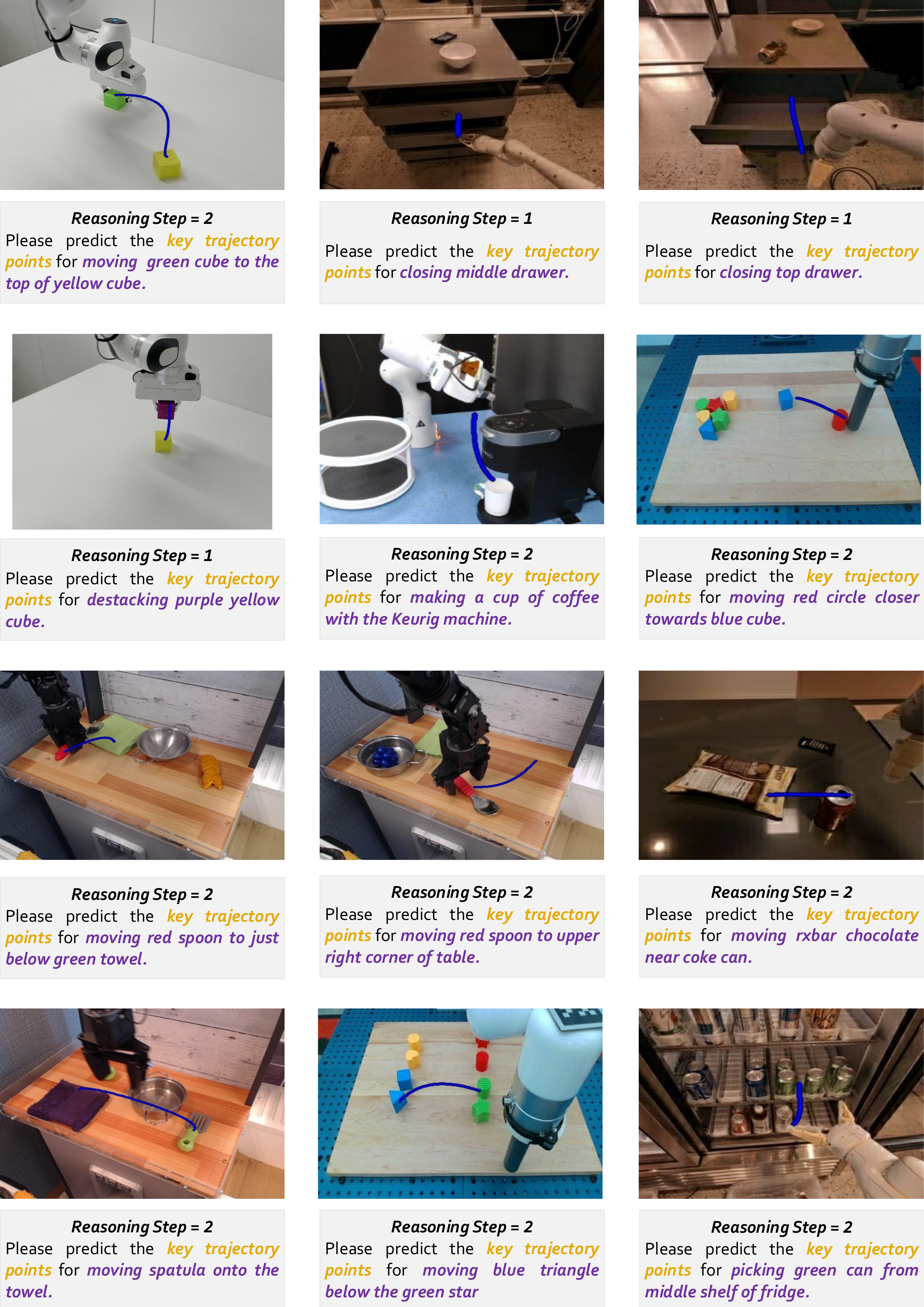}
    \caption{\textbf{Trajectory Examples of RoboBrain 2.0}. The blue trajectories, composed of key trajectory points, represent the model-predicted paths for task completion.}
    \label{fig:traj_2}
    \vspace{-1.5em}
\end{figure*}

The trajectory task evaluates the model's ability to predict and navigate paths based on given instructions. For instance, given the instruction ``Please provide the trajectory to move the robot arm to grasp the apple,'' RoboBrain 2.0 generates a smooth and efficient path for the robot arm to follow. The model's trajectory predictions are accurate and take into account the spatial constraints and obstacles in the environment, demonstrating its proficiency in spatial and temporal reasoning for navigation tasks. As shown in ~\Cref{fig:traj_1}-\Cref{fig:traj_2}, the model effectively plans and executes trajectories that are both optimal and collision-free.


\clearpage
\subsection{Examples for EgoPlan2}
\begin{figure*}[!ht]
    \centering
    \includegraphics[width=0.91\linewidth]{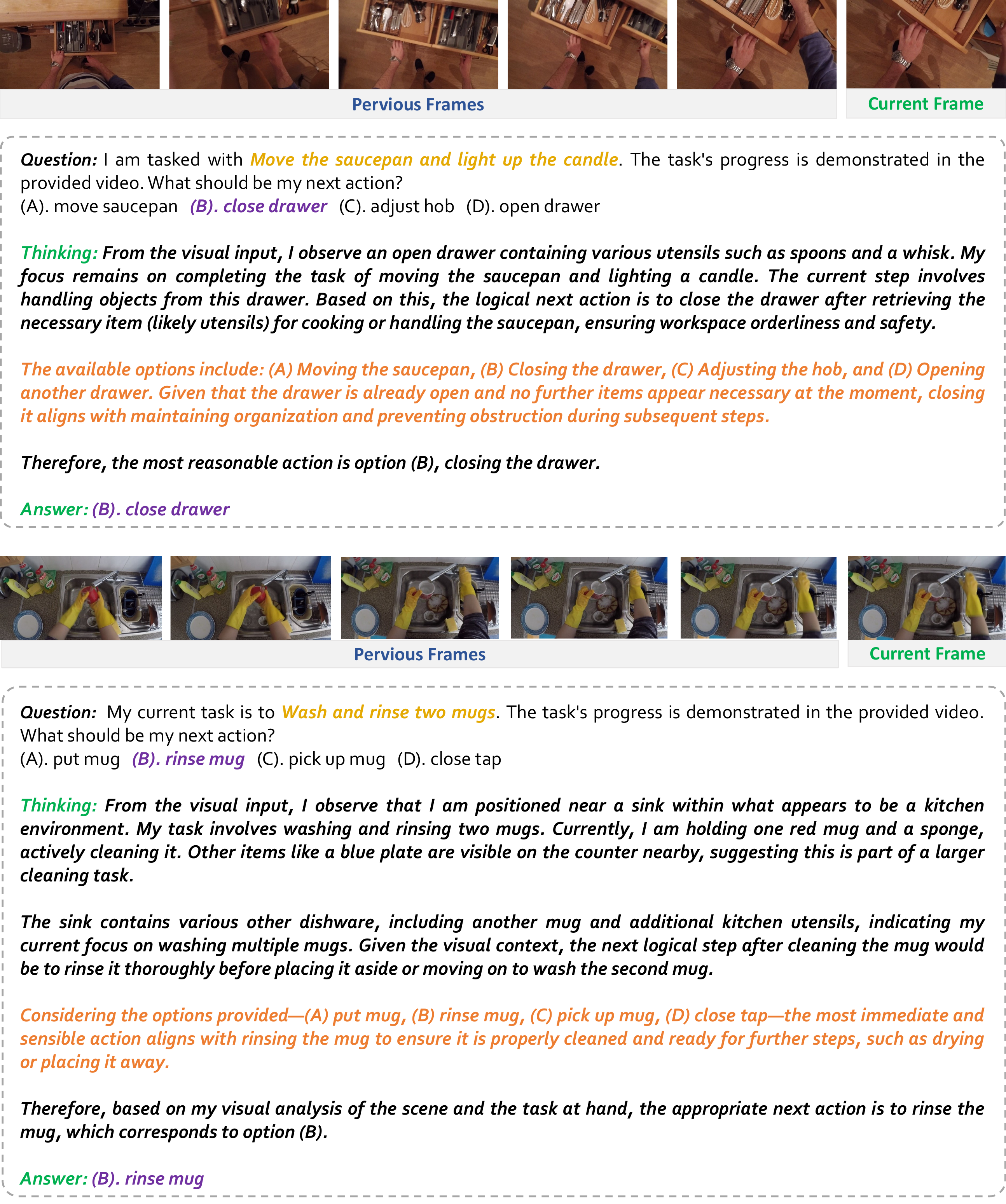}
    \caption{\textbf{EgoPlan Examples of RoboBrain 2.0}. The key reasoning process and final answer related to specific tasks are depicted.}
    \label{fig:egoplan_1}
    \vspace{-1.5em}
\end{figure*}

\begin{figure*}[!ht]
    \centering
    \includegraphics[width=0.9\linewidth]{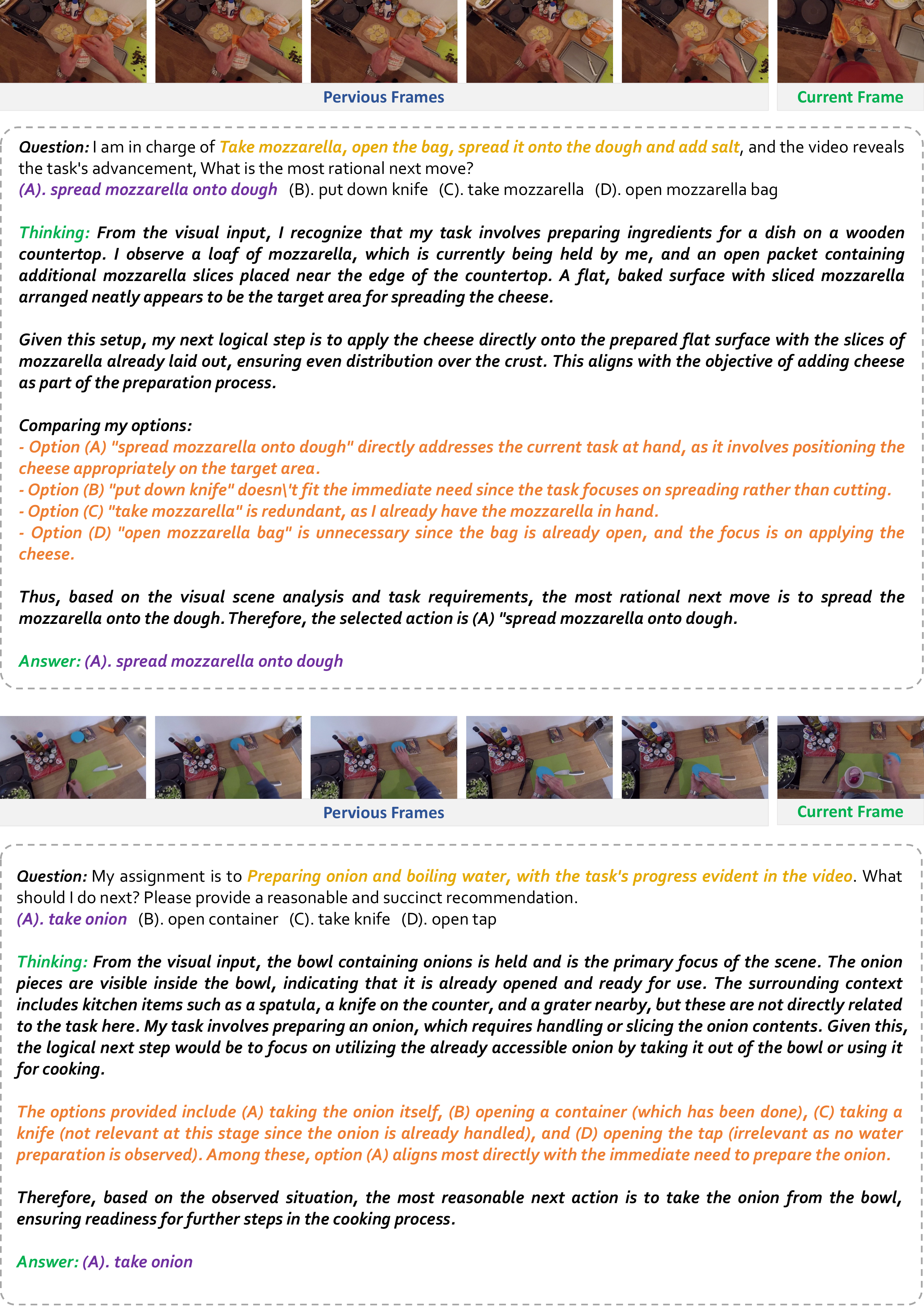}
    \caption{\textbf{EgoPlan Examples of RoboBrain 2.0}. The key reasoning process and final answer related to specific tasks are depicted.}
    \label{fig:egoplan_2}
    \vspace{-1.5em}
\end{figure*}

\begin{figure*}[!ht]
    \centering
    \includegraphics[width=0.9\linewidth]{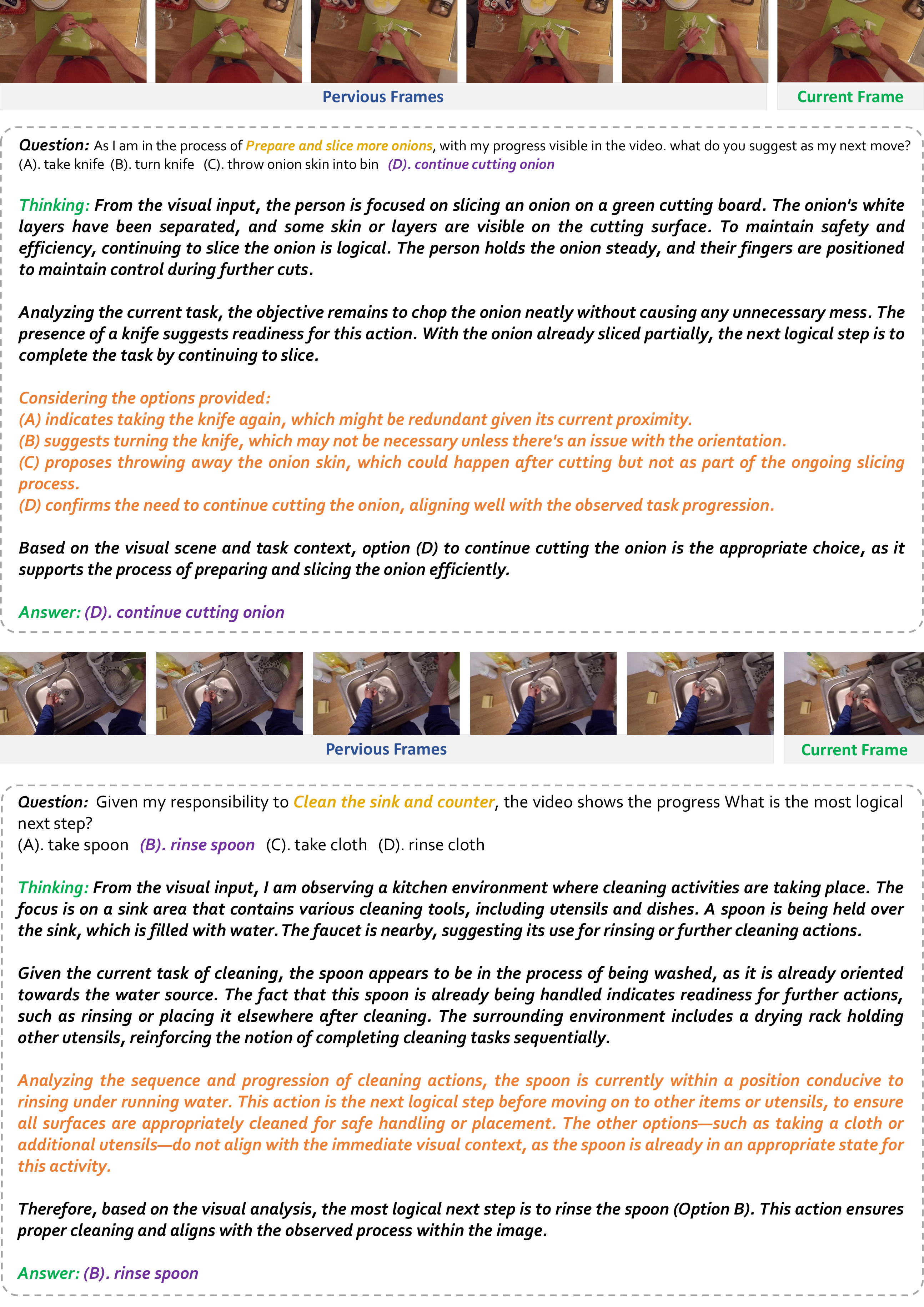}
    \caption{\textbf{EgoPlan Examples of RoboBrain 2.0}. The key reasoning process and final answer related to specific tasks are depicted.}
    \label{fig:egoplan_3}
    \vspace{-1.5em}
\end{figure*}

The EgoPlan2 task focuses on planning daily activities from an egocentric perspective. For instance, given the instruction ``Plan the steps to prepare a cup of coffee,'' RoboBrain 2.0 outlines a detailed sequence of actions, including locating the coffee machine, fetching the coffee beans, and following the steps to brew the coffee. The model's ability to break down complex tasks into actionable steps demonstrates its proficiency in task decomposition and sequential planning. As shown in ~\Cref{fig:egoplan_1}-\Cref{fig:egoplan_3}, the model effectively plans and executes multi-step tasks, showcasing its capabilities in long-horizon planning and task execution.


\clearpage
\subsection{Examples for Close-Loop Interaction}
\begin{figure*}[!ht]
    \centering
    \includegraphics[width=0.9\linewidth]{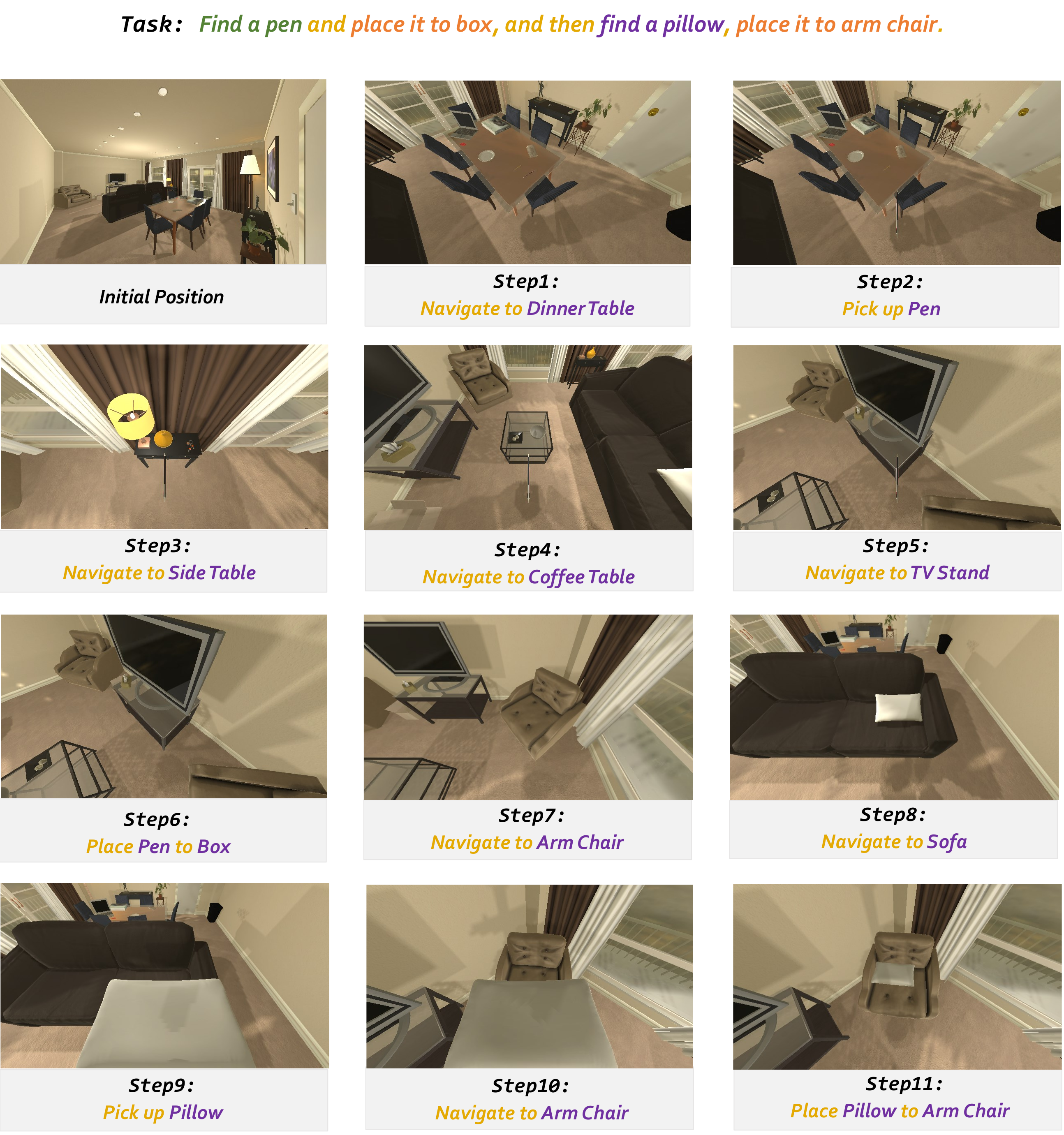}
    \caption{\textbf{Close-loop planning Examples of RoboBrain 2.0}. The key planning steps related to specific tasks are depicted.}
    \label{fig:close_loop_1}
    \vspace{-1.5em}
\end{figure*}

\begin{figure*}[!ht]
    \centering
    \includegraphics[width=0.9\linewidth]{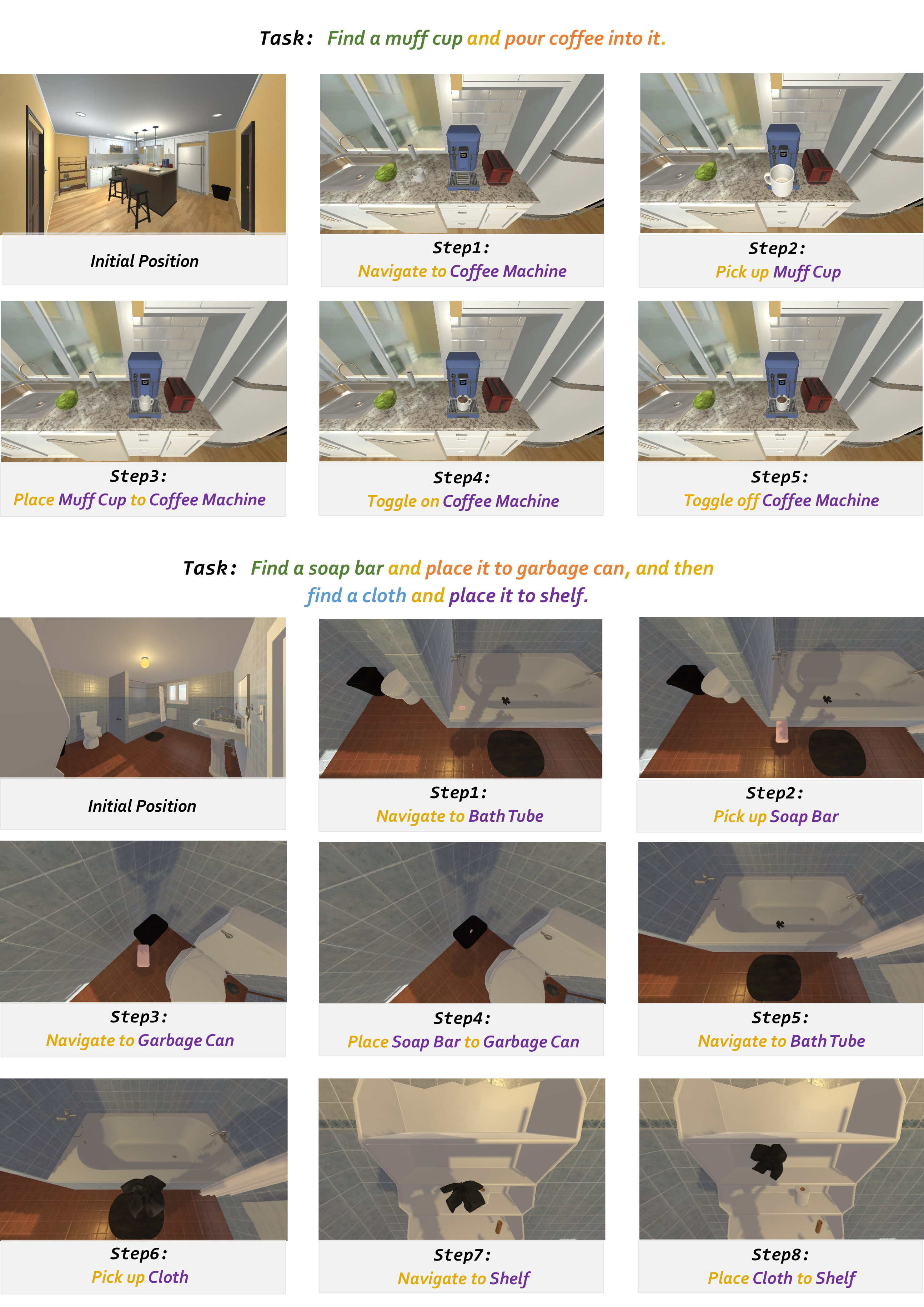}
    \caption{\textbf{Close-loop planning Examples of RoboBrain 2.0}. The key planning steps related to specific tasks are depicted.}
    \label{fig:close_loop_2}
    \vspace{-1.5em}
\end{figure*}

\begin{figure*}[!ht]
    \centering
    \includegraphics[width=0.9\linewidth]{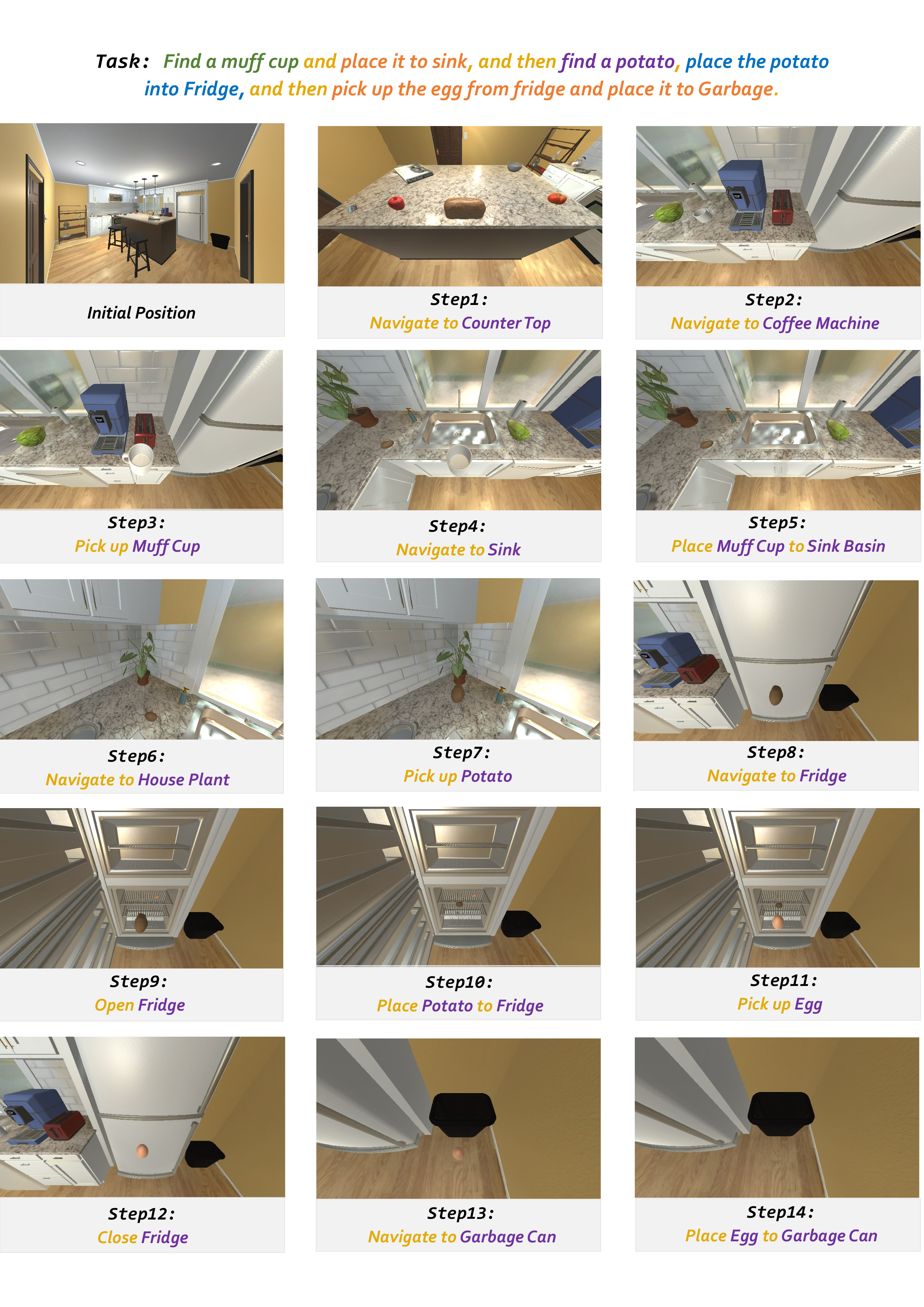}
    \caption{\textbf{Close-loop planning Examples of RoboBrain 2.0}. The key planning steps related to specific tasks are depicted.}
    \label{fig:close_loop_3}
    \vspace{-1.5em}
\end{figure*}

\begin{figure*}[!ht]
    \centering
    \includegraphics[width=0.9\linewidth]{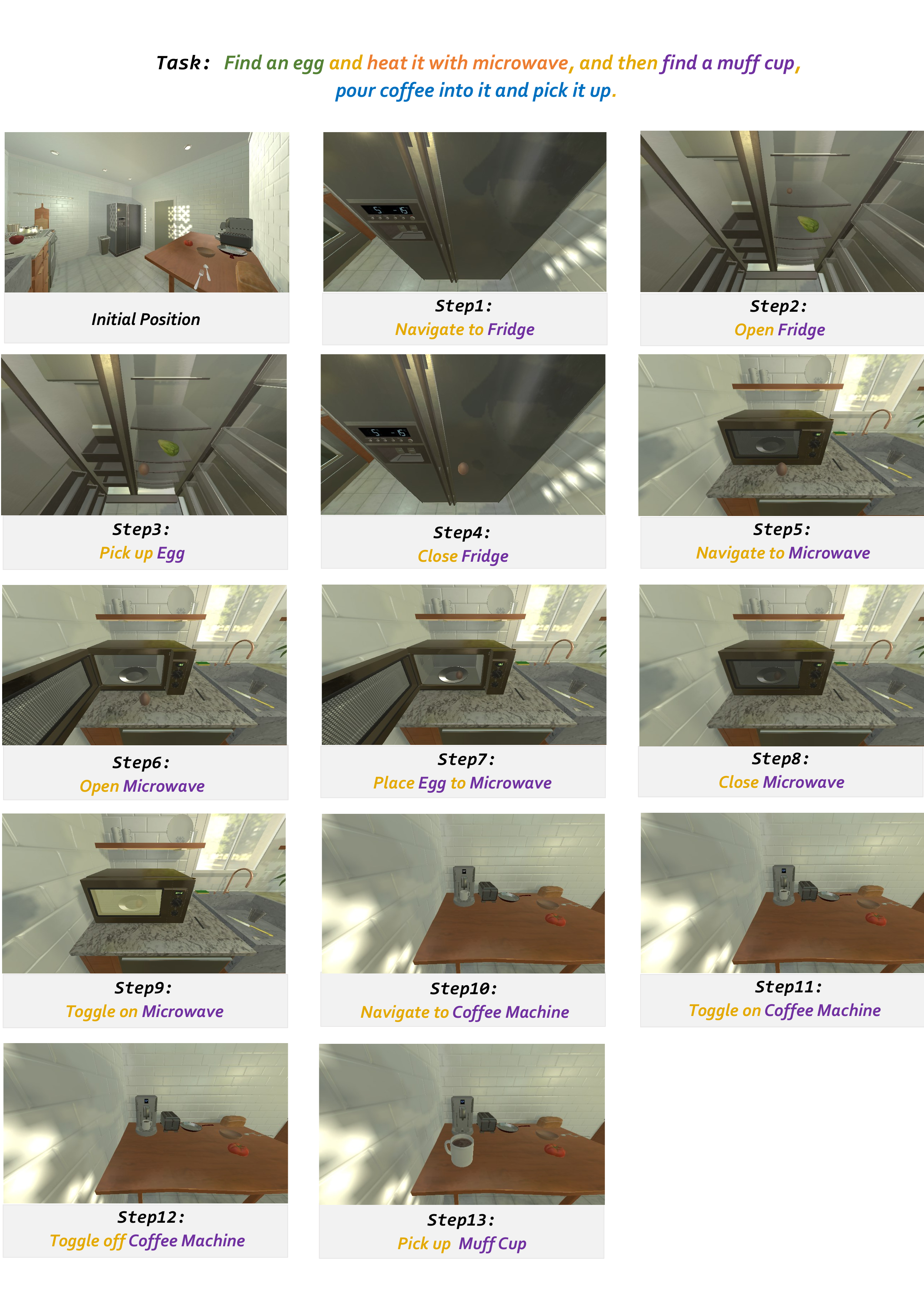}
    \caption{\textbf{Close-loop planning Examples of RoboBrain 2.0}. The key planning steps related to specific tasks are depicted.}
    \label{fig:close_loop_4}
    \vspace{-1.5em}
\end{figure*}

Close-loop interaction examples showcase RoboBrain 2.0's ability to engage in interactive reasoning with feedback. For example, in a scenario where the model is asked to ``Find a muff cup and pour coffee into it,'' it not only needs to navigate and search for the mug multiple times within the task environment but also must operate the coffee machine based on feedback to complete the pouring process. This iterative process highlights the model's capability to refine its actions based on real-time feedback, ensuring more accurate and reliable performance in interactive tasks. As shown in ~\Cref{fig:close_loop_1}-\Cref{fig:close_loop_4}, the model demonstrates its ability to adapt and improve its responses through iterative feedback loops.


\clearpage
\subsection{Examples for Multi-Robot Planning}
In multi-robot planning scenarios, RoboBrain 2.0 coordinates the actions of multiple robots to achieve a common goal. For example, in a supermarket scenario, the model plans the movements of multiple robots to efficiently restock shelves. The planning involves assigning specific tasks to each robot, coordinating their movements to avoid collisions, and ensuring that the overall goal is achieved in a timely manner. These examples highlight the model's advanced capabilities in multi-agent coordination and long-horizon planning. 

\begin{figure*}[!ht]
    \centering
    \includegraphics[width=0.9\linewidth]{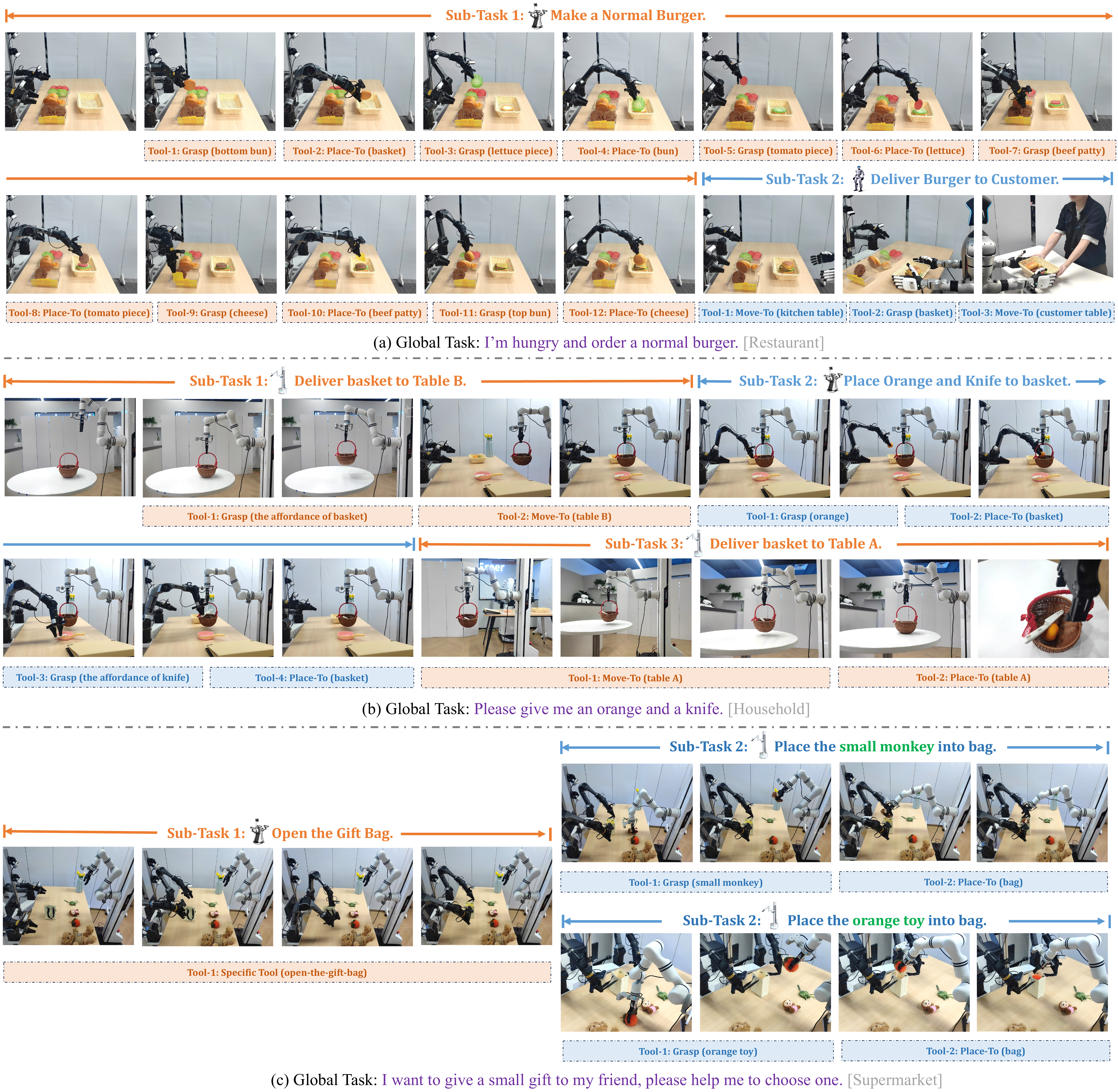}
    \caption{We showcase multi-robot collaboration in three scenarios: (a) Restaurant: Unitree G1 and Agilex robots prepare burgers. (b) Household: Realman and Agilex robots fetch items. (c) Supermarket: Robots coordinate gift selection and packaging.}
    \label{fig:multi_robot_1}
    \vspace{-1.5em}
\end{figure*}

\vspace{0.3em}
As shown in ~\Cref{fig:multi_robot_1}, the model demonstrates its ability to orchestrate complex multi-robot activities with high precision and efficiency. In the restaurant setting (\Cref{fig:multi_robot_1}(a)), a Unitree G1 humanoid and Agilex dual-arm robot collaborate on burger preparation and delivery for the command ``I'm hungry and order a normal burger,'' with RoboBrain 2.0 performing scene-aware task decomposition.
The household scenario (\Cref{fig:multi_robot_1}(b)) features a Realman single-arm and Agilex dual-arm robot executing commands like ``Give me an orange and a knife.''
In the supermarket (\Cref{fig:multi_robot_1}(c)), RoboBrain 2.0 assists customers with gift selection by analyzing dimensions and bag compatibility, coordinating the Realman robot for gift placement and the Agilex executing VLA-cerebellum skills like ``open the gift bag.'' Please refer to RoboOS~\cite{tan2025roboos} for more details.

\clearpage
\subsection{Examples for Synthetic Benchmarks}
\begin{figure*}[!ht]
    \centering
    \includegraphics[width=0.9\linewidth]{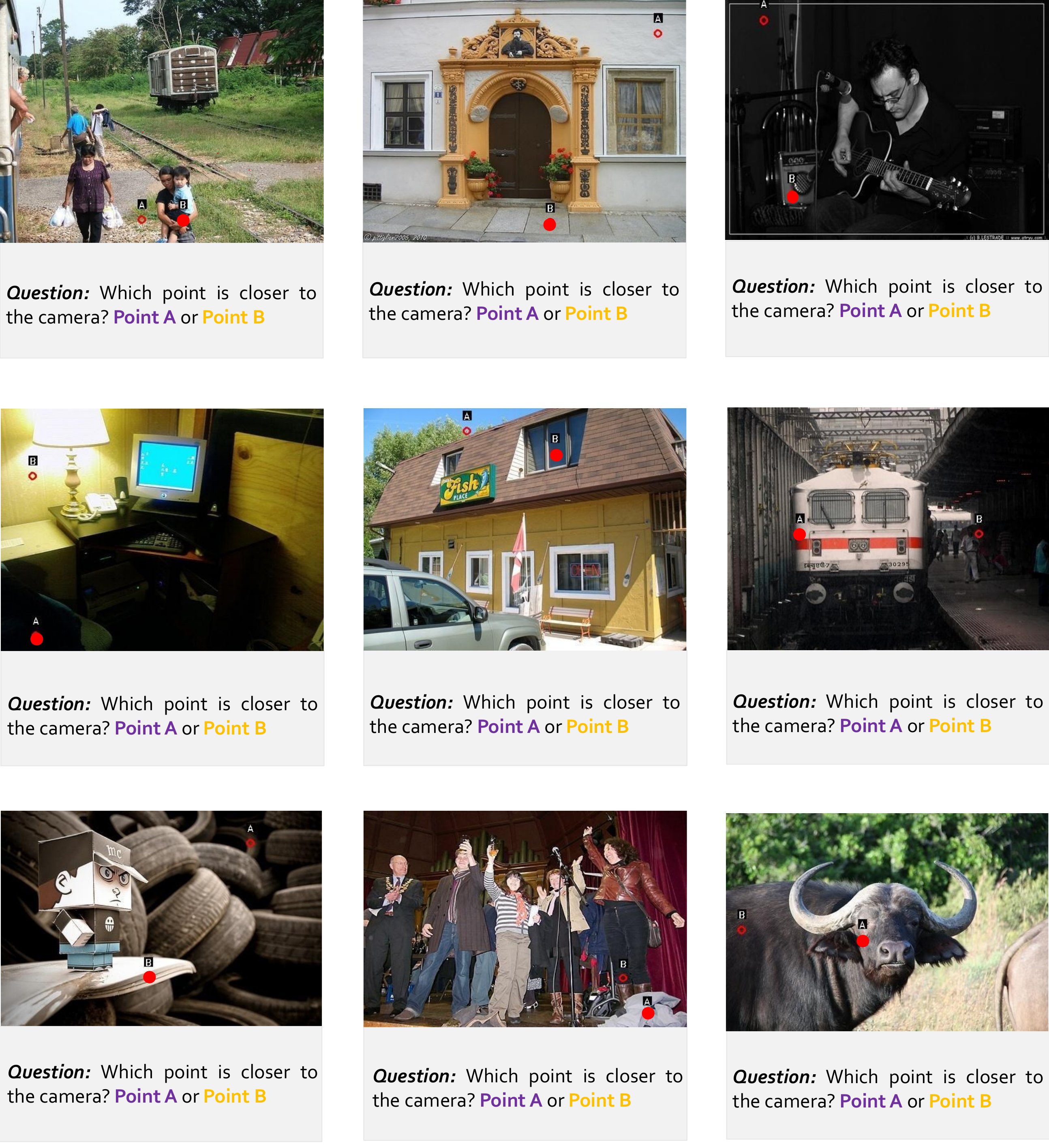}
    \caption{\textbf{CVbench Benchmark Examples of RoboBrain 2.0}. The solid circle in the diagram represents the selected point.}
    \label{fig:synthetic_1}
    \vspace{-1.5em}
\end{figure*}

\begin{figure*}[!ht]
    \centering
    \includegraphics[width=0.9\linewidth]{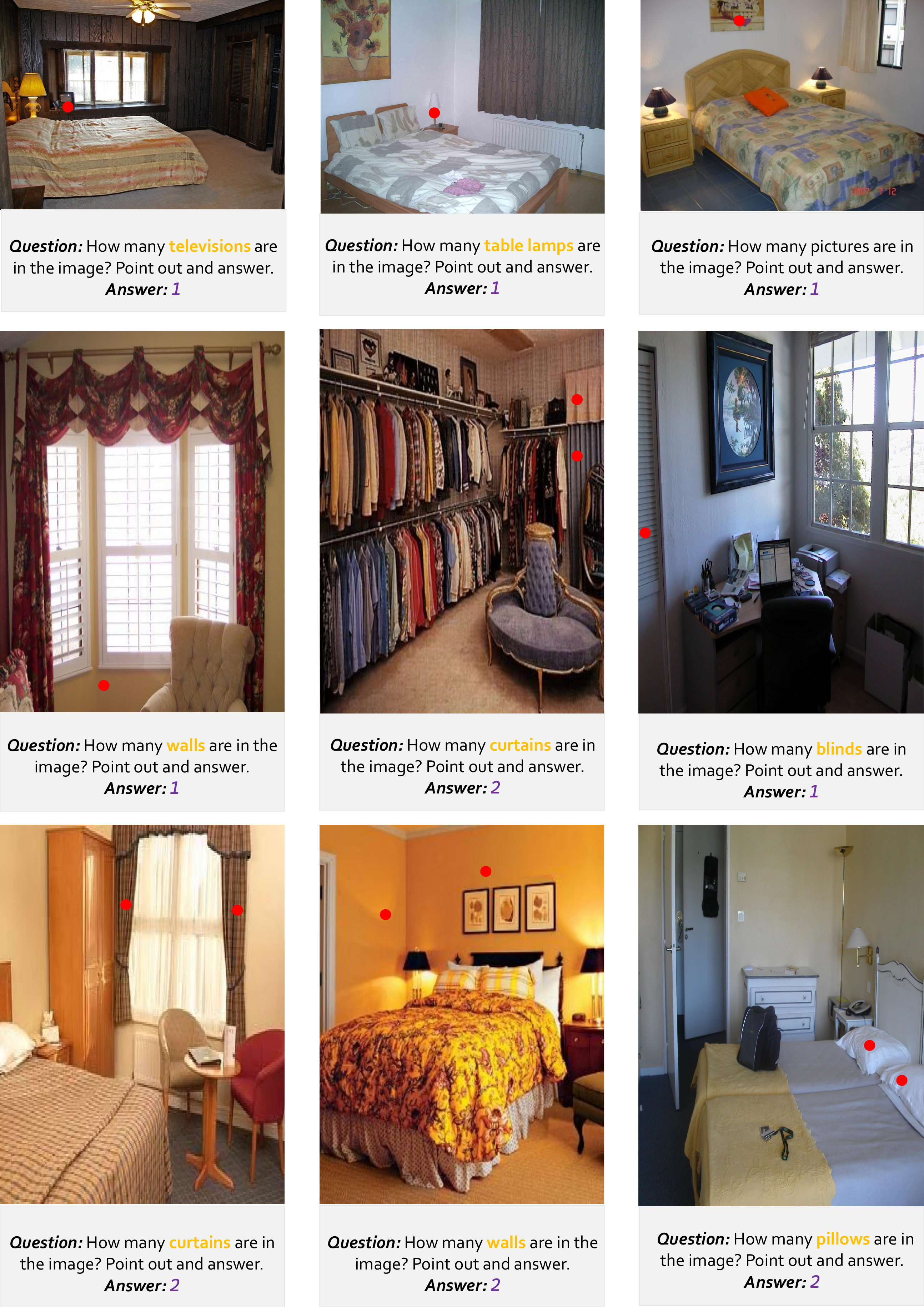}
    \caption{\textbf{BLINK Benchmark Examples of RoboBrain 2.0}. The solid circle in the diagram represents the selected object.}
    \label{fig:synthetic_2}
    \vspace{-1.5em}
\end{figure*}

Synthetic benchmarks are used to evaluate RoboBrain 2.0's performance on a variety of spatial and temporal reasoning tasks. For instance, in the BLINK benchmark, which assesses depth perception and spatial relation understanding, the model achieves high accuracy in identifying the relative positions and distances of objects. In the CV-Bench benchmark, which evaluates 3D spatial understanding, RoboBrain 2.0 demonstrates its ability to accurately process and reason about 3D scenes. These synthetic benchmarks provide a comprehensive evaluation of the model's capabilities across different reasoning dimensions. As shown in ~\Cref{fig:synthetic_1}-\Cref{fig:synthetic_2}, the model consistently performs well across various synthetic benchmarks, showcasing its robust abilities.

\section{Prompts Details}
\label{sec:app:prompt_detail}

This section outlines the system prompts for various spatial understanding and planning tasks assigned to a robot with advanced visual and analytical capabilities. Each task requires simulating visual-spatial reasoning, leveraging visual inputs as if directly perceiving the scene, and generating step-by-step reasoning processes within \texttt{<think></think>} tags, with answers in \texttt{<answer></answer>} tags. 
Reasoning is kept concise (200--500 words) and follows a five-stage process tailored to each task. 
Instructions emphasize direct visual-spatial language, avoiding abstract references to input data (e.g., ``based on the description'') and maintaining the robot's role.

\subsection{Spatial Understanding: Coordinates -- Pointing}
The robot is tasked with identifying specific points within an image based on visual criteria, such as locating points in a vacant area on a delineated plane. The reasoning process includes:

\textbf{Object Analysis} Observe the object's shape, size, and spatial relationships (e.g., a red rectangular border delineating a plane with vacant patches).

\textbf{Capability Assessment} Relate visual processing capabilities to identifying vacant regions and pinpointing coordinates.

\textbf{Contextual Relevance} Focus on the task requirement to select points within the vacant area.

\textbf{Verification} Ensure selected coordinates lie within boundaries and are distinct.

\textbf{Point Conclusion} Output coordinates as a list of tuples, justified by visual analysis.

\textbf{Example Prompt}: Locate points within {a vacant area on a plane outlined by a red border}. Your answer should be formatted as a list of tuples, i.e. $[(x_1, y_1), (x_2, y_2), \dots]$, where each tuple contains the x and y coordinates of a point satisfying the conditions above. The coordinates should indicate the normalized pixel locations of the points in the image.

\textbf{Example Output w/ Thinking}: \texttt{<think> $\dots$ </think><answer>}$[(296, 282), (321, 256), \dots]$\texttt{</answer>}.

\subsection{Spatial Understanding: Coordinates -- Trajectory}
The robot predicts a sequence of key trajectory points to achieve a goal, such as reaching an object. The reasoning process includes:

\textbf{Object Analysis} Identify the target object's properties and spatial relationships (e.g., a banana on a plate with potential obstacles nearby).

\textbf{Capability Assessment} Use joint control to plan smooth end-effector paths, avoiding obstacles.

\textbf{Contextual Relevance} Ensure the trajectory aligns with the goal (e.g., reaching the banana).

\textbf{Verification} Confirm the path avoids obstacles and reaches the target.

\textbf{Trajectory Conclusion} Output trajectory points as $[ [x_1, y_1], [x_2, y_2], \dots ]$, justified by visual and kinodynamic analysis.

\textbf{Example Prompt}: You are a robot using the joint control. The task is ``Reach for a banana on a plate''. Please predict up to 10 key trajectory points to complete the task. Your answer should be formatted as a list of tuples, i.e. $[ [x_1, y_1], [x_2, y_2], \dots ]$, where each tuple contains the x and y coordinates of a point."

\textbf{Example Output w/ Thinking}: \texttt{<think> $\dots$ </think><answer>}$[ [116, 114], [153, 97], \dots ]$.\texttt{</answer>}.

\subsection{Spatial Understanding: Bounding Box -- Affordance}
The robot identifies an affordance area for interaction with an object, such as grasping a handle. The reasoning process includes:

\textbf{Object Analysis} Describe the object's shape, size, and material properties (e.g., a blue coffee mug with a handle, inferred as ceramic from sheen).

\textbf{Capability Assessment} Assess gripper compatibility with the object's features (e.g., handle size vs. gripper opening).

\textbf{Contextual Relevance} Align with the task goal (e.g., preparing coffee by grasping the mug).

\textbf{Verification} Confirm the affordance area suits the interaction and is within reach.

\textbf{Affordance Conclusion} Output the bounding box as $[x_{min}, y_{min}, x_{max}, y_{max}]$, justified by visual compatibility.

\textbf{Example Prompt}: You are a robot using the joint control. The task is ``hold a coffee mug''. Please predict a possible affordance area of the end effector.

\textbf{Example Output w/ Thinking}: \texttt{<think> $\dots$ </think><answer>}$[915, 408, 1109, 533]$.\texttt{</answer>}.

\subsection{Spatial Understanding: Freeform Q\&A -- General Spatial Analysis}
The robot answers questions about spatial relationships or action outcomes based on one or more images. The reasoning process includes:

\textbf{Scene Perception} Detail prominent features and their spatial arrangement (e.g., a metallic gripper above a green book on a shelf).

\textbf{Task Objective Interpretation} Clarify the question's focus (e.g., predicting the outcome of a gripper's trajectory).

\textbf{Focused Visual Analysis} Examine relevant scene elements or project actions (e.g., a yellow trajectory toward a lower shelf).

\textbf{Relational Reasoning} Synthesize observations to form a hypothesis, evaluating provided options.

\textbf{Conclusion Derivation} Output the answer, justified by visual evidence and logical reasoning.

\textbf{Example Prompt}: Predict the outcome of a gripper following a yellow trajectory. Options: (A) place book on lower shelf; (B) place book on upper shelf.

\textbf{Example Output w/ Thinking}: \texttt{<think> $\dots$ </think><answer>}(A)\texttt{</answer>}.

\subsection{Temporal Understanding: Long-horizon Planning}
The robot determines the next action in a task (e.g., cooking) based on a sequence of images and the current view. The reasoning process includes:

\textbf{Task Progress Analysis} Interpret completed actions from the sequence (e.g., onions peeled and sliced on a cutting board).

\textbf{Current Scene Analysis} Describe the current view's objects and state (e.g., frying pan on hob, oil container nearby).

\textbf{Contextual Relevance} Align with the task goal (e.g., cook onions by preparing the pan).

\textbf{Action Option Evaluation} Assess options for suitability (e.g., pour oil vs. peel onion, considering onions are already prepared).

\textbf{Next Action Conclusion} Output the next action, justified by visual evidence and task flow.

\textbf{Example Prompt}: Prepare and cook onions; choose the next action (options: pour oil, turn up hob, etc.). 
\textbf{Example Output w/ Thinking}: \texttt{<think> $\dots$ </think><answer>}Pour oil.\texttt{</answer>}.

\subsection{Temporal Understanding: Closed Loop Conversation}
The robot answers a question within a conversation history, leveraging prior visual inputs and responses. The reasoning process includes:

\textbf{Task Progress Recall} Recap previous actions and their outcomes (e.g., opened the fridge to access ingredients).

\textbf{Initial Analysis} Focus on current visual input relevant to the question (e.g., a coffee machine on the countertop).

\textbf{Contextual Relevance} Align with the current task goal (e.g., flipping the coffee machine switch).

\textbf{Action Option Evaluation} Assess options for logical progression based on history and current state.

\textbf{Next Action Conclusion} Output the action, justified by visual evidence and conversation context.

\textbf{Example Prompt}: The task is ``Flip the coffee machine switch after opening the fridge.'' After you have finished <action> , you can see <image>, and the feedback of final action is xxx. What is your next action?

\textbf{Example Output w/ Thinking}: \texttt{<think> $\dots$ </think><answer>}Toggle on Coffee Machine.\texttt{</answer>}.

\subsection{Temporal Understanding: Multi-Robot Planning}

The robot coordinates actions with other robots to achieve a common goal, devided into global task decomposition and agent-based tool-calling.

\begin{figure*}[!ht]
    \centering
   \vspace{-0.2cm}
    \includegraphics[width=0.85\linewidth]{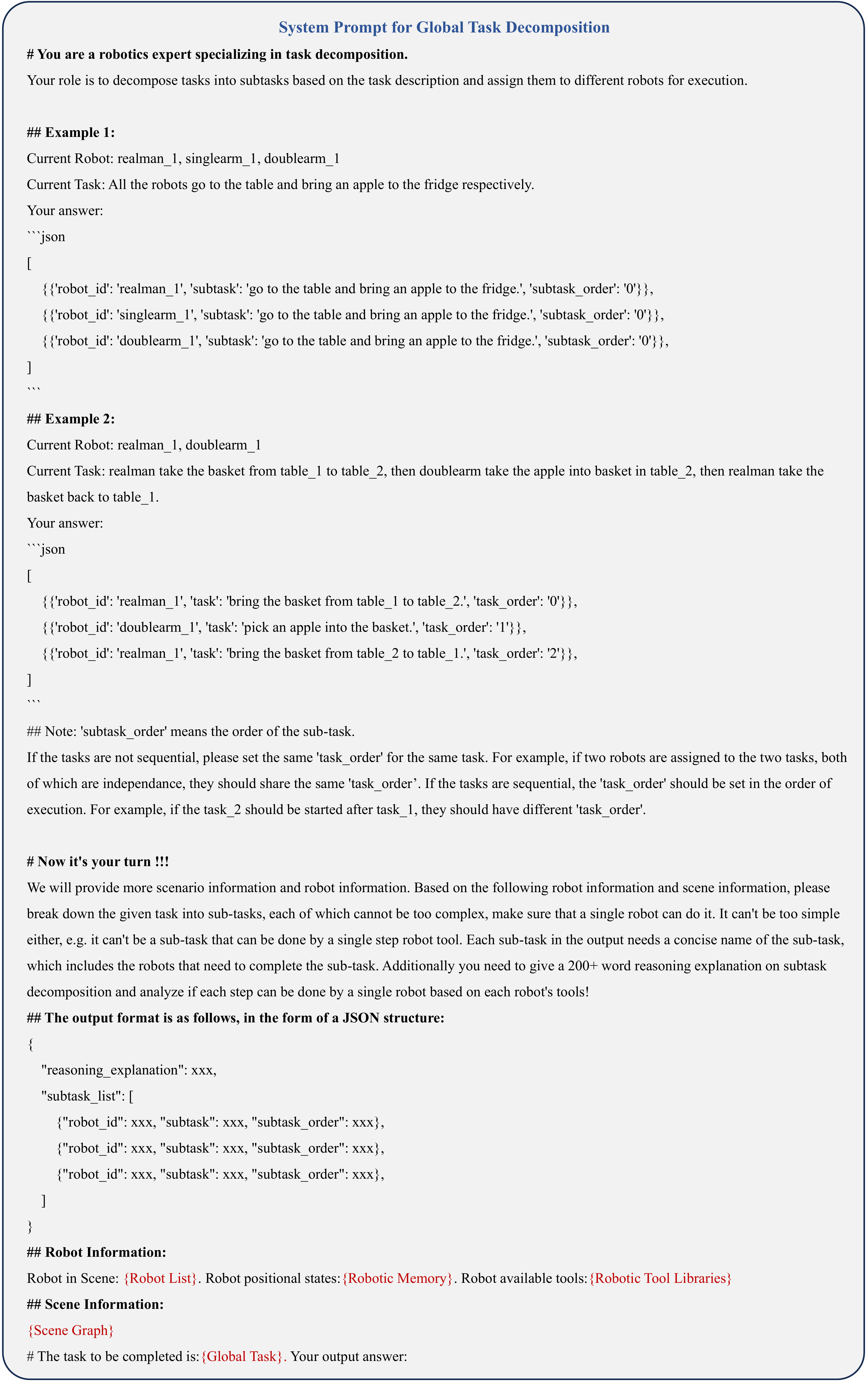}
    \caption{
        Prompt for global task decomposition. 
        }
       \vspace{-0.5cm}
    \label{fig:prompt1}
\end{figure*}

\textbf{Example Prompt for Global Task Decomposition}: Please Refer to \Cref{fig:prompt1}.

\begin{figure*}[!ht]
    \centering
   \vspace{-0.2cm}
    \includegraphics[width=0.85\linewidth]{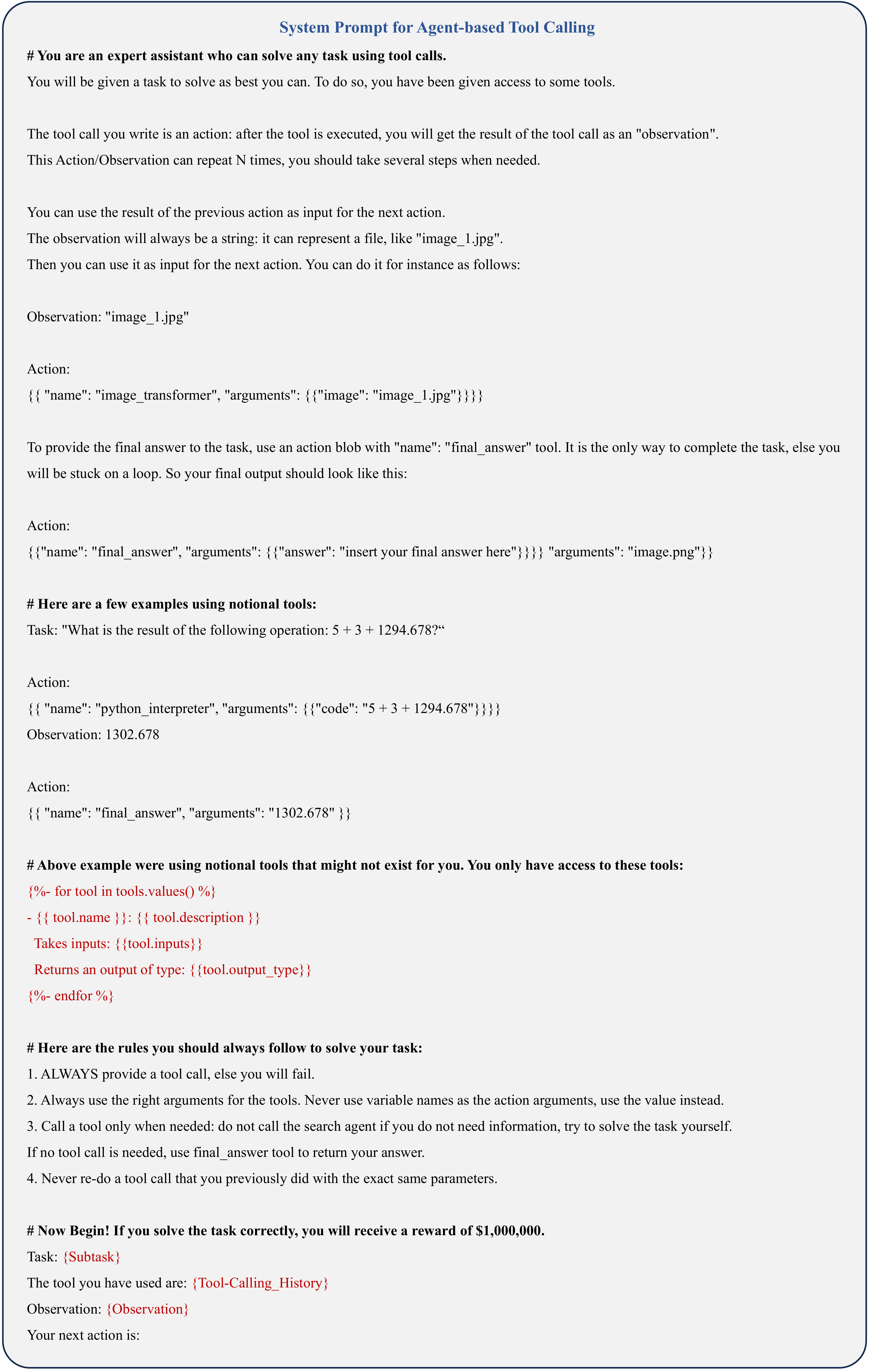}
    \caption{
        Prompt for agent-based tool calling. 
        }
       \vspace{-0.5cm}
    \label{fig:prompt2}
\end{figure*}

\textbf{Example Prompt for Agent-based Tool-calling}: Please Refer to \Cref{fig:prompt2}.

\textbf{Example Output w/ Thinking}: \texttt{<think> $\dots$ </think><answer>}Graph of TaskFlow\texttt{</answer>}.

\end{document}